\documentclass{article}

\usepackage[utf8]{inputenc}
\usepackage{authblk}
\usepackage{setspace}
\usepackage[margin=1.25in]{geometry}
\usepackage{graphicx}
\graphicspath{ {./figures/} }
\usepackage{amsmath}

\usepackage{bm}
\usepackage{amsthm,amsmath,amssymb}
\usepackage{mathrsfs}
\usepackage{booktabs}
\usepackage{multirow}
\usepackage{multicol}
\usepackage{float}
\usepackage{subfig}
\usepackage[table]{xcolor}
\usepackage{graphicx}
\usepackage{multirow}
\usepackage{enumitem}
\usepackage{bbding}
\usepackage{mathrsfs}
\usepackage{color}
\usepackage{colortbl}
\usepackage{rotating}
\usepackage{array}

\newcommand{\tabincell}[2]{\begin{tabular}{@{}#1@{}}#2\end{tabular}}
\definecolor{LightCyan}{rgb}{0.88,1,1}
\newcommand{\gray}[1]{\textcolor{gray}{#1}}

\usepackage{xcolor}
\usepackage{adjustbox}

\usepackage{tabularx}
\usepackage{etoolbox}
\usepackage{makecell}
\usepackage{pifont}

\usepackage{wrapfig}
\usepackage{cutwin}
\usepackage{alltt}

\definecolor{LightCyan}{RGB}{0.88,1,1}
\definecolor{sgreen}{RGB}{30, 150, 30} 
\definecolor{cvprblue}{RGB}{0.21,0.49,0.74}

\usepackage[style=nejm, citestyle=numeric-comp, sorting=none]{biblatex}
\usepackage[breaklinks,colorlinks,allcolors=cvprblue]{hyperref}

\title{Scalable Audio-Visual Masked Autoencoders for Efficient Affective Video Facial Analysis}

\author[1]{Xuecheng Wu}
\author[2*]{Junxiao Xue}
\author[3]{Xinyi Yin}
\author[1]{Yunyun Shi}
\author[4]{Liangyu Fu}
\author[1]{Danlei Huang}
\author[5]{Yifan Wang}
\author[1]{Jia Zhang}
\author[6]{Jiayu Nie}
\author[2]{Jun Wang}

\affil[1]{School of Computer Science and Technology, Xi'an Jiaotong University}
\affil[2]{Research Center for Space Computing System, Zhejiang Lab}
\affil[3]{School of Cyber Science and Engineering, Zhengzhou University}
\affil[4]{School of Software, Northwestern Polytechnical University}
\affil[5]{Institute of Advanced Technology, University of Science and Technology of China}
\affil[6]{Inspur Group}
\affil[*]{Address correspondence to: xuejx@zhejianglab.cn} 

\date{}

\begin{document}
\maketitle

\begin{abstract}
Affective video facial analysis (AVFA) has emerged as a key research field for building emotion-aware intelligent systems, yet this field continues to suffer from limited data availability. In recent years, the self-supervised learning (SSL) technique of Masked Autoencoders (MAE) has gained momentum, with growing adaptations in its audio-visual contexts. While scaling has proven essential for breakthroughs in general multi-modal learning domains, its specific impact on AVFA remains largely unexplored. Another core challenge in this field is capturing both intra- and inter-modal correlations through scalable audio-visual representations. To tackle these issues, we propose AVF-MAE++, a family of audio-visual MAE models designed to efficiently investigate the scaling properties in AVFA while enhancing cross-modal correlation modeling. Our framework introduces a novel dual masking strategy across audio and visual modalities and strengthens modality encoders with a more holistic design to better support scalable pre-training. Additionally, we present the Iterative Audio-Visual Correlation Learning Module, which improves correlation learning within the SSL paradigm, bridging the limitations of previous methods. To support smooth adaptation and reduce overfitting risks, we further introduce a progressive semantic injection strategy, organizing the model training into three structured stages. Extensive experiments conducted on 17 datasets, covering three major AVFA tasks, demonstrate that AVF-MAE++ achieves consistent state-of-the-art performance across multiple benchmarks. Comprehensive ablation studies further highlight the importance of each proposed component and provide deeper insights into the design choices driving these improvements. Our code and models have been publicly released at \href{https://github.com/XuecWu/AVF-MAE++}{Github}.
\end{abstract}

\begin{figure}[t!]
\centering
\includegraphics[scale=0.5]{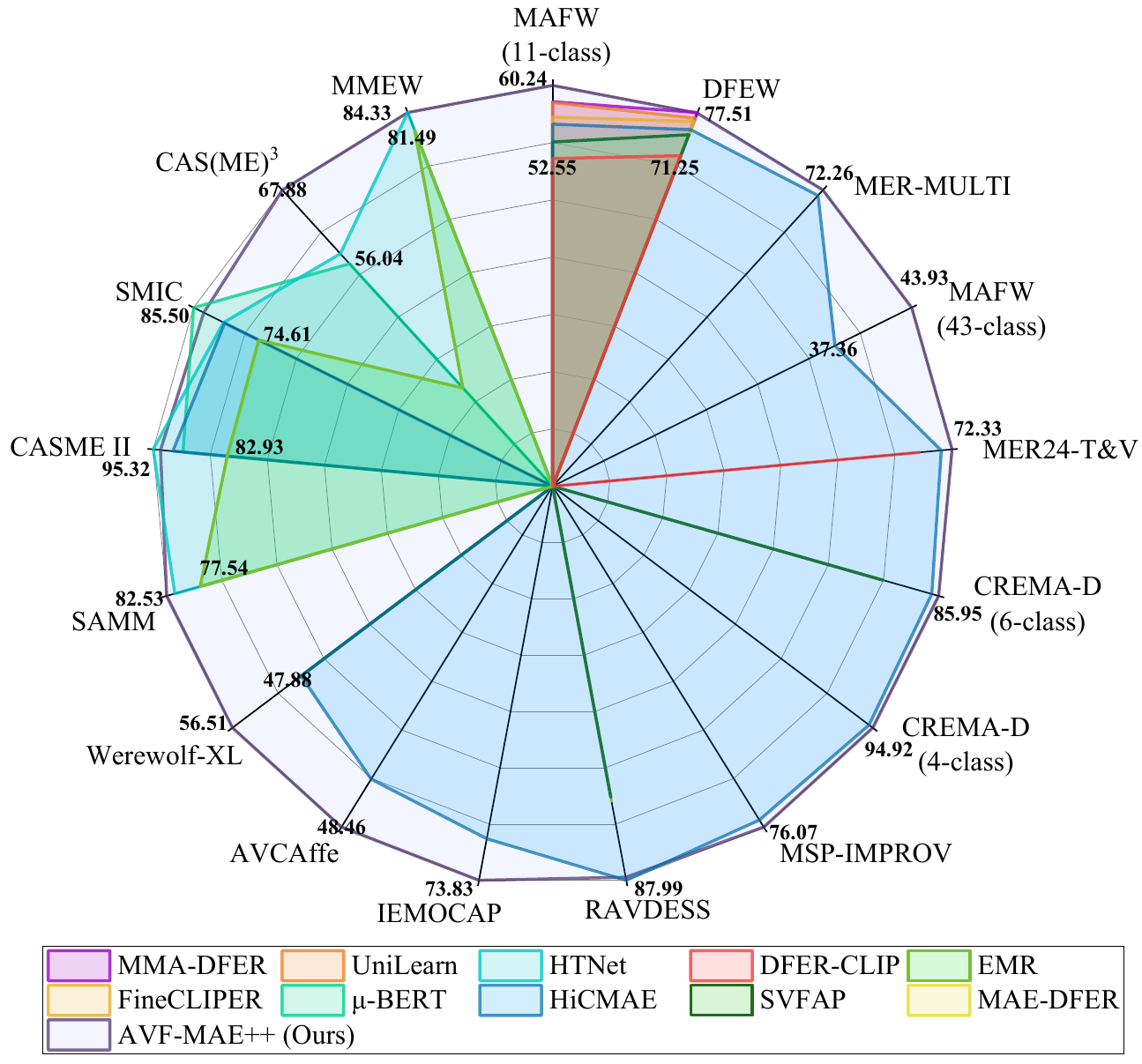}
\vspace{-0.5em}
\caption{Performance comparisons of AVF-MAE++ and state-of-the-art AVFA methods on 17 datasets across CEA, DEA, and MER tasks. Notably, we report the averaged results over dimensions on both Werewolf-XL~\cite{62_zhang2021werewolf} and AVCAffe~\cite{61_sarkar2023avcaffe} datasets.}
\label{radar-fig}
\end{figure}

\section{Introduction}
\label{sec:intro}

Affective Video Facial Analysis (AVFA) focuses on recognizing and interpreting human emotions from facial video streams and has attracted significant interest due to its potential applications in areas such as human–computer interaction (HCI) \cite{14_pantic2000automatic} and conversational systems \cite{99_schuller2018age,wang2024dual}. Typically, individuals express and perceive emotions through the synergies of multiple modalities, with early psychological studies confirming that the combined audio-visual cues (\textit{e.g.}, facial expressions and prosody) account for roughly 93\% of emotion perception \cite{16_mehrabian2017communication,17_wu2014survey}. Consequently, audio-visual AVFA has witnessed rapid advances over recent decades. In particular, with the rise of deep learning and the availability of annotated datasets, supervised deep neural networks (DNNs) have largely supplanted hand-crafted features as the dominant framework for AVFA \cite{18_zhang2023transformer,19_zhang2017learning,20_hossain2019emotion}. Despite these advances, two critical limitations hinder further progress: \textbf{(1)} High-capacity DNNs are prone to overfitting when trained on limited data, which restricts performance improvements; \textbf{(2)}The acquisition of large-scale high-quality annotations is costly, as emotional expressions are often sparse in video sequences and subjective to human interpretation \cite{3_sun2024svfap,wu2025towards}. These two challenges jointly constrain the supervised AVFA from achieving real-world deployment.

To address the inherent drawbacks of supervised approaches, a natural solution is to leverage abundant unlabeled video data to alleviate the data scarcity in AVFA. Consequently, self-supervised learning (SSL) techniques have gained considerable traction in this field, with Masked Autoencoders (MAE) emerging as a particularly influential paradigm~\cite{22_he2022masked}. Specifically, MAE aims to reconstruct the raw data from masked facial videos, leading to the emergence of various visual and audio-visual AVFA MAE methods~\cite{1_sun2024hicmae,2_sun2023mae,3_sun2024svfap}. Meanwhile, following the promising findings in image and language domains~\cite{22_he2022masked,23_bao2021beit}, recent work such as VideoMAE V2~\cite{4_wang2023videomae} underscores that scaling both model size and training data is critical to unlocking substantial performance improvements. However, very little research has explored the scaling properties of MAE pre-training for AVFA, which is more severe in audio-visual field. While \cite{1_sun2024hicmae,3_sun2024svfap} provides models with varying capacities, their largest size generally reaches the ten-million scale, lagging behind those in general domains. More importantly, successful large-scale audio-visual MAE pre-training hinges on robustly modeling both intra- and inter-modal dependencies, given that emotions are often conveyed through co-expressive cues. Nonetheless, current SSL-based AVFA methods still struggle to effectively capture such cross-modal correlations~\cite{5_wu2023emotions,6_zong2024self}.

To fill these gaps, we aim to explore the scaling properties of audio-visual MAE for AVFA, with particular attention to capturing both intra- and inter-modal correlations and pushing performance boundaries on diverse downstream datasets. Building upon HiCMAE~\cite{1_sun2024hicmae}, we scale the audio-visual MAE and further conduct million-level data scaling for the pre-training stage to fully exploit its potential. Moreover, we introduce dedicated components to explicitly enhance the capture of audio-visual correspondences, thereby addressing limitations observed in prior AVFA approaches. Nevertheless, achieving robust training of high-capacity audio-visual MAEs on such large-scale data and ensuring consistent gains across all downstream tasks requires careful consideration of several remaining issues:

\textbf{(1)} Computational costs and memory consumption remain the major bottlenecks in scaling audio-visual MAE. Although~\cite{1_sun2024hicmae} adopts the asymmetric encoder-decoder design from~\cite{7_tong2022videomae}, it still struggles to fully support the pre-training for large-scale models. Motivated by the dual masking strategy for the asymmetric architecture from~\cite{4_wang2023videomae}, we adaptively present the audio-visual dual masking strategy, yielding a more efficient self-supervised proxy objective. Meanwhile, the standard global spatio-temporal attention mechanism introduces quadratic scaling complexity, and large redundancy (\textit{e.g.}, facial symmetry) exists in 3D facial video data, rendering the expenses suboptimal. We thus flexibly introduce a local-global interaction attention paradigm within modality encoders, while elevating the holistic perspective to compensate for its weakness in global information flow. \textbf{(2)} A huge amount of unlabeled data is indispensable for scalable pre-training. However, unlike image-based affective analysis, existing AVFA datasets are relatively small. A straightforward yet effective remedy is to combine data from diverse sources. Following prior work~\cite{1_sun2024hicmae,2_sun2023mae}, we curate a large-scale corpus by integrating unlabeled videos from speaker recognition datasets, ultimately constructing a pre-training dataset of approximately 1.36M clips. \textbf{(3)} Cross-modal correlation modeling remains insufficient in existing approaches, which typically rely on self-attention and cross-attention layers. These designs limit inter-modal interactions and lack hierarchical-aware multi-scale semantic integration, while often underutilizing multi-modal features in building comprehensive representations. To address this, we propose the IAV-CL Module (\textbf{I}teratively \textbf{A}udio-\textbf{V}isual \textbf{C}orrelations \textbf{L}earning Module), which enhances the capture of intra- and inter-modal dependencies in a multi-level manner. \textbf{(4)} A key challenge for SSL methods is smoothly adapting the pre-trained models to downstream datasets. Direct fine-tuning on small-scale datasets usually leads to severe overfitting, preventing pre-trained models from realizing their full potential.  To alleviate this, we introduce a Progressive Semantics Injection (PSI) strategy, which leverages supervised hybrid datasets from diverse sources as an intermediate stage, effectively bridging pre-training and downstream fine-tuning.

Based on the above analysis, we propose a series of audio-visual MAE termed AVF-MAE++. By jointly scaling model capacity and data size, along with the introduced IAV-CL Module, we further present the PSI strategy to construct a three-stage progressive training pipeline. This pipeline consists of (i) large-scale audio-visual masked pre-training, (ii) post-pretraining with supervised hybrid datasets, and (iii) task-specific fine-tuning on downstream benchmarks. To validate the effectiveness of our framework, we conduct extensive pre-training and evaluate model performance across three key downstream tasks involving 17 datasets. As shown in Fig.~\ref{radar-fig}, AVF-MAE++ consistently surpasses state-of-the-art supervised and self-supervised methods. Notably, it is the first approach to achieve over 60\% WAR on the challenging MAFW (11-class)~\cite{63_liu2022mafw} benchmark. To the best of our knowledge, this represents the first systematic exploration of scaling properties for audio-visual MAEs in AVFA, offering a solid foundation for future studies. Beyond advancing affective video facial analysis, our contributions extend to broader areas, including multimodal large language models (MLLMs)~\cite{10_cheng2024emotion,wang2024multimodal}, talking-face generation~\cite{11_wang2022one}, deepfake detection~\cite{157_wang2024building,wu2025hola}, and explainable emotional reasoning~\cite{13_lian2023explainable,5_wu2023emotions}.

This paper extends our preliminary conference version~\cite{wucvpr} with several major enhancements that significantly strengthen the study. In particular, \textbf{(1)} we provide enriched discussions on the motivation, datasets, model architecture, experimental findings, and implementation details; \textbf{(2)} we elaborate the methodology with finer granularity, including the design motivation, implementation details, and parameter settings of each module, making the pipeline more transparent, coherent, and reproducible; \textbf{(3)} we give a detailed account of dataset construction, especially the preprocessing strategies adopted across the three training stages, offering practical insights for future AVFA research; \textbf{(4)} we contribute extensive qualitative and quantitative analyses at both dataset and method levels, aiming to establish solid foundations for subsequent advances; \textbf{(5)} to more thoroughly validate the scalability and applicability of AVF-MAE++, we further explore its potential applications in deepfake detection~\cite{157_wang2024building}, talking-face generation~\cite{11_wang2022one}, and MLLMs~\cite{10_cheng2024emotion,wang2024multimodal,zhang2025hkd4vlm}.

\section{Related Work}
\label{sec:Related Work}

\subsection{Audio-visual Affective Video Facial Analysis}

Most current works in AVFA concentrate on the supervised paradigm, where powerful representations are acquired to enhance performance \cite{25_chumachenko2024mma,26_zhao2023prompting,27_zhang2023weakly}. Recently, audio-visual AVFA has attracted extensive attention owing to the natural correspondence of audio-visual modalities \cite{5_wu2023emotions,28_goncalves2024versatile}. The main aspects of audio-visual AVFA chiefly involve unimodal feature extractions and audio-visual information fusion. Along with the advancement of deep learning and large-scale datasets, various audio-visual feature extractors have appeared \cite{30_wang2023cam++,31_liu2022video,21_zhang2024mart}. Recently, with the significant discoveries of SSL in general domains, many large pre-trained models show success in audio or video emotion analysis \cite{86_hsu2021hubert,2_sun2023mae,3_sun2024svfap}. Concerning audio-visual information fusion \cite{1_sun2024hicmae,29_ma2019audio}, the most frequently used fusion strategy is model-level fusion, which is primarily constructed on self-attention and cross-attention mechanisms \cite{18_zhang2023transformer,33_sun2023efficient,34_tsai2019multimodal}. Despite encouraging results, they still show limitations in capturing intra- and inter-modal correlations. Moreover, the prevalent supervised paradigm in AVFA is heavily limited by the shortage of emotional data and domain shifts.

\subsection{Masked Audio-Visual Modeling}

Masked data modeling acquires representations by reconstructing the masked content of input. Inspired by its recent achievements in the general vision domain, many works \cite{35_shi2022learning,36_guo2024crossmae,37_liu2024masked,38_woo2024speech,42_huang2024mavil,40_georgescu2023audiovisual} have tried to extend it to the audio-visual domain, showing impressive results across various downstream tasks. Among them, MAE-style methods have gained considerable attention due to their efficient learning capabilities, such as AV-MAE \cite{40_georgescu2023audiovisual}, CAV-MAE \cite{39_gong2022contrastive}, and DeepAVFusion \cite{41_mo2024unveiling}. AV-MAE \cite{40_georgescu2023audiovisual} introduces audio-visual MAE to mutually enhance self-supervised representation learning. CAV-MAE \cite{39_gong2022contrastive} integrates MAE and contrastive learning, resulting in the joint and coordinated audio-visual representations. Mo et al. \cite{41_mo2024unveiling} introduce DeepAVFusion using MAE and the early fusion transformers. Although these MAE-style methods show exciting performance on general audio-visual tasks, the learned representations are usually unsuitable for AVFA, as they are trained on generic scenarios or action videos rather than facial videos in AVFA. VQ-MAE-AV \cite{43_sadok2024audiovisual}, recently introduced, is a vector-quantized MAE for audio-visual AVFA using two-stage training. Xiang et al. \cite{44_xiang2024multimae} present various sequence fusion strategies using pre-trained MAE. Sun et al. \cite{1_sun2024hicmae} introduce HiCMAE with the three-pronged learning strategies, obtaining remarkable results. Meanwhile, the scaling properties extensively investigated in the general vision domain have not been fully examined in AVFA, causing substantial gaps. In this work, we introduce AVF-MAE++, building on HiCMAE as the foundational method, to bridge these gaps and advance the development of AVFA.

\subsection{Masked Autoencoders Scaling}

Based on the foundational success of MAE, researchers have extensively investigated its scaling properties across various vision fields. Singh et al. \cite{45_singh2023effectiveness} present an additional pre-pretraining stage to improve MAE initializations. Han et al. \cite{46_han2024efficient} develop \textit{Efficient} MAE with a novel loss and a decoder masking strategy. VideoMAE \cite{7_tong2022videomae} and MAE-ST \cite{48_feichtenhofer2022masked} have trained the video transformers with millions of parameters. VideoMAE V2 \cite{4_wang2023videomae} scales the VideoMAE \cite{7_tong2022videomae} in both model and data with a dual masking strategy. SimMIM \cite{47_xie2023data} systematically investigates the data scaling capability of masked data modeling. In AVFA, there exist some works that have initially investigated the scaling properties of MAE \cite{1_sun2024hicmae,2_sun2023mae,3_sun2024svfap}. However, existing efforts only concentrate on scaling up model sizes on a limited scale, with few explorations of data scaling. To this end, we first examine the scaling properties of audio-visual MAE in both model and data with the currently largest pretraining dataset for AVFA.
\section{Methodology}
\label{sec:Mthods}

\begin{figure*}[t!]
\setlength{\belowcaptionskip}{-1.5em}
\centering
\includegraphics[height=10.2cm,width=\textwidth]{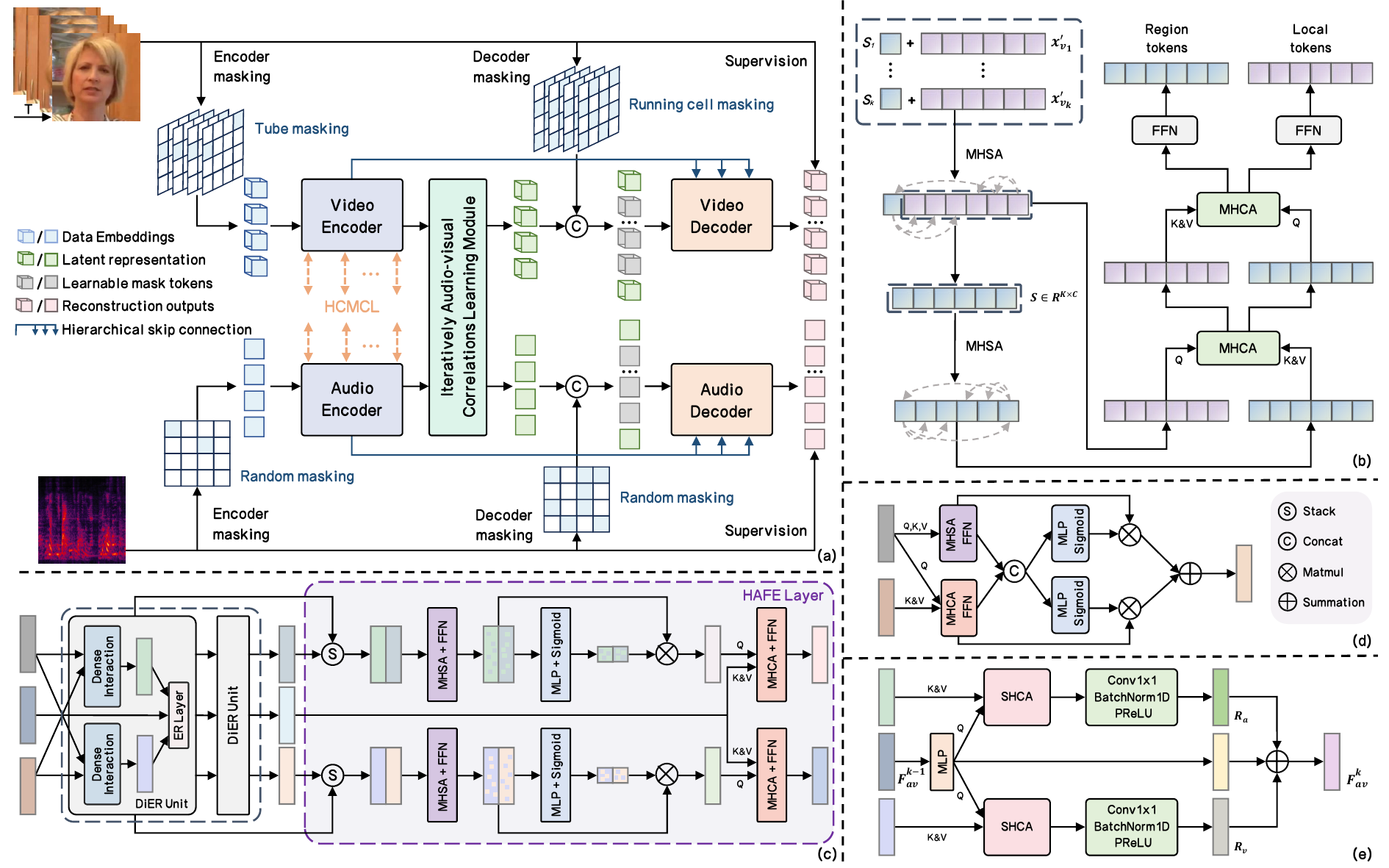}
\vspace{-1.0em}
\caption{The illustrations of AVF-MAE++. (a) The pre-training pipeline with our new audio-visual dual masking strategy. (b) One layer of improved modality encoder. (c) IAV-CL Module. (d) \& (e) The \textit{dense interaction} and \textit{evolutionary refinement} layers of one introduced DiER Unit.}
\label{fig-AVF-MAE++}
\end{figure*}

In this section, we begin by revisiting the foundational work HiCMAE~\cite{1_sun2024hicmae} in Sec.~\ref{3.1}. We then introduce the audio-visual dual masking strategy (Sec.~\ref{3.2}), improved modality encoder (Sec.~\ref{3.3}), and the IAV-CL Module (Sec.~\ref{3.4}), as shown in Fig.~\ref{fig-AVF-MAE++}. Finally, we elaborate on the details of dual scaling and the progressive adaptation strategy (Sec.~\ref{3.5}).

\subsection{HiCMAE Revisited}
\label{3.1}
HiCMAE~\cite{1_sun2024hicmae} follows the asymmetric encoder-decoder architecture of~\cite{22_he2022masked} and proposes a three-pronged hierarchical strategy. Next, we briefly revisit its implementation details. 

\noindent \textbf{Data Embedding.} A cube embedding layer and a patch embedding layer are first utilized to divide $\mathbf{X}_v \in \mathbb{R}^{T_v\times H \times W \times 3}$ and $\mathbf{X}_a \in \mathbb{R}^{T_a \times F}$, leading to token lists: $\mathbf{X}_v^{'} = \Phi_{emb}^{v} (\mathbf{X}_v)$ and $\mathbf{X}_a^{'} = \Phi_{emb}^{a} (\mathbf{X}_a)$, where $\mathbf{X}_v^{'} = \{\mathbf{X}_v^{'i}\}_{i=1}^{N_v}$ and $\mathbf{X}_a^{'} = \{\mathbf{X}_a^{'j}\}_{j=1}^{N_a}$ are the token sequences, $(\mathbf{X}_{v}^{'i}, \mathbf{X}_{a}^{'j}) \in \mathbb{R}^{1 \times C}$ are the tokens output by the embedding layers and then added with positional embeddings. Here, $N_v = \frac{T_v}{2} \times \frac{H}{16} \times \frac{W}{16}$ and $N_a = \frac{T_a}{16} \times \frac{F}{16}$ refer to the lengths of video and audio token sequences, while $C$ denotes the feature channels.

\noindent \textbf{Token Masking.} 
HiCMAE deploys the tube masking and random masking for the video and audio branches, using high masking ratios ($\rho_v = 90\%$ and $\rho_a = 80\%$). Next, only the visible tokens $\mathbf{X}_v^{''}$ and $\mathbf{X}_a^{''}$ will run into the encoders, where $\mathbf{X}_v^{''} = \{\mathbf{X}_v^{'i}\}_{i \in (1-\mathbb{M}(\rho_{v}))}$, $\mathbf{X}_a^{''} = \{\mathbf{X}_a^{'j}\}_{j \in (1-\mathbb{M}(\rho_{a}))}$, and their token lengths are $N_v' = 0.1N_v$ and $N_a' = 0.2N_a$. $\mathbb{M}(\rho_{v})$ and $\mathbb{M}(\rho_{a})$ here are the audio-visual masking maps.

\noindent \textbf{Encoder.} The encoder of HiCMAE operates on the visible tokens $\mathbf{X}_v^{''}$ and $\mathbf{X}_a^{''}$ with two modality-specific encoders and a cross-modal fusion encoder: $\mathbf{E}_{a \rightarrow v}$, $\mathbf{E}_{v \rightarrow a}$ $=$ $\Phi_{enc}^{a \leftrightarrow v}(\Phi_{enc}^{v}(\mathbf{X}_v^{''}),\Phi_{enc}^{a}(\mathbf{X}_a^{''}))$, where the modality encoders are vanilla ViT~\cite{24_dosovitskiy2020image}, and the fusion encoder is mainly implemented using multi-head cross-attention components.

\noindent \textbf{Decoder.} The video and audio decoders, including hierarchical skip connections, respectively take the {\em combined} tokens as inputs and reconstruct data with narrower and shallower ViT: $\hat{\mathbf{X}}_m$ $=$ $\Phi_{dec}^{m}(\mathbf{E}_m^{c})$, where the {\em combined} tokens $\mathbf{E}_m^{c}$ is the concatenated sequence of encoded tokens $\mathbf{E}_{\bar{m} \rightarrow m}$ and the learnable masked tokens $[\texttt{MASK}]_m$ (with position embeddings), the token length $N_m^d$ $=$ $N_m$, and $m \in \{a,v\}$.

\noindent \textbf{Pre-training Loss.} The pre-training object is to minimize the combination of modality-specific \textit{Mean Square Error} (MSE) Losses and the introduced \textit{HCMCL} Loss~\cite{1_sun2024hicmae}, \textit{i.e.},
\begin{equation}
\mathcal{L} = (\mathcal{L}_{\textrm{MSE}}^a + \mathcal{L}_{\textrm{MSE}}^v) + \lambda \cdot \sum_{k=1}^{N_c} \mathcal{L}_{\textrm{InfoNCE}}(\mathbf{e}^{k}_{a}, \mathbf{e}^{k}_{v}),
\label{eq1}
\end{equation}
where $\lambda$ is the weight factor, $N_c$ is the number of selected encoder layers in hierarchical skip connections, and $\mathbf{e}^{k}_{m}$ is a batch of sample-level features to adopt HCMCL Loss.

\noindent \textbf{Downstream Fine-tuning.} After pre-training, the overall encoder incorporating hierarchical feature fusion will be deployed to target fine-tune on the downstream tasks.

\subsection{Audio-Visual Dual Masking Strategy}
\label{3.2}

As detailed in Section~\ref{3.1}, a significant redundancy exists in HiCMAE because its decoders are required to process the complete set of tokens. To address this inefficiency, the more recent VideoMAE V2~\cite{4_wang2023videomae} employs a dual masking strategy. This approach enhances the efficiency of video pre-training by providing the decoder with inputs from two sources: the tokens that were visible under the encoder mask, $\mathbb{M}_e = \mathcal{M}_e(\rho^e)$, along with a subset of the remaining tokens that are revealed by a distinct decoder mask, $\mathbb{M}_d = \mathcal{M}_d(\rho^d)$.

Building on this concept, we introduce an audio-visual dual masking strategy that incorporates both encoder masking, $\mathcal{M}_e$, and decoder masking, $\mathcal{M}_d$, for the audio and video streams, as illustrated in Figure~\ref{fig-AVF-MAE++}~(a).

For the specific implementation, the encoder masking, $\mathcal{M}_e^m$, is consistent with the method used in HiCMAE. The video decoder mask, $\mathcal{M}_d^v$, follows the work of~\cite{4_wang2023videomae} by adaptively employing running cell masking~\cite{50_qing2023mar} to enhance complementary information during the partial reconstruction process. For the audio decoder mask, $\mathcal{M}_d^a$, we utilize random masking, as prior research~\cite{49_huang2022masked} has shown that audio MAE models learn efficiently by predicting adjacent contexts. In line with the approach in ~\cite{4_wang2023videomae}, the decoder masking ratio, $\rho^d$, is set to 50\% for both the audio and video branches. The introduction of this dual masking strategy results in a new formulation for the \textit{combined} token sequence that is input to the modality decoder:

\begin{equation}
\mathbf{E}_m^{c} = \mathbf{E}_{\bar{m} \rightarrow m} \cup \{\mathbf{M}_i^{m}\}_{i \in \mathbb{M}_d^{m}},
\label{eq2}
\end{equation}
where $\mathbf{E}_{\bar{m} \rightarrow m}$ denotes the latent representations from encoder, $\mathbf{M}_i^{m}$ is the learnable masking token with related positional embeddings, and $m \in \{a,v\}$. With this updated sequence $\mathbf{E}_m^{c}$, decoder only regards the visible tokens as the reconstruction targets. The final MSE Loss can be given as: 
\begin{equation}
\mathcal{L}_{\textrm{MSE}}^{m} = \frac{1}{(1-\rho^d_m) \cdot N_m} \sum_{i \in \mathbb{M}_d^{m} \cap \mathbb{M}_e^{m}} |\mathbf{X}_m^{i} - \hat{\mathbf{X}}_m^{i}|^2,
\label{eq3}
\end{equation}
where $\mathbf{X}_m$ and $\hat{\mathbf{X}}_m$ refer to the original input and the reconstructed output of the audio-visual modalities, respectively.

\subsection{Improved Modality Encoder}
\label{3.3}

Two primary factors impede efficient, large-scale pre-training on facial videos: the significant redundancy inherent in the data and the high computational cost of the global space-time self-attention mechanism in a standard ViT~\cite{24_dosovitskiy2020image}. To address this bottleneck, we employ an adaptively modified and enhanced version of LGI-Former~\cite{2_sun2023mae} for the uni-modal encoder, leveraging its demonstrated ability to reduce computational demands. For simplicity, our explanation will focus on a single encoder layer during the fine-tuning phase. The main distinction between this stage and pre-training lies in the number of visible tokens within each region. The original LGI-Former~\cite{2_sun2023mae} is proposed for video, which can be decomposed into three stages: \textbf{(\text{I})} local intra-region self-attention, \textbf{(\text{II})} global inter-region self-attention, and \textbf{(\text{III})} local-global interaction.

In stage \text{I}, the 3D tokens $\mathbf{X}_v^{'} \in \mathbb{R}^{\frac{T_v}{2} \times \frac{H}{16} \times \frac{W}{16} \times C}$ is first divided into $K$ non-overlapping local spatio-temporal regions of equal size $Z_v = t \times h \times w$, leading to $\mathbf{X}_{v_i}^{'} \in \mathbb{R}^{Z_v \times C}$, and $\mathbf{X}_{v_i}^{'}$ is then added with a learnable region token $\mathbf{S}_{i} \in \mathbb{R}^{1 \times C}$ ($i \in \{1, 2, ..., K\}$, $K = \frac{N_v}{Z_v}$). The self-attention then operates on their concatenation to promote local-aware features learning and aggregate information into the region token $\mathbf{S}_{i}$:
\begin{equation}
\mathbf{\hat{X}}_{v_{i}}^{'} = \textrm{MHSA}(\textrm{LN}(\textrm{C}(\mathbf{S}_i, \mathbf{X}_{v_i}^{'}))) + \textrm{C}(\mathbf{S}_i, \mathbf{X}_{v_i}^{'}),
\label{eq4}
\end{equation}
where $\mathbf{\hat{X}}_{v_{i}}^{'} \in \mathbb{R}^{(Z_{v}+1) \times C}$, $\textrm{MHSA}(\cdot)$ is the vanilla multi-head self-attention, $\textrm{LN}(\cdot)$ and $\textrm{C}(\cdot)$ denote layer normalization and concatenation operation. In particular, the calculation of $\textrm{MHSA}$ can be formulated as:
\begin{align}
\textrm{MHSA}(\mathbf{X}) &= \textrm{C}(\textrm{h}_{1}, ..., \textrm{h}_{H})\mathbf{W}^O, \\
\textrm{h}_{j} &= \textrm{Att}(\mathbf{X} \mathbf{W}^{Q}_j, \mathbf{X} \mathbf{W}^{K}_j, \mathbf{X} \mathbf{W}^{V}_j), \\
\textrm{Att}(\mathbf{Q}, \mathbf{K}, \mathbf{V}) &= \textrm{Softmax}(\mathbf{Q} \mathbf{K}^\top / \sqrt{d})\mathbf{V},
\label{eq5-7}
\end{align}
where $\mathbf{W}^{*}_j \in \mathbb{R}^{C\times d}$ ($* \in \{Q,K,V \}$), $\mathbf{W}^{O} \in \mathbb{R}^{C\times C}$, $H$ is the number of attention heads, $d=\frac{C}{H}$ is the feature dimension of each head~\cite{2_sun2023mae}, $j \in \{1,..., H\}$. In stage \text{II}, all the region tokens $\{\mathbf{S}_{i}\}_{i=1}^{K}$ are first aggregated, self-attention is then employed to exchange inter-region information between different regions with negligible costs, \textit{i.e.},
\begin{equation}
\mathbf{S} = \textrm{MHSA}(\textrm{LN}(\textrm{C}(\mathbf{S}_1, ..., \mathbf{S}_K))) + \textrm{C}(\mathbf{S}_1, ..., \mathbf{S}_K),
\label{eq5}
\end{equation}
where $\mathbf{S} \in \mathbb{R}^{K \times C}$ is the aggregated region tokens. So far, the region token $\mathbf{S}_i$ has been consolidated by discriminative information from other regions, holding a global perspective of the overall input tokens. As a result, in stage \text{III}, multi-head cross-attention between $\mathbf{X}_{v_i}^{'}$ and $\mathbf{S}$ is explicitly exploited to enable the original local tokens to access the global-aware selective information,~\textit{i.e.},
\begin{equation}
\mathbf{X}_{v_i}^{'} = \textrm{MHCA}(\textrm{LN}(\mathbf{X}_{v_i}^{'}), \textrm{LN}(\mathbf{S})) + \mathbf{X}_{v_i}^{'},
\label{eq6}
\end{equation}
where $\textrm{MHCA}(\cdot)$ refers to the vanilla multi-head cross-attention. Additionally, since the region tokens are rather important to the global evolutionary information flow across multiple encoder layers, which should emphasize the more holistic master for the local-aware intra-region information, we thus introduce the stage \text{IV} (global-local interaction):
\begin{equation}
\mathbf{S} = \textrm{MHCA}(\textrm{LN}(\mathbf{S}), \textrm{LN}(\mathbf{X}_{v_i}^{'})) + \mathbf{S}.
\end{equation}

Following this process, both the local and region tokens are passed through feed-forward networks (FFNs) to further refine their feature representations. When applying this encoder to the audio branch, the main difference is the patch embedding layer outputs 2D tokens $\mathbf{X}_a^{'} \in \mathbb{R}^{\frac{T_a}{16} \times \frac{F}{16} \times C}$, leading to different region shape. In stage \text{I}, we split $\mathbf{X}_a^{'}$ into $K$ non-overlapping local regions of equal size $Z_a = h_a \times w_a$, resulting in $\mathbf{X}_{a_i}^{'} \in \mathbb{R}^{Z_a \times C}$ ($i \in \{1, 2, ..., K\}$, $K=\frac{N_a}{Z_a}$). After conducting region division, the remaining process keeps consistent with the video branch. Finally, we take the region tokens $(\mathbf{S}_v, \mathbf{S}_a) \in \mathbb{R}^{K \times C}$ as the outputs from one encoder layer, and the overall modality encoders both consist of $N_l$ sequentially stacked layers.

\subsection{Iteratively Audio-Visual Correlations Learning}
\label{3.4}

As illustrated in Figure~\ref{fig-AVF-MAE++}~(c), we introduce the IAV-CL Module, a component designed for the iterative capture of complementary correlations. This module is constructed from two primary elements: the Dense Interactions and Evolutionary Refinement (DiER) Units and the Hierarchical Aggregations and Feedback Enhancement (HAFE) Layer.

During fine-tuning, we first stack $\{\mathbf{S}_v^n, \mathbf{S}_a^n\}_{n=1}^{N_l}$, leading to the uni-modal features $(\mathbf{F}_v$, $\mathbf{F}_a$) $\in \mathbb{R}^{K \times N_l \times C}$. We then utilize the learnable layer weights to dynamically unify features across different encoder layers followed by concatenation to output the original multi-modal feature $\mathbf{F}_{av}^0$, \textit{i.e.},
\vspace{-0.65em}
\begin{equation}
\mathbf{F}_{av}^0 = \textrm{C}(\sum_{l=1}^{N_l} \alpha^{a}_l \mathbf{F}^{l}_{a}, \sum_{l=1}^{N_l} \alpha^{v}_l \mathbf{F}^{l}_{v}),
\label{eq13}
\vspace{-0.35em}
\end{equation}
where $\mathbf{F}_{av}^0 \in \mathbb{R}^{K \times 2C}$, $\sum_{l=1}^{N_l} \alpha^{m}_l = 1$. We then simply use poolings to reshape $\mathbf{F}_v$ and $\mathbf{F}_a$ as ($\mathbf{F}_{v}^1$, $\mathbf{F}_{a}^1$) $\in \mathbb{R}^{K \times C}$. The DiER Unit is proposed to perform dense audio-visual interactions and evolutionarily refine the multi-modal feature in the simultaneous manner, which is detailed as follows:

\noindent \textbf{Dense Audio-Visual Interactions.}
Recognizing that sequentially connecting MHSA and MHCA blocks ~\cite{52_cheng2020look,1_sun2024hicmae} offers suboptimal support for dense audio-visual interactions, our approach utilizes a parallel arrangement, as depicted in Figure~\ref{fig-AVF-MAE++}~(d). The process begins by concatenating the outputs from the parallel attention blocks along the channel dimension. Subsequently, we compute channel-wise attention scores, which are then used to refine the features through a linear layer and a sigmoid function. Finally, a summation operation is applied to yield the densely interacted output features:

\begin{align}
\mathbf{F}_m^2 &= \resizebox{0.361\textwidth}{!}{$
    \sigma\left(\mathbf{W}_{s} \mathbf{F}_m^{sc} + \mathbf{b}_s\right)\mathbf{F}_{m}^s
    + \sigma\left(\mathbf{W}_{c} \mathbf{F}_m^{sc} + \mathbf{b}_c\right)\mathbf{F}_{m}^c
$}, \\
\mathbf{F}_{m}^s &= \textrm{MHSA}\big(\textrm{LN}(\mathbf{F}_m^1)\big) + \mathbf{F}_m^1, \\
\mathbf{F}_{m}^c &= \textrm{MHCA}\big(\textrm{LN}(\mathbf{F}_m^1),\ \textrm{LN}(\mathbf{F}_{\bar{m}}^1)\big) + \mathbf{F}_m^1,
\end{align}
where $\mathbf{F}_m^{sc} = \textrm{C}\left(\mathbf{F}_{m}^s, \mathbf{F}_{m}^c\right)$, $\mathbf{\sigma}(\cdot)$ is sigmoid function, $\mathbf{F}_m^2 \in \mathbb{R}^{K \times C}$, $\mathbf{W}_{*}$ and $\mathbf{b}_{*}$ ($* \in \{s,c \}$) are learnable parameters.

\noindent \textbf{Evolutionary Refinement (ER) Layer} iteratively refines the multi-modal feature, leading to the following feedback enhancements of correlations capture. We first simply employ the linear layer to transform $\mathbf{F}_{av}^0$ into $\mathbf{F}_{av}^1 \in \mathbb{R}^{K \times C}$, then attend $\mathbf{F}_{av}^1$ to audio-visual features using the single-head cross-attention (SHCA) block, which dynamically aggregates uni-modal useful information into $\mathbf{F}_{av}^1$, as illustrated in Fig.~\ref{fig-AVF-MAE++}~(e). Inspired by~\cite{53_hu2023cross}, the convolutional block incorporating one $1 \times 1$ convolution followed by the Batch Normalization and PReLU sub-layers is then introduced to generate the residual features $\mathbf{R}_a$ and $\mathbf{R}_v$, which discriminatively learn invariant audio-visual representations, \textit{i.e.},
\begin{align}
\mathbf{R}_m &= \textrm{Conv}(\textrm{Att}(\mathbf{F}_{av}^1, \mathbf{F}_m^2, \mathbf{F}_m^2)),
\label{eq-residual}
\end{align}
where $\textrm{Conv}(\cdot)$ and $\textrm{Att}(\cdot)$ refer to the convolutional and SHCA blocks. Next, we sum multi-modal feature with $\mathbf{R}_m$ to produce features with highly correlated information:
\begin{align}
\mathbf{F}_{av}^{k} &= \textrm{LN}(\mathbf{F}_{av}^{k-1} + \mathbf{R}_{a}^{k-1} + \mathbf{R}_{v}^{k-1}),
\label{eq19}
\end{align}
where $k \in \{2, ..., N_c\}$ is the unit index. The parameters of each ER Layer are shared to facilitate evolutionary refinements. Finally, the outputs $\{\mathbf{F}_{a_i}^2, \mathbf{F}_{v_i}^2\}_{i=1}^{N_c}$ of all the units are preserved as features at multi-semantic scales, while $\mathbf{F}_{av}^{N_c}$ will be utilized for the following feedback enhancement.

We then present the HAFE Layer to hierarchically aggregate preserved features and promote correlated relationships modeling in reverse. Since features across units have distinct semantic scales, simply using pooling integrates the hierarchical representations inadequately. We thus first stack $\{\mathbf{F}_{m_i}^2\}_{i=1}^{N_c}$ along $N_c$ to merge the scale-aware features, then deploy the unit-level MHSA followed by FFNs to provide aggregatively contextual integrations, which further consider the intra-modal correspondences, \textit{i.e.}, 
\begin{align}
\mathbf{\gamma}_m &= \textrm{MHSA}(\textrm{LN}(\mathbf{F}_m^s) + \mathbf{F}_m^s,
\label{eq20}
\end{align}
where $\mathbf{F}_m^s = \textrm{Stack}(\mathbf{F}_{m_1}^2, ..., \mathbf{F}_{m_{N_c}}^2)$. To select the most useful representations, we first apply the linear projection and sigmoid function to dynamically assign weights across different granularities. The weighted summation is then conducted to output the compatibly integrated features, \textit{i.e.},
\begin{align}
\mathbf{F}_{m}^3 &= \sum_{l=1}^{N_c} \ \sigma\left(\mathbf{W}_{sf} \cdot \gamma_{m}^{l} + \mathbf{b}_{sf}\right) \cdot \gamma_{m}^{l},
\label{eq14}
\end{align}

Afterwards, we deploy MHCA followed by FFNs to facilitate the complementary correlated information learning with $\mathbf{F}_{av}^{N_c}$ under a feedback manner, which can be given as:
\begin{align}
\mathbf{F}_{m}^4 &= \textrm{MHCA}(\textrm{LN}(\mathbf{F}_{m}^3), \textrm{LN}(\mathbf{F}_{av}^{N_c})) + \mathbf{F}_{m}^3.
\label{eq16}
\end{align}

\begin{table}[t!]
\centering
\caption{The statistics of data utilized for three training stages. AC: Acquisition Condition. Mix: Wild \& Lab Environments. CEA: Categorical Emotion Analysis. DEA: Dimensional Emotion Analysis. MER: Micro-Expression Recognition.}
\setlength{\abovecaptionskip}{0.31em}
\setlength{\arrayrulewidth}{0.40pt}
\renewcommand{\arraystretch}{0.95}
\resizebox{0.8\linewidth}{!}{
\begin{tabular}{c|ccccc}
\toprule[0.40pt]
Stage & Task & Dataset & \#Emos & Num & AC \\
\midrule[0.40pt]
Pre-training & \textbf{--} & Unlabeled Hybrid & \textbf{--} & 1,360,531 & Mix \\ \midrule[0.40pt]
Supervised & \textbf{--} & CEA Labeled Hybrid & 13 & 31,218 & Mix \\
Post-pre-training & \textbf{--} & MER Labeled Hybrid & 3 & 1,007 & Lab \\ \midrule[0.40pt]
 & \multirow{2}{*}{CEA} & \multirow{2}{*}{MAFW~\cite{63_liu2022mafw}} & 11 & 9,172 & Wild \\
 &  & & 43 & 8,996 & Wild \\
 & CEA & DFEW~\cite{9_jiang2020dfew} & 7 & 11,697 & Wild \\
 & CEA & MER-MULTI~\cite{64_lian2023mer} & 6 & 3,784 & Wild \\
 & CEA & MER24-T\&V~\cite{56_lian2024mer} & 6 & 5030 & Wild \\
 & CEA & IEMOCAP~\cite{65_busso2008iemocap} & 4 & 5,531 & Lab \\
 & \multirow{2}{*}{CEA} & \multirow{2}{*}{CREMA-D~\cite{66_cao2014crema}} & 6 & 7,442 & Lab \\
Targeted &  & & 4 & 4,896 & Lab \\
Fine-tuning & CEA & RAVDESS~\cite{67_livingstone2018ryerson} & 8 & 1,440 & Lab \\
 & CEA & MSP-IMPROV~\cite{68_busso2016msp} & 4 & 7,798 & Lab \\
 & DEA & Werewolf-XL~\cite{62_zhang2021werewolf} & 3 & 14,632 & Lab \\
 & DEA & AVCAffe~\cite{61_sarkar2023avcaffe} & 2 & 58,112 & Wild \\
 & MER & SAMM~\cite{72_davison2016samm} & 3 & 133 & Lab \\
 & MER & CASME II~\cite{70_yan2014casme} & 3 & 145 & Lab \\
 & MER & SMIC~\cite{73_li2013spontaneous} & 3 & 164 & Lab \\
 & MER & CAS(ME)$^3$~\cite{71_li2022cas} & 3 & 943 & Lab \\
 & MER & MMEW~\cite{69_ben2021video} & 3 & 300 & Lab \\
\bottomrule[0.40pt]
\end{tabular}
}
\label{tab:dataset_summary}
\vspace{-1.1em}
\end{table}

In the final processing step, we apply a pooling operation along the token dimension to reshape the features into $\mathbf{F}_{m}^4 \in \mathbb{R}^{C}$. These reshaped features are then concatenated and passed through a designated linear layer to generate the final output, $\mathbf{F}_f$, for the fully tuned model. For downstream tasks, the model utilizes a cross-entropy loss for classification and a mean square error loss for regression. It is important to note that the primary distinction between this fine-tuning stage and the pre-training phase is the number of visible tokens in the input features.

\subsection{Dual Scaling and Progressive Training}
\label{3.5}
\noindent \textbf{Model Scaling.}
The model capacity is the foremost force in improving performance. Following the scaling behaviors of~\cite{1_sun2024hicmae,7_tong2022videomae}, we scale the capacity of AVF-MAE++ by constructing uni-modal encoders of varying dimensions, attention heads, and depths, leading to three versions (\textit{i.e.}, Base, Large, and Huge), which are detailed in the supplementary material. The stacked number of our IAV-CL Module remains constant. Besides, we adhere to~\cite{1_sun2024hicmae,4_wang2023videomae} by using lightweight vanilla ViT~\cite{24_dosovitskiy2020image} as decoders, while keeping the decoder capacity consistent across different model versions.

\noindent \textbf{Data Scaling.}
We construct an unlabeled hybrid cross-linguistic facial video dataset to better support audio-visual MAE pre-training, originating from CN-Celeb series~\cite{55_fan2020cn}, MER2024~\cite{56_lian2024mer}, VoxCeleb2-dev~\cite{57_chung2018voxceleb2}, AV-Speech~\cite{58_ephrat2018looking}, and CelebV-HQ~\cite{59_zhu2022celebv}, as illustrated in Tab.~\ref{tab:dataset_summary}. After collection, we filter and crop videos using the pre-processing pipeline from~\cite{100_chen2023cn} to reduce redundancy, resulting in a hybrid pre-training dataset with 1.36M clips.\textbf{ \textit{To our best knowledge, this is the largest dataset utilized for AVFA self-supervised pre-training}}. More details are illustrated in Sec.~\ref{sec:data} below.

\noindent \textbf{Progressive Adaptation Training.}
In contrast to the settings in~\cite{7_tong2022videomae,4_wang2023videomae}, the AVFA benchmark presents significant adaptation and overfitting challenges. These issues stem from two primary factors: a distribution gap between the pre-training and fine-tuning data, and the limited size of the fine-tuning dataset. Together, these problems hinder the ability of pre-trained models to reach their full potential. To address this, we introduce the progressive semantics injection (PSI) strategy, which draws inspiration from~\cite{4_wang2023videomae,23_bao2021beit}. This method facilitates a gradual adaptation to downstream tasks by incorporating supervised semantic signals from multiple sources, thereby establishing a three-stage training pipeline. Concretely, we first conduct self-supervised pre-training on the unlabeled hybrid dataset. We then perform supervised post-pre-training on the labeled hybrid datasets to inject downstream semantics into pre-trained models. As displayed in Tab.~\ref{tab:dataset_summary}, the labeled hybrid datasets are built by merging datasets for different downstream tasks and aligning their label semantics. Finally, we fine-tune models on targeted datasets to transfer the general semantics to task-specific knowledge.

\section{Datasets}
\label{sec:data}

In this section, we provide details regarding the construction of pre-training datasets and intermediate tuning datasets, together with the specifics of targeted downstream datasets.

\subsection{Unlabeled Hybrid for Pre-training}

Our unlabeled hybrid dataset comprises a combination of unlabeled facial videos sourced from the CN-Celeb series~\cite{55_fan2020cn,100_chen2023cn}, MER2024~\cite{56_lian2024mer}, VoxCeleb2~\cite{57_chung2018voxceleb2}, AV-Speech~\cite{58_ephrat2018looking}, along with CelebV-HQ~\cite{59_zhu2022celebv}. The detailed components of the unlabeled hybrid dataset are presented in Tab.\ref{tab:unlabeledhybrid}. Subsequently, we briefly specify the handling of each dataset.

\noindent \textbf{VoxCeleb2}. VoxCeleb2~\cite{57_chung2018voxceleb2} contains over 1 million clips featuring more than 6,000 celebrities, derived from approximately 150,000 interview videos on YouTube. VoxCeleb2 is split into a development (\textit{dev}) set and a test set. Here, we initially select portions of the clips from the \textit{dev} set, followed by pre-processing, resulting in 515K processed samples.

\noindent \textbf{AV-Speech}. AV-Speech~\cite{58_ephrat2018looking} provides video clips sourced from educational lectures (\textit{e.g.}, TED Talks) and instructional videos available on the YouTube platform. In total, approximately 4,700 hours of video content are included in this dataset, featuring around 150,000 unique speakers and exhibiting diverse characteristics in terms of individuals, languages and facial orientations. For our research, we extract portions from the complete collection and apply pre-processing procedures to these chosen clips, obtaining 371K processed samples.

\noindent \textbf{MER2024}. MER2024~\cite{56_lian2024mer} represents an enhanced iteration of MER2023, comprising 115,595 video clips sourced from the Internet. Through pre-processing procedures, we derive 85K samples for the pre-training.

\noindent \textbf{CN-Celeb series}. CN-Celeb series dataset constitutes a large-scale continuous visual-speech benchmark in Mandarin Chinese, comprising short clips gathered from TV news and Internet speech shows. A subset of data is randomly selected and we process the non-standard clips. Ultimately, we employ 370K clips for the pre-training stage.

\noindent \textbf{CelebV-HQ}. CelebV-HQ~\cite{59_zhu2022celebv} comprises 35,666 video clips featuring minimum resolution of 512 $\times$ 512, encompassing 15,653 identities. Upon completion of pre-processing, there are 18K clips available for use in our AVFA pre-training.

To reduce computational cost and improve efficiency, we propose a multi-stage, multi-granularity data processing pipeline, which can be divided into five stages: \textbf{(1)} Considering the requirements of AVAF, we first perform scene change detection and cutting on the videos. Using LibAV, we obtain segments without scene changes and remove overly short or incomplete clips. \textbf{(2)} We then apply face detection to filter out low-quality samples with multiple faces, no faces, or severely distorted faces. To balance efficiency and accuracy, we sample several frames from each clip and use dlib for face detection. \textbf{(3)} We use librosa to detect silent frames and further segment each video for more precise audio-video alignment, aiding subsequent training. \textbf{(4)} To ensure continuity and stability in facial sequences, we use RetinaFace to extract face regions and calculate their center points. A Savitzky-Golay filter is applied to smooth the trajectory, eliminating noise and fluctuations. A dynamic window is then created based on the maximum face size across all frames to extract stable face images. 
\textbf{(5)} Finally, we employ the SyncNet ~\cite{Syncnet} model to perform audio-visual synchronization detection. By jointly modeling lip motion features from the video and speech features from the audio, SyncNet evaluates their temporal alignment and effectively identifies potential misalignments or semantic inconsistencies. Clips with mismatched lip movements and speech rhythms are discarded as unsynchronized, while those with blur, watermarks, or other visual artifacts are also filtered out.
Through the above preprocessing pipeline, we construct a large-scale pre-training dataset containing approximately 1.36M high-quality audio-visual clips, thereby providing sufficient and diverse samples for large-scale self-supervised cross-modal pre-training and facilitating representation learning with temporal alignment and semantic consistency.

\begin{table}[t]
\centering
\caption{The detailed components of our unlabeled hybrid dataset. We build this unlabeled dataset by collecting clips from multiple sources to better support AVF-MAE++ pre-training.}
\setlength{\abovecaptionskip}{0.3em} 
\setlength{\arrayrulewidth}{0.40pt}  
\renewcommand\arraystretch{1.1}      
\begin{tabular}{l|cc}
\toprule
Dataset & Size & Source \\
\hline
VoxCeleb2~\cite{57_chung2018voxceleb2} & 515K & YouTube \\
AV-Speech~\cite{58_ephrat2018looking} & 371K & YouTube \\
MER2024~\cite{56_lian2024mer} & 85K & Open-Media \\
CN-Celeb series~\cite{55_fan2020cn,100_chen2023cn} & 370K & Open-Media \\
CelebV-HQ~\cite{59_zhu2022celebv} & 18K & Open-Media \\
\hline
\textbf{Unlabeled Hybrid} & 1.36M & Multi-Source \\
\bottomrule

\end{tabular}
\label{tab:unlabeledhybrid}
\end{table}

\subsection{Labeled Hybrid for Intermediate Tuning}

For the post-pre-training stage of our AVF-MAE++, we develop labeled hybrid datasets targeting downstream tasks, thereby establishing the progressive training pipeline through unifying various downstream targeted datasets.

Using the CEA task as an illustration, we initially perform label semantic alignment across the downstream targeted datasets, as shown in Tab.\ref{CEA-mapping} above. Subsequently, label remapping of the downstream datasets is conducted to generate annotations for the labeled hybrid dataset. Given that label semantic alignment cannot be applied to both datasets in the DEA task, labeled hybrid datasets are constructed exclusively for the CEA and MER tasks. Regarding the MER task, the five selected datasets all conform to the widely-adopted three-emotion paradigm, ensuring that both targeted datasets and the labeled hybrid dataset maintain consistency with this paradigm. It should be noted that the MSP-IMPROV~\cite{68_busso2016msp} dataset is excluded from the CEA labeled hybrid dataset construction owing to its distinctive annotation characteristics.

\begin{table*}[t]
\centering
\caption{The emotional labels of the built labeled hybrid and related downstream targeted datasets. Overall, there are 13 emotional labels in the labeled hybrid dataset for the CEA task.}
\setlength{\abovecaptionskip}{0.3em} 
\setlength{\arrayrulewidth}{0.40pt}  
\renewcommand\arraystretch{1.1}      
\resizebox{\linewidth}{!}{%
\begin{tabular}{l|c}
\toprule
Dataset & Emotional Labels  \\
\hline
\multirow{2}{*}{MAFW~\cite{63_liu2022mafw}} & Anger, Disgust, Fear, Happiness, Neutral, Sadness  \\
& Surprise, Contempt, Anxiety, Helplessness, Disappointment \\
DFEW~\cite{9_jiang2020dfew} & Happy, Sad, Neutral, Angry, Surprise, Disgust, Fear  \\
MER-MULTI~\cite{64_lian2023mer} \& MER24-T\&V~\cite{56_lian2024mer} & Worried, Happy, Neutral, Angry, Surprise, Sad  \\
IEMOCAP~\cite{65_busso2008iemocap} & Anger, Happy, Neutral, Sad  \\
CREMA-D~\cite{66_cao2014crema} & Anger, Disgust, Fear, Happy, Neutral, Sadness  \\
RAVDESS~\cite{67_livingstone2018ryerson} & Neutral, Calm, Happy, Sad, Angry, Fearful, Disgust, Surprised  \\
\hline
\multirow{2}{*}{\textbf{Labeled Hybrid}} & Anger, Disgust, Fear, Happy, Neutral, Sadness, Surprise \\ & Worried, Clam, Contempt, Anxiety, Helplessness, Disappointment \\
\bottomrule
\end{tabular}
}
\label{CEA-mapping}
\vspace{-1em}
\end{table*}

\subsection{Targeted Fine-tuning}

\noindent \textbf{MAFW}~\cite{63_liu2022mafw} represents a multi-modal compound in-the-wild affective dataset. It comprises 10,045 clips annotated with 11 common emotions. Each video clip is additionally paired with several textual sentences describing the subject's affective behaviors. The dataset offers an 11-class single-labeled set with 9,172 clips and a 43-class compound set with 8,996 clips. We adhere to the original paper by adopting a 5-fold cross-validation protocol for model performance evaluation.

\noindent \textbf{DFEW}~\cite{9_jiang2020dfew} encompasses 16,372 clips derived from movies. This dataset exhibits several challenging characteristics, such as extreme illumination and occlusion. We conduct 5-fold cross-validation on 11,697 single-labeled clips for evaluations to maintain consistency with previous works.

\noindent \textbf{MER-MULTI}~\cite{64_lian2023mer} offers 3,373 training clips originating from Chinese TV series and movies. We adhere to the original paper by conducting 5-fold cross-validation on 3,373 clips for hyper-parameter tuning and assessing performance on a held-out test set with 411 clips.

\noindent \textbf{MER24-T\&V}~\cite{56_lian2024mer} (MER2024-Train~\&~Val) is expanded by merging the labeled samples of MER2023 Challenge~\cite{64_lian2023mer}. According to the original paper, we fine-tune models on the \textit{Train} set and assess performance on the \textit{Val} set.

\noindent \textbf{IEMOCAP}~\cite{65_busso2008iemocap} encompasses roughly 12 hours of videos from 10 subjects recorded across five sessions. According to common practice~\cite{1_sun2024hicmae}, we employ 5,531 samples of five emotional categories (\textit{i.e.}, Anger, Neutral, Happiness, Excitement, Sadness), then combine Excitement into Happiness to establish a four-emotion form. Additionally, we conduct 5-fold cross-validation in a session-independent manner.

\noindent \textbf{CREMA-D}~\cite{66_cao2014crema} represents a high-quality dataset for analyzing multi-modal patterns of acted emotions. It comprises 7,442 clips recorded by 91 actors. Since no official split exists, we adhere to previous works~\cite{1_sun2024hicmae,95_tran2023saaml} by conducting 5-fold cross-validation in a \textit{subject-independent} manner. We also perform experiments on a subset of four emotions (\textit{i.e.}, Happiness, Sadness, Anger, Neutral) and only report the performance of the final fold under this setup.

\noindent \textbf{RAVDESS}~\cite{67_livingstone2018ryerson} comprises emotional speech and songs, featuring 2,880 clips with 24 professional actors. According to ~\cite{1_sun2024hicmae}, we exclusively use the speech portion and employ a \textit{subject-independent} 6-fold cross-validation protocol for performance evaluation~\cite{1_sun2024hicmae,2_sun2023mae}.

\noindent \textbf{MSP-IMPROV}~\cite{68_busso2016msp} represents an acted audio-visual corpus for exploring emotional behaviors during conversational dyadic interactions. MSP-IMPROV~\cite{68_busso2016msp} encompasses 8,438 clips recorded across six sessions from 12 actors. According to~\cite{1_sun2024hicmae,95_tran2023saaml}, we exclusively use samples of four emotions and perform 6-fold cross-validation in a \textit{session-independent} manner.

\noindent \textbf{AVCAffe}~\cite{61_sarkar2023avcaffe} represents a large-scale audio-visual affect dataset simulating remote work scenarios, featuring almost 108 hours of videos along with self-reported labels for cognitive load and affect (\textit{i.e.}, Arousal, Valence). Since arousal and valence scores are provided on a scale of 1-4, we adhere to ~\cite{61_sarkar2023avcaffe,1_sun2024hicmae} by formulating their predictions as a classification task. We employ the official split (86 subjects for training and 20 subjects for testing) for performance evaluation.

\noindent \textbf{Werewolf-XL}~\cite{62_zhang2021werewolf} represents a database for studying spontaneous emotions during competitive group interactions of Werewolf games. It encompasses roughly 15 hours of audio-visual recordings. To maintain consistency with previous works, we employ 14,632 samples with dimensional annotations (\textit{i.e.}, Arousal, Valence, Dominance) and perform \textit{subject-independent} 5-fold cross-validation for performance assessment.

\noindent \textbf{CASME II}~\cite{70_yan2014casme} encompasses videos of 24 subjects, totaling 145 samples. All samples are captured using lab cameras, with a frame rate of 200 FPS. After merging into three emotional categories, the numbers of Negative, Positive, and Surprise are 88, 32, and 25, respectively.

\noindent \textbf{SAMM}~\cite{72_davison2016samm} offers various facial expression data, encompassing action unit coding as well as indices for micro-expression onset, offset, and apex. All videos have a resolution of 2040 $\times$ 1088 pixels with a frame rate of 200 FPS. After grouping videos into three emotions, SAMM has 92 Negative, 26 Positive, and 15 Surprise samples.

\noindent \textbf{SMIC}~\cite{73_li2013spontaneous} comprises video data from 16 subjects, totaling 164 samples. All samples are recorded using a lab camera with a frame rate of 100 FPS. The original frame size of each sample is 640 $\times$ 480 pixels. The sample numbers of Negative, Positive, and Surprise emotions are 70, 51, and 43, respectively.

\noindent \textbf{CAS(ME)$^3$}~\cite{71_li2022cas} is characterized by its inclusion of multi-source information. In this work, we exclusively select the {\em PartA} for model performance evaluation, which comprises data from 100 subjects, totaling 943 samples. The samples are captured using a lab camera, and have an original resolution of 1280 $\times$ 720 pixels. The overall numbers of Negative, Positive, and Surprise emotional samples are 508, 64, and 201.

\noindent \textbf{MMEW}~\cite{69_ben2021video} encompasses both macro- and micro-expressions. It comprises 300 MEs and 900 macro-expression samples with a large resolution (\textit{i.e.}, 1920 $\times$ 1080) at 90 FPS. According to previous works
~\cite{74_nguyen2023micron,81_zhou2022feature}, we merge the seven emotions into three emotions, then assess performance.
\definecolor{cvprblue}{RGB}{0.21,0.49,0.74}

\section{Experiments}
\label{sec:exprs}

\subsection{Model Configurations}
We construct three distinct variants of AVF-MAE++, specifically Base: AVF-MAE++ (B), Large: AVF-MAE++ (L), and Huge: AVF-MAE++ (H), in order to comprehensively investigate the scaling characteristics of audio-visual MAE in AVFA applications. The key distinctions among these three model variants lie in the architectural configurations of their modality-specific encoders, as presented in Tab.~\ref{tab_configs} above.

\begin{table*}[t!]
\centering
\caption{The overall illustrations of configuration details about AVF-MAE++ across three different versions. Note that the index of hierarchical skipping connections is from 0 to (encoder depth~-~1).}
\setlength{\abovecaptionskip}{0.3em} 
\setlength{\arrayrulewidth}{0.40pt}  
\renewcommand{\arraystretch}{1.1}    
\resizebox{\linewidth}{!}{%
\begin{tabular}{l|ccc}
\toprule
Configurations  & AVF-MAE++ (B)         & AVF-MAE++ (L) & AVF-MAE++ (H)  \\
\hline
patch size                        & 16  & 16   & 16          \\
embedding dimensions (encoder)                        & 512  & 640   & 768          \\
number of attention heads (encoder)                          & 8 & 10   & 12          \\
encoder depth                          & 10  & 12   & 15          \\
embedding dimensions (decoder)                          & 384  & 512   & 640          \\
number of attention heads (decoder)                          & 6  & 8   & 8          \\
decoder depth                          & 4  & 4   & 4         \\
number of attention heads (fusion)                          & 8  & 10   & 12          \\
fusion depth                          & 2  & 2   & 2          \\
index of hierarchical skip connections~\cite{1_sun2024hicmae}                          & [3, 6, 9]  & [3, 7, 11]   & [4, 9, 14]          \\
\bottomrule
\end{tabular}
}
\label{tab_configs}
\end{table*}

\begin{table}[t!]
\centering
\caption{Pre-training settings. We only show the details of AVF-MAE++ (B) here for the example.}
\begin{tabular}{l|c}
\toprule
Configurations & Value \\
\hline
video encoder mask type & tube \\
video encoder mask ratio & 0.9 \\
audio encoder mask type & random \\
audio encoder mask ratio & 0.8125 \\
video input size & 3 $\times$ 16 $\times$ 160 $\times$ 160 \\
audio input size & 1 $\times$ 256 $\times$ 128 \\
video decoder mask type & running cell \\
video decoder mask ratio & 0.5 \\
audio decoder mask type & random \\
audio decoder mask ratio & 0.5 \\
optimizer & AdamW~\cite{101_loshchilov2017decoupled} \\
base learning rate & 1.5\textit{e}-4 \\
weight decay & 0.05 \\
target normalizations & Yes \\
loss weight factor & 0.0025 \\
Mel filterbank sequence length & 128 \\
audio augmentation & Yes \\
video augmentation & MultiScaleCrop \\
optimizer momentum & $\beta_1, \beta_2 = 0.9, 0.95$ \\
base batch size & 164 \\
contrastive temperature & 0.07 \\
video region size & (2, 5, 10) \\
audio region size & (4, 4) \\
repeated augmentation~\cite{102_hoffer2020augment} & No \\
learning rate schedule & cosine decay \\
frame difference optimization & Yes \\
warmup epoch & 20 \\
epoch & 200 \\
frame & 16 \\
sampling rate & 4 \\
audio sampling rate & 16000 \\
clip grading & None (B \& L), 0.7 (H) \\
\bottomrule
\end{tabular}
\label{tab:pretrain_set}
\end{table}

\subsection{Implementation Details} 
\label{sec:details}

We conduct pre-training of three AVF-MAE++ versions on the constructed unlabeled hybrid dataset utilizing a machine equipped with 8 $\times$ NVIDIA RTX A6000 GPUs. In addition to the computational efficiency improvements methods proposed in the main paper, we incorporate mix-precision training at the engineering level to accelerate pre-training processes. In accordance with~\cite{4_wang2023videomae,7_tong2022videomae,1_sun2024hicmae}, we implement FP-16 mixed precision for encoder training and FP-32 precision for decoder training to prevent potential precision overflow risk during model pre-training. The repeated augmentation for video data is not incorporated during pre-training. The learning rate undergoes linear scaling based on the total batch size (\textit{i.e.}, $\text{lr} = \text{base\_lr} \times  \text{batch\_size}~ / ~256$). The comprehensive pre-training settings are presented in Tab.~\ref{tab:pretrain_set}.

During the supervised post-pre-training stage, we perform fine-tuning of the pre-trained encoder on the labeled hybrid dataset for various downstream AVFA tasks. To effectively preserve the pre-training effects, we moderately increase the drop path rate and incorporate the repeated augmentation. Subsequently, we execute the specific fine-tuning to generate the targeted models for categorical emotion analysis (CEA), dimensional emotion analysis (DEA), and micro-expression recognition (MER) three tasks. In particular, we utilize nearly identical pipelines for the three downstream tasks, with the exception that we implement the MSE Loss and incorporate activation functions before the model head for the DEA task. The comprehensive information regarding the post pre-training and targeted fine-tuning settings of AVF-MAE++ is provided in Tab.~\ref{tab:post-ft-settings}. For simplicity, we exclude the parameter configurations that remain consistent with the pre-training phase in Tab.~\ref{tab:pretrain_set}.

\begin{table}[t!]
\centering
\caption{Post pre-training and targeted fine-tuning settings. Due to space constraints, we only show the detailed settings of AVF-MAE++ (B) on the CEA downstream task.}
\begin{tabular}{l|cc}
\toprule
Configurations & \makecell{Post \\ Pre-training} & \makecell{Targeted \\ Fine-tuning} \\
\hline
optimizer & \multicolumn{2}{c}{AdamW~\cite{101_loshchilov2017decoupled}} \\
base learning rate & 1\textit{e}-3 & 5\textit{e}-4 \\
weight decay & \multicolumn{2}{c}{0.05} \\
optimizer momentum & \multicolumn{2}{c}{$\beta_1, \beta_2 = 0.9, 0.999$} \\
inference segment & \multicolumn{2}{c}{2} \\
inference crop & \multicolumn{2}{c}{2} \\
learning rate schedule & \multicolumn{2}{c}{cosine decay} \\
warmup epoch & \multicolumn{2}{c}{5} \\
epoch & \multicolumn{2}{c}{100} \\
drop path & \multicolumn{2}{c}{0.1} \\
layer decay & \multicolumn{2}{c}{0.75} \\
base batch size & \multicolumn{2}{c}{32}  \\
tubelet size & \multicolumn{2}{c}{2} \\
color jitter factor & \multicolumn{2}{c}{0.4} \\
mix up~\cite{105_zhang2017mixup} & \multicolumn{2}{c}{0.8} \\
RandAug~\cite{103_cubuk2020randaugment} & \multicolumn{2}{c}{(0, 0.25)} \\
label smoothing~\cite{104_szegedy2016rethinking} & \multicolumn{2}{c}{0.1} \\
repeated augmentation~\cite{102_hoffer2020augment} & \multicolumn{2}{c}{2} \\
\bottomrule
\end{tabular}
\label{tab:post-ft-settings}
\end{table}

\subsection{Downstream AVFA Tasks}

To validate the effectiveness and broad applicability of AVF-MAE++, we perform comprehensive evaluations across diverse datasets covering three representative AVFA tasks, as summarized in Tab.~\ref{tab:dataset_summary}.

\noindent \textbf{Categorical Emotion Analysis (CEA).}~As one of the core tasks in AVFA, CEA focuses on classifying each clip to a specific emotion category. Following ~\cite{1_sun2024hicmae}, we conduct an in-depth study of this task to explore the scalability of audio-visual MAE models. Our evaluation involves ten datasets, with UAR, WAR, and WA-F1 adopted as performance metrics.

\noindent \textbf{Dimensional Emotion Analysis (DEA).}~DEA models emotions in a continuous space, making it more demanding than CEA due to its finer-grained annotation. In line with ~\cite{1_sun2024hicmae,3_sun2024svfap}, we adopt AVCAffe~\cite{61_sarkar2023avcaffe} and WereWolf-XL~\cite{62_zhang2021werewolf} to assess our framework. We do not construct a \textit{labeled hybrid} set for DEA, as both datasets already provide sufficient samples and he semantic alignment for continuous annotations cannot be explicitly applied. For evaluation, WA-F1 is used on ~\cite{61_sarkar2023avcaffe}, while ~\cite{62_zhang2021werewolf} is measured by PCC under A-V-D conditions.

\noindent \textbf{Micro-Expression Recognition (MER).}~MER focuses on detecting fleeting and subtle facial movements that reflect underlying emotions. The objective is to reliably identify these rapid changes in video sequences to gain a richer understanding of affective states. We employ five widely used benchmarks and report UF1 as the evaluation metric.

\begin{table*}[]
\centering
\caption{Performance comparisons of AVF-MAE++ with state-of-the-art CEA methods on MAFW (11-class). AN: Anger. DI: Disgust. FE: Fear. HA: Happiness. NE: Neutral. SA: Sadness. SU: Surprise. CO: Contempt. AX: Anxiety. HL: Helplessness. DS: Disappointment. UAR: Unweighted Average Recall. WAR: Weighted Average Recall. *: The pre-trained models is not deployed for initialization. \textbf{--}: Unaccessible Results. We highlight the best performance in \textbf{bold} and \underline{underline} the second performance.}
\setlength{\abovecaptionskip}{0.3em} 
\setlength{\arrayrulewidth}{0.40pt}  
\renewcommand\arraystretch{1.27}      
\resizebox{\textwidth}{!}{%
\begin{tabular}{lccccccccccccccccc}
\toprule
\multirow{2}{*}{Method}  & \multirow{2}{*}{Venue} & \multirow{2}{*}{SSL} & \multirow{2}{*}{Modality} & \multirow{2}{*}{\#Params (M)}    
& \multicolumn{11}{c}{Accuracy of Each Emotion (\%)}  & \multicolumn{2}{c}{Metrics (\%)} \\ 
\cmidrule(lr){6-16} \cmidrule(lr){17-18} 
&  & &  & &  AN    & DI    & FE    & HA    & NE    & SA    & SU    & CO   & AX    & HL   & DS   & UAR  & WAR  \\ 
\midrule

Wav2Vec2.0~\cite{96_baevski2020wav2vec} & NeurIPS'20 & \checkmark  &  A  &  95    & 59.01 &  9.39 & 26.08 & 31.47 & 32.04  & 46.52  & 9.91  & 1.69  & 12.23  & 3.05  & 6.04  & 21.59 & 29.69         \\
HuBERT~\cite{32_hsu2021hubert}     & TASLP'21      & \checkmark  &  A  &  95    & 54.97 & 15.49 & 31.20 & 28.64 & 36.88  & 58.39  & 12.52 & 2.54  & 12.55  & 5.34  & \underline{16.48} & 25.00 & 32.60        \\
WavLM-Plus~\cite{87_chen2022wavlm}   & J-STSP'22    & \checkmark  &  A  &  95    & 55.62 & 17.21 & 40.48 & 36.65 & 36.53  & 57.44  & 11.12 & 2.12  & 11.35  & \underline{9.54}  & 11.54 & 26.33 & 34.07        \\

\hline%

ResNet-18~\cite{90_he2016deep}  & CVPR'16     &  $\times$   &  V  &  11    & 45.02 & 9.25  & 22.51 & 70.69 & 35.94 & 52.25 & 39.04 & 0.00 & 6.67  & 0.00 & 0.00 & 25.58 & 36.65          \\

ViT~\cite{24_dosovitskiy2020image} & ICLR'21  &  $\times$   &  V  &  \textbf{--}    & 46.03 & 18.18 & 27.49 & 76.89 & 50.70 & 68.19 & 45.13 & 1.27 & 18.93 & 1.53 & 1.65 & 32.36 & 45.04          \\

S2D~\cite{111_chen2024static} & TAFFC'24 &  $\times$ &  V  &  9   & \textbf{--}   & \textbf{--} & \textbf{--} & \textbf{--} & \textbf{--} & \textbf{--} & \textbf{--} & \textbf{--} & \textbf{--} & \textbf{--} & \textbf{--} & 43.40 & 57.37 \\

C3D~\cite{107_tran2015learning}   & ICCV'15    &  $\times$   &  V  &  78   & 51.47 & 10.66 & 24.66 & 70.64 & 43.81 & 55.04 & 46.61 & 1.68 & 24.34 & 5.73 & 4.93 & 31.17 & 42.25          \\

ResNet-18+LSTM~\cite{63_liu2022mafw} & MM'22 &  $\times$   &  V  &  \textbf{--}    & 46.25 & 4.70  & 25.56 & 68.92 & 44.99 & 51.91 & 45.88 & 1.69 & 15.75 & 1.53 & 1.65 & 28.08           & 39.38          \\

FE-Adapter~\cite{112_gowda2024fe} & FG’24 &  $\times$ &  V  &  7   & \textbf{--}   & \textbf{--} & \textbf{--} & \textbf{--} & \textbf{--} & \textbf{--} & \textbf{--} & \textbf{--} & \textbf{--} & 2.84 & \textbf{--} & 39.41 & 55.02 \\

ViT+LSTM~\cite{63_liu2022mafw}    & MM'22   &  $\times$   &  V  &  \textbf{--}   & 42.42 & 14.58 & 35.69 & 76.25 & 54.48 & 68.87 & 41.01 & 0.00 & 24.40 & 0.00 & 1.65 & 32.67           & 45.56          \\

C3D+LSTM~\cite{63_liu2022mafw}    & MM'22   &  $\times$   &  V  &  \textbf{--}    & 54.91 & 0.47  & 9.00  & 73.43 & 41.39 & 64.92 & 58.43 & 0.00 & 24.62 & 0.00 & 0.00 & 29.75           & 43.76          \\

Former-DFER~\cite{92_zhao2021former} & MM'21 &  $\times$   &  V  &  18  & 58.23 & 11.45 & 31.29 & 75.06 & 43.07 & 63.81 & 46.02 & 0.42 & 26.22 & 2.88 & 2.25 & 32.79 & 45.31 \\

T-ESFL~\cite{63_liu2022mafw}      & MM'22  &  $\times$   &  V  &  \textbf{--}    & 62.70 & 2.51  & 29.90 & 83.82 & 61.16 & 67.98 & 48.50 & 0.00 & 9.52  & 0.00 & 0.00 & 33.28 & 48.18     \\
T-MEP~\cite{85_zhang2023transformer}  & TCSVT'23 &  $\times$   &  V  &  5   & 52.91 & 17.41 & 28.01 & 80.79 & 49.42 & 58.73 & 49.54 & 0.00 & 26.18 & 2.25 & 3.56 & 33.53 & 47.53 \\

DFER-CLIP~\cite{26_zhao2023prompting} & BMVC’23 &  \checkmark &  V  &  153   & \textbf{--}   & \textbf{--} & \textbf{--} & \textbf{--} & \textbf{--} & \textbf{--} & \textbf{--} & \textbf{--} & \textbf{--} & \textbf{--} & \textbf{--} & 39.89 & 52.55 \\

SVFAP~\cite{3_sun2024svfap}   & TAFFC'24    & \checkmark  &  V   &  78  & 64.60 & 25.20 & 35.68 & 82.77 & 57.12 & 70.41 & 58.58 & 8.05 & 32.42 & 8.40 & 9.89 & 41.19 & 54.28        \\

EmoCLIP~\cite{113_foteinopoulou2024emoclip}  & FG’24 &  \checkmark &  V  &  \textbf{--}   & \textbf{--}   & \textbf{--} & \textbf{--} & \textbf{--} & \textbf{--} & \textbf{--} & \textbf{--} & \textbf{--} & \textbf{--} & \textbf{--} & \textbf{--} & 34.24 & 41.46 \\

MAE-DFER~\cite{2_sun2023mae}     & MM'23   & \checkmark    &  V   &  85  & 67.77 & 25.35 & 34.88 & 77.13 & 58.26 & 71.09 & 57.46 & \textbf{8.90} & 33.08 & \textbf{11.83} & 12.09 & 41.62 & 54.31 \\
UniLearn~\cite{84_chen2024unilearn} & arXiv'24 &  \checkmark &  V  &  101   & \textbf{--}   & \textbf{--} & \textbf{--} & \textbf{--} & \textbf{--} & \textbf{--} & \textbf{--} & \textbf{--} & \textbf{--} & \textbf{--} & \textbf{--} & 43.72 & 58.44 \\
A$^3$lign-DFER~\cite{114_tao20243} & arXiv'24 &  \checkmark &  V  &  \textbf{--}   & \textbf{--}   & \textbf{--} & \textbf{--} & \textbf{--} & \textbf{--} & \textbf{--} & \textbf{--} & \textbf{--} & \textbf{--} & \textbf{--} & \textbf{--} & 42.07 & 53.24\\

\hline%

ResNet-18+LSTM~\cite{63_liu2022mafw} & MM'22 & $\times$ &  A+V  &  \textbf{--}       & 54.47 & 11.89 & 7.07 & 82.73 & 54.85 & 55.06 & 39.35 & 0.00 & 15.99 & 0.39 & 0.00  & 29.26  & 42.69 \\
C3D+LSTM~\cite{63_liu2022mafw}      & MM'22         & $\times$ &  A+V  &  \textbf{--}       & 62.47 & 3.17  & 15.74 & 77.30 & 42.20 & 65.30 & 42.67 & 0.00 & 19.14 & 0.00 & 0.00  & 30.47  & 44.15 \\
AMH~\cite{108_yoon2020attentive}      & ICASSP'20       & $\times$ &  A+V  &  \textbf{--}       & 51.73 & 18.68 & 28.13 & 79.14 & 52.55 & 52.26 & 46.29 & 0.26 & 29.62 & 1.74 & 2.39 & 32.98 & 48.83 \\
T-ESFL~\cite{63_liu2022mafw}       & MM'22          & $\times$ &  A+V  &  \textbf{--}       & 60.73 & 1.26  & 21.40 & 80.31 & 58.24 & 75.31 & 53.23 & 0.00 & 14.93 & 0.00 & 0.00 & 33.35 & 48.70 \\
T-MEP*~\cite{85_zhang2023transformer}   & TCSVT'23    & $\times$ &  A+V  &  61      & 54.98 & 22.11 & 32.23 & 82.79 & 50.90 & 62.50 & 49.93 & 0.87 & 29.27 & 8.09 & 6.70 & 36.40 & 48.17 \\
T-MEP~\cite{85_zhang2023transformer}    & TCSVT'23    & $\times$ &  A+V  &  61      & 57.04 & 24.85 & 36.09 & 78.96 & 50.83 & 61.85 & 51.28 & 1.29 & \textbf{38.47} & 6.46 & 1.70 & 37.17 & 51.15 \\
MMA-DFER~\cite{25_chumachenko2024mma} & CVPR'24 &  \checkmark &  A+V  &  \textbf{--}   & \textbf{--}   & \textbf{--} & \textbf{--} & \textbf{--} & \textbf{--} & \textbf{--} & \textbf{--} & \textbf{--} & \textbf{--} & \textbf{--} & \textbf{--} & 44.25 & 58.45 \\
HiCMAE-T~\cite{1_sun2024hicmae}   & IF'24    & \checkmark &  A+V  &  20       & 67.72 & 24.73 & 34.56 & 75.81 & 55.63 & 73.74 & 56.45 & 2.97 & 29.69 & 6.87 & 13.74 & 40.17 & 53.41 \\
HiCMAE-S~\cite{1_sun2024hicmae}   & IF'24     & \checkmark &  A+V  &  46       & 67.94 & 26.13 & 36.00 & 75.00 & 56.51 & 73.33 & 58.41 & \underline{8.47} & 34.39 & 7.25 & 14.84 & 41.66 & 54.45 \\
HiCMAE-B~\cite{1_sun2024hicmae}    & IF'24   & \checkmark &  A+V  &  81       & 69.24 & 29.73 & 34.72 & 78.32 & 59.15 & \underline{77.69} & 60.65 & 6.78 & 31.11 & 8.02 & 13.74 & 42.65 & 56.17 \\

\rowcolor{cvprblue!20} 
\textbf{AVF-MAE++ (B)} & \textbf{--}   & \checkmark &  A+V  &  169       & \textbf{76.14} & 22.55 & \underline{44.32} & \underline{84.79} & 59.16 & 76.60 & 60.46 & 1.69 & 29.91 & 8.37 & 12.13 & 43.10 & 57.50 \\
\rowcolor{cvprblue!20}
\textbf{AVF-MAE++ (L)} & \textbf{--}   & \checkmark &  A+V  &  303       & \underline{72.25} & \underline{32.40} & \textbf{46.56} & 81.54 & \underline{63.93} & \textbf{78.50} & 61.95 & 4.68 & 31.44 & 5.34 & \textbf{20.33} & \underline{45.36} & \underline{59.13} \\
\rowcolor{cvprblue!20}
\textbf{AVF-MAE++ (H)} & \textbf{--}   & \checkmark &  A+V  &  521       & 71.89 & \textbf{38.19} & 40.80 & 84.25 & \textbf{68.49} & 76.40 & \textbf{64.66} & 3.83 & \underline{36.24} & 8.02 & 13.80 & \textbf{46.05} & \textbf{60.24} \\

\hline%
\gray{FineCLIPER~\cite{82_chen2024finecliper}} & \gray{MM'24} &  \gray{\checkmark} &  \gray{T+V}  &  \gray{20}   & \gray{\textbf{--}}   & \gray{\textbf{--}} & \gray{\textbf{--}} & \gray{\textbf{--}} & \gray{\textbf{--}} & \gray{\textbf{--}} & \gray{\textbf{--}} & \gray{\textbf{--}} & \gray{\textbf{--}} & \gray{\textbf{--}} & \gray{\textbf{--}} & \gray{45.01} & \gray{56.91} \\

\gray{ResNet18+MDRE~\cite{109_yoon2018multimodal}} & \gray{SLT'18}  & \gray{$\times$} & \gray{A+T+V} &  \gray{\textbf{--}}       & \gray{45.59}  & \gray{9.35}  & \gray{24.30} & \gray{76.31} & \gray{51.10} & \gray{74.87} & \gray{28.82} & \gray{2.08} & \gray{30.99} & \gray{0.00} & \gray{0.00} & \gray{31.22} & \gray{48.33} \\

\gray{AMH~\cite{108_yoon2020attentive}} & \gray{ICASSP'20} & \gray{$\times$} & \gray{A+T+V} & \gray{\textbf{--}}       & \gray{54.91}  & \gray{19.41} & \gray{30.01} & \gray{82.79} & \gray{51.42} & \gray{60.73} & \gray{51.54} & \gray{0.00} & \gray{28.18} & \gray{0.00} & \gray{0.00} & \gray{34.45} & \gray{49.87} \\

\gray{Rajan et al.~\cite{110_rajan2022cross}} & \gray{ICASSP'22}   & \gray{$\times$} & \gray{A+T+V} & \gray{\textbf{--}}      & \gray{56.10}  & \gray{9.96}  & \gray{41.58} & \gray{84.13} & \gray{60.39} & \gray{63.95} & \gray{44.59} & \gray{0.00} & \gray{24.26} & \gray{2.69} & \gray{1.76} & \gray{35.40} & \gray{48.78} \\

\gray{T-ESFL~\cite{63_liu2022mafw}}    & \gray{MM'22}     & \gray{$\times$} & \gray{A+T+V} & \gray{\textbf{--}}       & \gray{61.89}  & \gray{1.10}  & \gray{7.69}  & \gray{\textbf{85.90}} & \gray{\textbf{--}}    & \gray{71.87} & \gray{\underline{62.17}} & \gray{0.00} & \gray{36.00} & \gray{0.00} & \gray{0.00} & \gray{31.00} & \gray{50.29} \\

\gray{T-MEP*~\cite{85_zhang2023transformer}}    & \gray{TCSVT'23}     & \gray{$\times$} & \gray{A+T+V} & \gray{111}  & \gray{53.03}  & \gray{19.32} & \gray{40.65} & \gray{79.94} & \gray{55.89} & \gray{74.17} & \gray{53.48} & \gray{2.15} & \gray{26.61} & \gray{1.15} & \gray{5.10} & \gray{37.41} & \gray{50.96} \\

\gray{T-MEP~\cite{85_zhang2023transformer}}    & \gray{TCSVT'23}      & \gray{$\times$} & \gray{A+T+V} & \gray{111}  & \gray{56.95}  & \gray{18.19} & \gray{42.89} & \gray{81.62} & \gray{60.14} & \gray{71.60} & \gray{58.22} & \gray{3.21} & \gray{30.53} & \gray{2.27} & \gray{7.51} & \gray{39.37} & \gray{52.85} \\

\hline%

\end{tabular}
}
\label{tab_mafw_sota_single}
\end{table*}

\begin{table}[htbp]
\centering
\caption{The Comparative results of AVF-MAE++ with state-of-the-art methods on MAFW (43-class). Macro-F1: macro-averaged F1-score.}
\setlength{\abovecaptionskip}{0.3em} 
\setlength{\arrayrulewidth}{0.40pt}  
\renewcommand\arraystretch{1.14}     
\resizebox{0.90\linewidth}{!}{
\begin{tabular}{lccccccccccccccc}
\toprule
Method &     SSL &     Modality &  UAR   & WAR   & Macro-F1    \\ 
\midrule

Wav2Vec2.0~\cite{96_baevski2020wav2vec} (NeurIPS'20) & \checkmark  &   A   &  5.27  & 20.38 & \textbf{--}       \\
HuBERT~\cite{32_hsu2021hubert} (TASLP'21)      & \checkmark  &   A   &  5.36  & 20.70 & \textbf{--}       \\
WavLM-Plus~\cite{87_chen2022wavlm} (J-STSP'22)        & \checkmark  &   A   &  5.51  & 21.09 & \textbf{--}      \\

\hline

ResNet-18~\cite{90_he2016deep} (CVPR'16)         & $\times$   &   V   &  6.18  & 23.83 & 4.89  \\
ViT~\cite{24_dosovitskiy2020image} (ICLR'21)     & $\times$   &   V   &  8.62  & 31.76 & 7.46  \\    
C3D~\cite{107_tran2015learning} (ICCV'15)         & $\times$   &   V   &  9.51  & 28.12 & 6.73  \\    
ResNet-18+LSTM~\cite{63_liu2022mafw} (MM'22)   & $\times$   &   V   &  6.93  & 26.60 & 5.56  \\
ViT+LSTM~\cite{63_liu2022mafw} (MM'22)        & $\times$   &   V   &  8.72  & 32.24 & 7.59  \\
C3D+LSTM~\cite{63_liu2022mafw} (MM'22)        & $\times$   &   V   &  7.34  & 28.19 & 5.67  \\
T-ESFL~\cite{63_liu2022mafw} (MM'22)        & $\times$   &   V   &  9.15  & 34.35 & 7.18  \\
Former-DFER~\cite{92_zhao2021former} (MM'21)   & $\times$         &   V   & 10.21  & 32.07 &  \textbf{--}       \\
T-MEP~\cite{85_zhang2023transformer} (TCSVT'23)  & $\times$    &   V   &  9.50  & 31.54 &  \textbf{--}      \\

\hline

ResNet-18+LSTM~\cite{63_liu2022mafw} (MM'22)            & $\times$  &  A+V  &  7.85  & 31.03 & 5.95  \\ 
C3D+LSTM~\cite{63_liu2022mafw} (MM'22)            & $\times$  &  A+V  &  7.45  & 29.88 & 5.76  \\
T-ESFL~\cite{63_liu2022mafw} (MM'22)            & $\times$  &  A+V  &  9.93  & 34.67 & 8.44  \\
T-MEP*~\cite{85_zhang2023transformer} (TCSVT'23)   & $\times$  &  A+V  & 11.51  & 34.11 &  \textbf{--}       \\
T-MEP~\cite{85_zhang2023transformer} (TCSVT'23)  & $\times$  &  A+V  & 13.22  & 36.58 &  \textbf{--}       \\
HiCMAE-T~\cite{1_sun2024hicmae} (IF'24)        & \checkmark &  A+V  & 12.07  & 34.84 & 10.01    \\
HiCMAE-S~\cite{1_sun2024hicmae} (IF'24)        & \checkmark &  A+V  & 13.47  & 36.29 & 11.53     \\
HiCMAE-B~\cite{1_sun2024hicmae} (IF'24)        & \checkmark &  A+V  & 13.29  & 37.36 & 12.16    \\
\rowcolor{cvprblue!20}
\textbf{AVF-MAE++ (B)}        & \checkmark &  A+V  & 15.42  & 43.41 & 14.28    \\
\rowcolor{cvprblue!20}
\textbf{AVF-MAE++ (L)}        & \checkmark &  A+V  & \underline{15.59}  & \textbf{43.93} & \underline{14.52}     \\
\rowcolor{cvprblue!20}
\textbf{AVF-MAE++ (H)}        & \checkmark &  A+V  & \textbf{17.25}  & \underline{43.83} & \textbf{15.25}    \\

\hline

\gray{ResNet-18+MDRE~\cite{109_yoon2018multimodal} (SLT'18)}  & \gray{$\times$}    & \gray{A+V+T} & \gray{9.02}  & \gray{33.64} & \gray{\textbf{--}}       \\
\gray{AMH~\cite{108_yoon2020attentive} (ICASSP'20)}              & \gray{$\times$} & \gray{A+V+T} & \gray{10.24} & \gray{35.35} & \gray{\textbf{--}}      \\
\gray{Rajan et al.~\cite{110_rajan2022cross} (ICASSP'22)}      & \gray{$\times$} & \gray{A+V+T} & \gray{11.09} & \gray{35.33} & \gray{\textbf{--}}      \\
\gray{T-ESFL~\cite{63_liu2022mafw} (MM'22)}            & \gray{$\times$} & \gray{A+V+T} & \gray{9.68}  & \gray{35.02} & \gray{8.65} \\
\gray{T-MEP*~\cite{85_zhang2023transformer} (TCSVT'23)}  & \gray{$\times$}          & \gray{A+V+T} & \gray{13.25} & \gray{37.69} & \gray{\textbf{--}}      \\
\gray{T-MEP~\cite{85_zhang2023transformer} (TCSVT'23)}   & \gray{$\times$}          & \gray{A+V+T} & \gray{15.22} & \gray{39.00} & \gray{\textbf{--}}       \\
\hline
\end{tabular}
}
\label{tab_mafw_sota_compound}
\vspace{-1.0em}
\end{table}

\begin{table}[htbp]
\centering
\caption{Performance comparisons of AVF-MAE++ with advanced methods on CREMA-D (6-class).}
\setlength{\abovecaptionskip}{0.3em} 
\setlength{\arrayrulewidth}{0.40pt}  
\renewcommand\arraystretch{1.13}     
\resizebox{0.8\linewidth}{!}{
\begin{tabular}{lcccccc}
\toprule
Method  & SSL & Modality  &  \#Params (M)  &  UAR             & WAR \\
\hline

AuxFormer~\cite{125_goncalves2022auxformer} (ICASSP'22)       & $\times$  &  A  &  \textbf{--}     & \textbf{--}                       & 58.70                    \\
LR+eGeMAPS \cite{129_keesing2023emotion,98_eyben2015geneva}    & \checkmark  &  A    &  \textbf{--}  & 52.70                       & \textbf{--}                    \\
LR+wav2vec \cite{129_keesing2023emotion,96_baevski2020wav2vec} & \checkmark  &  A    &  \textbf{--}  & 66.50                       & \textbf{--}                    \\

Wav2Vec2.0~\cite{96_baevski2020wav2vec} (NeurIPS'20)   & \checkmark & A &  95     & 72.57  & 72.41 \\
HuBERT~\cite{32_hsu2021hubert} (TASLP'21)        & \checkmark & A &  95     & 72.72  & 72.57 \\
WavLM-Plus~\cite{87_chen2022wavlm} (J-STSP'22)        & \checkmark & A &  95     & 73.34  & 73.39 \\

\hline

AuxFormer~\cite{125_goncalves2022auxformer} (ICASSP'22)        & $\times$  &  V  &  \textbf{--}  &  \textbf{--}                          & 53.10                    \\
VO-LSTM~\cite{126_ghaleb2019multimodal} (ACII'19)          & $\times$  &  V  &  \textbf{--}     & \textbf{--}                       & 66.80                    \\
Goncalves et al.~\cite{127_goncalves2022robust} (TAFFC'22)   & $\times$  &  V  &  \textbf{--}     & \textbf{--}                       & 62.20                    \\
Lei et al.~\cite{128_lei2023audio} (TAFFC'23)                & $\times$  &  V  &  \textbf{--}     & 64.68                   & 64.76                    \\

SVFAP~\cite{3_sun2024svfap} (TAFFC'24)                  & \checkmark & V &  78    & 77.31          & 77.37                    \\ 
MAE-DFER~\cite{2_sun2023mae} (MM'23)                 & \checkmark & V &  85    & 77.33          & 77.38                    \\ 

\hline

EF-GRU~\cite{130_tran2022pre} (ICASSP'22)                     & $\times$  & A+V &  \textbf{--}     & \textbf{--}                       & 57.06                    \\
LF-GRU~\cite{130_tran2022pre} (ICASSP'22)                     & $\times$  & A+V &  \textbf{--}     & \textbf{--}                       & 58.53                    \\
TFN~\cite{131_zadeh2017tensor} (EMNLP'17)                   & $\times$  & A+V &  \textbf{--}     & \textbf{--}                       & 63.09                    \\
MATER~\cite{132_ghaleb2020multimodal} (ICIP'20)             & $\times$  & A+V &  \textbf{--}     & \textbf{--}                       & 67.20                   \\
MulT-Base~\cite{130_tran2022pre} (ICASSP'22)                  & \checkmark& A+V &  38     & \textbf{--}                       & 68.87                    \\ 
MulT-Large~\cite{130_tran2022pre} (ICASSP'22)                 & \checkmark& A+V &  89     & \textbf{--}                       & 70.22                    \\ 
AuxFormer~\cite{125_goncalves2022auxformer} (ICASSP'22)       & $\times$  & A+V &  \textbf{--}     & \textbf{--}                       & 71.70                    \\
AV-LSTM~\cite{126_ghaleb2019multimodal} (ACII'19)           & $\times$  & A+V &  \textbf{--}     & \textbf{--}                       & 72.90                    \\  
AV-Gating~\cite{126_ghaleb2019multimodal} (ACII'19)         & $\times$  & A+V &  \textbf{--}     & \textbf{--}                       & 74.00                    \\
Goncalves et al.~\cite{127_goncalves2022robust} (TAFFC'22)   & $\times$  & A+V &  \textbf{--}     & \textbf{--}                       & 77.30                    \\
Ladder Networks~\cite{133_goncalves2023learning} (ICASSP'23) & $\times$  & A+V &  \textbf{--}     & \textbf{--}                       & 80.30                    \\
\tabincell{l}{VQ-MAE-AV+\\Attn. Pooling~\cite{43_sadok2024audiovisual}}   & \checkmark& A+V  & 30 &  \textbf{--}      & 78.40 \\
\tabincell{l}{VQ-MAE-AV+\\Query2Emo~\cite{43_sadok2024audiovisual}}       & \checkmark& A+V  & 30      & \textbf{--}  & 80.40 \\
HiCMAE-T~\cite{1_sun2024hicmae} (IF'24)          & \checkmark& A+V  & 20     & 83.84  & 83.74 \\
HiCMAE-S~\cite{1_sun2024hicmae} (IF'24)          & \checkmark& A+V  & 46     & 84.46  & 84.38 \\
HiCMAE-B~\cite{1_sun2024hicmae} (IF'24)          & \checkmark& A+V  & 81     & 84.91  & 84.89 \\
\rowcolor{cvprblue!20}
\textbf{AVF-MAE++ (B)}          & \checkmark& A+V  & 169     & 85.10  & 85.09 \\
\rowcolor{cvprblue!20}
\textbf{AVF-MAE++ (L)}          & \checkmark& A+V  & 303     & \underline{85.69}  & \underline{85.60} \\
\rowcolor{cvprblue!20}
\textbf{AVF-MAE++ (H)}           & \checkmark& A+V  & 521    & \textbf{86.02}  & \textbf{85.95} \\
\hline
\end{tabular}
}
\label{tab_cremad_sota_six_classes}
\vspace{-1.0em}
\end{table}

\begin{table}[htbp]
\centering
\caption{Performance comparisons of AVF-MAE++ with SOTA methods on MSP-IMPROV.}
\setlength{\abovecaptionskip}{0.3em} 
\setlength{\arrayrulewidth}{0.40pt}  
\renewcommand\arraystretch{1.1}      
\resizebox{\linewidth}{!}{
\begin{tabular}{lcccccc}
\toprule
Method  & SSL & Modality  &  \#Params (M)  &  UAR    & WAR \\
\midrule
AuxFormer~\cite{125_goncalves2022auxformer} (ICASSP'22)                              & $\times$   & A+V &  \textbf{--}     & 62.97 & 70.28    \\
Tran et al.~\cite{130_tran2022pre} (ICASSP'22)        & \checkmark & A+V &  \textbf{--}     & 59.41 & 65.29    \\
AV-HuBERT~\cite{35_shi2022learning} (ICLR'22)     & \checkmark & A+V & 103    & \textbf{--}     & 65.27    \\
FAV-HuBERT~\cite{95_tran2023saaml} (MM'23)      & \checkmark & A+V & 103    & 61.05 & 68.35    \\
TAPT-HuBERT~\cite{95_tran2023saaml} (MM'23)      & \checkmark & A+V & 103    & 63.95 & 70.46    \\
CTAPT-HuBERT~\cite{95_tran2023saaml} (MM'23)     & \checkmark & A+V & 103    & 60.83 & 68.02    \\
AW-HuBERT~\cite{95_tran2023saaml} (MM'23)      & \checkmark & A+V & 103    & 65.72 & 71.80    \\

HiCMAE-T~\cite{1_sun2024hicmae} (IF'24)          & \checkmark & A+V  & 20     & 63.16  & 72.78 \\
HiCMAE-S~\cite{1_sun2024hicmae} (IF'24)          & \checkmark & A+V  & 46     & 63.90  & 74.35 \\
HiCMAE-B~\cite{1_sun2024hicmae} (IF'24)          & \checkmark & A+V  & 81     & 65.78 & 74.95 \\

\rowcolor{cvprblue!20}
\textbf{AVF-MAE++ (B)}         & \checkmark& A+V  & 169     & 68.90  & 75.07 \\
\rowcolor{cvprblue!20}
\textbf{AVF-MAE++ (L)}         & \checkmark& A+V  & 303     & \underline{68.95}  & \underline{75.59} \\
\rowcolor{cvprblue!20}
\textbf{AVF-MAE++ (H)}         & \checkmark& A+V  & 521     & \textbf{70.05}  & \textbf{76.07} \\

\hline
\end{tabular}
}
\label{tab_msp_improv_sota}
\end{table}

\begin{table}[htbp]
\centering
\caption{Performance comparisons of AVF-MAE++ with SOTA methods on CREMA-D (4-class).}
\setlength{\abovecaptionskip}{0.3em} 
\setlength{\arrayrulewidth}{0.40pt}  
\renewcommand\arraystretch{1.1}      
\resizebox{\linewidth}{!}{
\begin{tabular}{lcccccc}
\toprule
Method  & SSL & Modality &\#Params (M)  &  UAR   & WAR \\
\midrule
AuxFormer~\cite{125_goncalves2022auxformer} (ICASSP'22) & $\times$   & A+V &  \textbf{--}     & 91.10 & 91.62    \\
Tran et al.~\cite{130_tran2022pre} (ICASSP'22)           & \checkmark & A+V &  \textbf{--}    & 83.29 & 83.46    \\
AV-HuBERT~\cite{35_shi2022learning} (ICLR'22) & \checkmark & A+V & 103    & \textbf{--}     & 85.47    \\
FAV-HuBERT~\cite{95_tran2023saaml} (MM'23)  & \checkmark & A+V & 103    & 87.34 & 87.61    \\
TAPT-HuBERT~\cite{95_tran2023saaml} (MM'23) & \checkmark & A+V & 103    & 92.78 & 92.84    \\
CTAPT-HuBERT~\cite{95_tran2023saaml} (MM'23) & \checkmark & A+V & 103    & 90.52 & 90.39    \\
AW-HuBERT~\cite{95_tran2023saaml} (MM'23) & \checkmark & A+V & 103    & 93.65 & 93.65    \\

HiCMAE-T~\cite{1_sun2024hicmae} (IF'24)          & \checkmark& A+V  & 20     & 92.47  & 92.67 \\
HiCMAE-S~\cite{1_sun2024hicmae} (IF'24)          & \checkmark& A+V  & 46     & 93.34  & 93.48 \\
HiCMAE-B~\cite{1_sun2024hicmae} (IF'24)          & \checkmark& A+V  & 81     & \underline{94.00}  & \underline{94.13} \\

\rowcolor{cvprblue!20}
\textbf{AVF-MAE++ (B)}          & \checkmark& A+V  & 169     & 93.81  & 94.04 \\
\rowcolor{cvprblue!20}
\textbf{AVF-MAE++ (L)}          & \checkmark& A+V  & 303     & 93.05  & 93.16 \\
\rowcolor{cvprblue!20}
\textbf{AVF-MAE++ (H)}          & \checkmark& A+V  & 521     & \textbf{94.82}  & \textbf{94.92} \\

\hline

\end{tabular}
}
\label{tab_cremad_sota_four_cls}
\end{table}

\begin{table}[htbp]
\centering
\caption{Performance comparisons of AVF-MAE++ with advanced CEA methods on DFEW.}
\setlength{\abovecaptionskip}{0.3em} 
\setlength{\arrayrulewidth}{0.40pt}  
\renewcommand\arraystretch{1.10}      
\resizebox{0.94\linewidth}{!}{
\begin{tabular}{lcccccc}
\toprule
Method  & SSL & Modality  & \#Params (M)  &  UAR             & WAR \\
\hline
Wav2Vec2.0~\cite{96_baevski2020wav2vec} (NeurIPS'20)   & \checkmark & A &  95     & 36.15  & 43.05 \\
HuBERT~\cite{32_hsu2021hubert} (TASLP'21)        & \checkmark & A &  95     & 35.98  & 43.24 \\
WavLM-Plus~\cite{87_chen2022wavlm} (J-STSP'22)        & \checkmark & A &  95     & 37.78  & 44.64 \\

\hline%

C3D~\cite{107_tran2015learning} (ICCV'15)             &  $\times$  &  V  &   78     &  42.74 & 53.54  \\
R(2+1)D-18~\cite{115_tran2018closer} (CVPR'18)        &  $\times$  &  V  &   33     &  42.79 & 53.22 \\
3D ResNet-18~\cite{116_hara2018can} (CVPR'18)         &  $\times$  &  V  &   33     & 46.52 & 58.27    \\ 
EC-STFL~\cite{9_jiang2020dfew} (MM'20)            &  $\times$  &  V  &  \textbf{--}    & 45.35 & 56.51    \\ 
ResNet-18+LSTM~\cite{92_zhao2021former} (MM'21)    &  $\times$  &  V  &  \textbf{--}    & 51.32 & 63.85    \\ 
ResNet-18+GRU~\cite{92_zhao2021former} (MM'21)    &  $\times$  &  V  &  \textbf{--}    & 51.68 & 64.02    \\ 
Former-DFER~\cite{92_zhao2021former} (MM'21)       &  $\times$  &  V  &  18   & 53.69 & 65.70   \\ 
CEFLNet~\cite{117_liu2022clip} (IS'22)              &  $\times$  &  V  &  13    & 51.14 & 65.35   \\ 
EST~\cite{118_liu2023expression} (PR'23)            &  $\times$  &  V  &  43   & 53.43 & 65.85   \\
STT~\cite{119_ma2022spatio} (arXiv'22)                 &  $\times$  &  V  &  \textbf{--}    & 54.58 & 66.65   \\ 
NR-DFERNet~\cite{120_li2022nr} (arXiv'22)              &  $\times$  &  V  &  \textbf{--}    & 54.21 & 68.19   \\
DPCNet~\cite{121_wang2022dpcnet} (MM'22)            &  $\times$  &  V  &  51   & 57.11 & 66.32   \\ 
IAL~\cite{122_li2023intensity} (AAAI'23)              &  $\times$  &  V  &  19   & 55.71 & 69.24   \\ 
M3DFEL~\cite{123_wang2023rethinking} (CVPR'23)        &  $\times$  &  V  &  \textbf{--}     & 56.10 & 69.25       \\
T-MEP~\cite{85_zhang2023transformer} (TCSVT'23)      &  $\times$  &  V  &  5     & 54.14 & 65.22 \\

Video Swin-T~\cite{31_liu2022video} (CVPR'22)              & $\times$ & V &  88   & 59.38 & 71.90 \\ 
UniLearn~\cite{84_chen2024unilearn} (arXiv'24)              & \checkmark & V &  101   & \underline{66.80} & \underline{76.68} \\ 
CLIPER~\cite{124_li2024cliper} (ICME’24)      & \checkmark & V &  88   & 57.56 & 70.84 \\ 
S2D~\cite{111_chen2024static} (TAFFC'24)              & \checkmark & V &  9   &  65.45 & 74.81 \\ 
DFER-CLIP~\cite{26_zhao2023prompting} (BMVC’23)       &  \checkmark  & V & 153   & 59.61 & 71.25\\
SVFAP~\cite{3_sun2024svfap} (TAFFC'24)              & \checkmark & V &  78   & 62.83 & 74.27 \\
EmoCLIP~\cite{113_foteinopoulou2024emoclip} (FG’24)      & \checkmark & V &  \textbf{--}   & 58.04 & 62.12 \\ 
MAE-DFER~\cite{2_sun2023mae} (MM'23)              & \checkmark & V &  85   & 63.41 & 74.43 \\ 
VideoMAE~\cite{7_tong2022videomae} (NeurIPS'22)              & \checkmark & V &  86   & 58.32 & 70.94 \\
MoCo~\cite{158_he2020momentum} (CVPR'20)              & \checkmark & V &  32   & 53.47 & 67.45 \\ 
A$^3$lign-DFER~\cite{114_tao20243} (arXiv'24)              & \checkmark & V &  \textbf{--}   & 64.09 & 74.20 \\ 
\hline%

ResNet-18+LSTM~\cite{85_zhang2023transformer} (TCSVT'23)    &  $\times$& A+V   &  \textbf{--}      & 52.41  & 64.32 \\
C3D+LSTM~\cite{85_zhang2023transformer} (TCSVT'23)         &  $\times$& A+V   &  \textbf{--}      & 53.77  & 65.17 \\
AMH~\cite{108_yoon2020attentive} (ICASSP'20)                 &  $\times$& A+V   &  \textbf{--}      & 54.48  & 66.51 \\
T-MEP*~\cite{85_zhang2023transformer} (TCSVT'23)           &  $\times$& A+V   &  61     & 55.06  & 66.30 \\
T-MEP~\cite{85_zhang2023transformer} (TCSVT'23)            &  $\times$& A+V   &  61      & 57.16  & 68.85 \\
HiCMAE-T~\cite{1_sun2024hicmae} (IF'24)          & \checkmark& A+V  & 20     & 60.13  & 72.43 \\
HiCMAE-S~\cite{1_sun2024hicmae} (IF'24)           & \checkmark& A+V  & 46     & 63.05  & 74.33 \\
HiCMAE-B~\cite{1_sun2024hicmae} (IF'24)           & \checkmark& A+V  & 81     & 63.76  & 75.01 \\

\rowcolor{cvprblue!20}
\textbf{AVF-MAE++ (B)}         & \checkmark& A+V  & 169     & 63.74  & 75.42 \\
\rowcolor{cvprblue!20}
\textbf{AVF-MAE++ (L)}          & \checkmark& A+V  & 303     & 65.14  & 76.24 \\
\rowcolor{cvprblue!20}
\textbf{AVF-MAE++ (H)}           & \checkmark& A+V  & 521    & \textbf{66.88}  & \textbf{77.45} \\

\hline%

\gray{FineCLIPER~\cite{82_chen2024finecliper} (MM'24)}           & \gray{\checkmark} & \gray{T+V}  & \gray{20}     & 65.98  & 76.21  \\

\hline%
\end{tabular}
}
\label{tab_dfew_sota}
\end{table}

\begin{table}[htbp]
\centering
\caption{Performance comparisons of AVF-MAE++ with SOTA CEA methods on IEMOCAP.}
\setlength{\abovecaptionskip}{0.3em} 
\setlength{\arrayrulewidth}{0.40pt}  
\renewcommand\arraystretch{0.98}     
\resizebox{0.9\linewidth}{!}{
\begin{tabular}{lcccccc}
\toprule
Method  & SSL & Modality  &\#Params (M)  &  UAR             & WAR \\
\midrule
FBANK~\cite{134_tseng2024av} (ICASSP'24)         & $\times$ & A   &  \textbf{--}  &  \textbf{--}        & 51.52    \\
AV-HuBERT~\cite{35_shi2022learning} (arXiv'22) & \checkmark & A &  90 &  \textbf{--}      & 58.54    \\
RepLAI~\cite{135_mittal2022learning} (NeurIPS'22) & \checkmark & A &   5 &  \textbf{--}      & 57.53    \\
AVBERT~\cite{97_lee2020parameter} (ICLR'21)   & \checkmark & A &  10 &  \textbf{--}      & 60.94    \\
MAViL~\cite{97_lee2020parameter} (ICLR'21)    & \checkmark & A &  86 &  \textbf{--}     & 59.46    \\
Wav2Vec2.0~\cite{96_baevski2020wav2vec} (NeurIPS'20) & \checkmark & A &  95    & 69.88  & 67.32    \\
HuBERT~\cite{32_hsu2021hubert} (TASLP'21)           & \checkmark & A &  95    & 68.33  & 66.34    \\
WavLM-Plus~\cite{87_chen2022wavlm} (J-STSP'22)       & \checkmark & A &  95    & 68.64  & 67.12    \\

\hline

HOG~\cite{136_dalal2005histograms} (CVPR'05)   & $\times$   & V & \textbf{--}      & \textbf{--}   & 35.83    \\
AV-HuBERT~\cite{35_shi2022learning} (arXiv'22) & \checkmark & V & 103    & \textbf{--}   & 26.59    \\
RepLAI~\cite{135_mittal2022learning} (NeurIPS'22) & \checkmark & V &  15    & \textbf{--}   & 40.72    \\
AVBERT~\cite{97_lee2020parameter} (ICLR'21)   & \checkmark & V &  37    & \textbf{--}   & 45.80    \\
MAViL~\cite{97_lee2020parameter} (ICLR'21)    & \checkmark & V &  87    & \textbf{--}   & 43.03    \\

\hline

AV-HuBERT~\cite{35_shi2022learning} (ICLR'22) & \checkmark & A+V & 103    & \textbf{--}   & 46.45    \\
AVBERT~\cite{97_lee2020parameter} (ICLR'21)   & \checkmark & A+V &  43    & \textbf{--}   & 61.87    \\
MAViL~\cite{97_lee2020parameter} (ICLR'21)    & \checkmark & A+V & 187    & \textbf{--}   & 54.94    \\
HiCMAE-T~\cite{1_sun2024hicmae} (IF'24)          & \checkmark& A+V  & 20     & 66.85  & 66.62 \\
HiCMAE-S~\cite{1_sun2024hicmae} (IF'24)          & \checkmark& A+V  & 46     & 67.46  & 67.49 \\
HiCMAE-B~\cite{1_sun2024hicmae} (IF'24)          & \checkmark& A+V  & 81     & 68.21  & 68.36 \\

\rowcolor{cvprblue!20}
\textbf{AVF-MAE++ (B)}         & \checkmark& A+V  & 169     & 69.53  & 71.47 \\
\rowcolor{cvprblue!20}
\textbf{AVF-MAE++ (L)}         & \checkmark& A+V  & 303     & \underline{69.86}  & \underline{71.65} \\
\rowcolor{cvprblue!20}
\textbf{AVF-MAE++ (H)}          & \checkmark& A+V  & 521     & \textbf{72.71}  & \textbf{73.83} \\

\hline
\end{tabular}
}
\label{tab_iemocap_sota}
\end{table}

\begin{table}[htbp]
\centering
\caption{Performance comparisons of AVF-MAE++ with advanced methods on MER24-T\&V.}
\setlength{\abovecaptionskip}{0.3em} 
\setlength{\arrayrulewidth}{0.40pt}  
\renewcommand\arraystretch{0.98}     
\resizebox{0.9\linewidth}{!}{
\begin{tabular}{lcccccc}
\toprule
Method  & SSL & Modality  & \#Params (M)  &  WAR  & WA-F1 \\
\midrule
Whisper~\cite{93_radford2023robust} (ICML'23)  & $\times$ & A &  1550   & 63.27  & 63.23 \\
eGeMAPS~\cite{98_eyben2015geneva} (TAFFC'15) & $\times$ & A &  \textbf{--}   & 42.88  & 39.68 \\
VGGish~\cite{137_hershey2017cnn} (ICASSP'17) & $\times$ & A  &  \textbf{--}   & 50.20  & 48.60 \\
emotion2vec~\cite{153_ma2023emotion2vec} (ACL'24) & \checkmark & A &  94   & 56.48  & 56.08 \\
Wav2Vec2.0~\cite{96_baevski2020wav2vec} (NeurIPS'20) & \checkmark  & A &  95   & 65.83  & 65.50 \\
HuBERT~\cite{32_hsu2021hubert} (TASLP'21) & \checkmark & A  &  95   & 69.43  & 69.26 \\

\midrule

EmoNet~\cite{154_kahou2016emonets} (JMUI'16) & $\times$ & V & \textbf{--}             & 53.18    & 51.76   \\
SENet-FER2013~\cite{152_hu2018squeeze} (CVPR'18) & $\times$ & V & 28             & 58.79    & 57.67   \\
ResNet-FER2013~\cite{90_he2016deep} (CVPR'16) & $\times$ & V & 26             & 59.66    & 58.73   \\
MANet-RAFDB~\cite{91_zhao2021learning} (TIP'21) & $\times$ & V & 51            & 61.10    & 59.91   \\
CLIP-base~\cite{155_radford2021learning} (ICML'21) & \checkmark & V & \textbf{--}             & 62.56    & 61.74   \\
CLIP-large~\cite{155_radford2021learning} (ICML'21) & \checkmark & V & \textbf{--}             & 67.17    & 66.66   \\
EVA-02~\cite{156_fang2024eva} (IVC'24) & \checkmark & V & 86             & 62.28    & 61.41   \\
DINOv2~\cite{94_oquab2023dinov2} (arXiv'23)   & \checkmark & V & \textbf{--}    & 59.57    & 58.44   \\
VideoMAE~\cite{7_tong2022videomae} (NeurIPS’22) & \checkmark & V & 86             & 64.93    & 64.50   \\

\midrule

HiCMAE~\cite{1_sun2024hicmae} (IF'24)          & \checkmark & A+V  & 81     & 70.95  & 70.18 \\

\rowcolor{cvprblue!20}
\textbf{AVF-MAE++ (B)}          & \checkmark & A+V  & 169     & 72.11  & 71.24 \\
\rowcolor{cvprblue!20}
\textbf{AVF-MAE++ (L)}          & \checkmark & A+V  & 303     & \textbf{72.33}  & \underline{71.64} \\
\rowcolor{cvprblue!20}
\textbf{AVF-MAE++ (H)}          & \checkmark & A+V  & 521     & \underline{72.28}  & \textbf{71.75} \\

\hline
\end{tabular}
}
\label{tab_MER24-T-V_sota}
\end{table}

\begin{table}[htbp]
\centering
\caption{Performance comparisons of AVF-MAE++ with advanced methods on RAVDESS.}
\setlength{\abovecaptionskip}{0.3em} 
\setlength{\arrayrulewidth}{0.40pt}  
\renewcommand\arraystretch{1.11}     
\resizebox{0.9\linewidth}{!}{
\begin{tabular}{lcccccc}
\toprule
Method  & SSL & Modality  & \#Params (M)  &  UAR             & WAR \\
\midrule

LR+eGeMAPS \cite{129_keesing2023emotion,98_eyben2015geneva}    & \checkmark  &  A    &  \textbf{--}  & 50.30                       & \textbf{--}                    \\
LR+wav2vec \cite{129_keesing2023emotion,96_baevski2020wav2vec} & \checkmark  &  A    &  \textbf{--}  & 68.80                       & \textbf{--}                    \\

Wav2Vec2.0~\cite{96_baevski2020wav2vec} (NeurIPS'20)   & \checkmark & A &  95     & 73.44  & 74.38 \\
HuBERT~\cite{32_hsu2021hubert} (TASLP'21)        & \checkmark & A &  95     & 74.15  & 74.37 \\
WavLM-Plus~\cite{87_chen2022wavlm} (J-STSP'22)        & \checkmark & A &  95     & 75.28  & 75.36 \\
\midrule

VO-LSTM~\cite{126_ghaleb2019multimodal} (ACII'19)           & $\times$  &  V  &  \textbf{--}     & \textbf{--}                       & 60.50                    \\

3D ResNeXt-50~\cite{145_su2020msaf} (arXiv'20)               & $\times$  &  V  & 26     & \textbf{--}                       & 62.99                    \\

SVFAP~\cite{3_sun2024svfap} (TAFFC'24)                 & \checkmark & V &  78    & 75.15          & 75.01                    \\ 
MAE-DFER~\cite{2_sun2023mae} (MM'23)                 & \checkmark & V &  85    & 75.91          & 75.56                    \\ 

\midrule

AV-LSTM~\cite{126_ghaleb2019multimodal} (ACII'19)           & $\times$  & A+V &  \textbf{--}     & \textbf{--}                       & 65.80                    \\  
AV-Gating~\cite{126_ghaleb2019multimodal} (ACII'19)         & $\times$  & A+V &  \textbf{--}     & \textbf{--}                       & 67.70                    \\
MCBP~\cite{146_fukui2016multimodal} (EMNLP'16)   & $\times$  & A+V & 51     & \textbf{--}                       & 71.32                    \\
MMTM~\cite{147_joze2020mmtm} (CVPR'20)          & $\times$  & A+V & 32     & \textbf{--}                       & 73.12                   \\
MSAF~\cite{145_su2020msaf} (arXiv'20)                        & $\times$  & A+V & 26    & \textbf{--}                       & 74.86                   \\
ERANNs~\cite{148_verbitskiy2022eranns} (PRL'22)            & $\times$  & A+V & \textbf{--}      & \textbf{--}                       & 74.80                   \\
CFN-SR~\cite{149_fu2021cross} (arXiv'21)                     & $\times$  & A+V & 26     & \textbf{--}                       & 75.76                   \\

MATER~\cite{132_ghaleb2020multimodal} (ICIP'20)            & $\times$  & A+V &  \textbf{--}     & \textbf{--}                       & 76.30                   \\
MulT~\cite{150_tsai2019multimodal} (ACL'19)                & $\times$  & A+V &  \textbf{--}     & \textbf{--}                       & 76.60                   \\
AVT~\cite{151_chumachenko2022self} (ICPR'22)               & $\times$  & A+V &  \textbf{--}     & \textbf{--}                       & 79.20                   \\

\tabincell{l}{VQ-MAE-AV+\\Attn. Pooling~\cite{43_sadok2024audiovisual}}   & \checkmark& A+V  & 30      & \textbf{--}  & 83.20 \\
\tabincell{l}{VQ-MAE-AV+\\Query2Emo~\cite{43_sadok2024audiovisual}}       & \checkmark& A+V  & 30      & \textbf{--}  & 84.80 \\

HiCMAE-T~\cite{1_sun2024hicmae} (IF'24)          & \checkmark & A+V  & 20     & 86.26  & 86.11 \\
HiCMAE-S~\cite{1_sun2024hicmae} (IF'24)          & \checkmark & A+V  & 46     & 86.85  & 86.67 \\
HiCMAE-B~\cite{1_sun2024hicmae} (IF'24)          & \checkmark & A+V  & 81     & \textbf{87.96}  & \textbf{87.99} \\

\rowcolor{cvprblue!20}
\textbf{AVF-MAE++ (B)}          & \checkmark & A+V  & 169     & 85.09  & 85.07 \\
\rowcolor{cvprblue!20}
\textbf{AVF-MAE++ (L)}          & \checkmark & A+V  & 303     & 86.98  & 87.22 \\
\rowcolor{cvprblue!20}
\textbf{AVF-MAE++ (H)}          & \checkmark & A+V  & 521     & \underline{87.44}  & \underline{87.57} \\

\hline
\end{tabular}
}
\label{tab_ravdess_sota}
\vspace{-1em}
\end{table}

\begin{table}[htbp]
\centering
\caption{Performance comparisons of AVF-MAE++ with advanced methods on MER-MULTI.}
\setlength{\abovecaptionskip}{0.3em} 
\setlength{\arrayrulewidth}{0.40pt}  
\renewcommand\arraystretch{1.11}     
\resizebox{0.9\linewidth}{!}{
\begin{tabular}{lcccccc}
\toprule
Method  & SSL & Modality  & \#Params (M)  &  UAR  & WA-F1 \\
\midrule
eGeMAPS~\cite{98_eyben2015geneva} (TAFFC'15) & $\times$ & A &  \textbf{--}   & \textbf{--}  & 17.28 \\
VGGish~\cite{137_hershey2017cnn} (ICASSP'17)   & $\times$ & A &  \textbf{--}  & \textbf{--}  & 40.76 \\
Wav2Vec2.0~\cite{96_baevski2020wav2vec} (NeurIPS'20)    & \checkmark & A &  95     & 51.36	& 51.48 \\
HuBERT~\cite{32_hsu2021hubert} (TASLP'21)         & \checkmark & A &  95     & 50.32 & 52.70 \\
WavLM-Plus~\cite{87_chen2022wavlm} (J-STSP'22)         & \checkmark & A &  95     & 53.43 & 54.16 \\
HuBERT-CH~\cite{89_zhang2022wenetspeech} (ICASSP'22) & \checkmark & A &  95     & \textbf{--}     & 61.16 \\

HiCMAE-T~\cite{1_sun2024hicmae} (IF'24)          & \checkmark & A &  8     & 48.35  & 51.33 \\
HiCMAE-S~\cite{1_sun2024hicmae} (IF'24)          & \checkmark & A & 18     & 51.09  & 54.16 \\
HiCMAE-B~\cite{1_sun2024hicmae} (IF'24)          & \checkmark & A & 32     & 51.43  & 55.33 \\

\midrule
ResNet-MSCeleb~\cite{90_he2016deep} (CVPR'16)    & $\times$ & V & 26    & \textbf{--}    & 40.32    \\
ResNet-ImageNet~\cite{90_he2016deep} (CVPR'16)   & $\times$ & V & 26    & \textbf{--}    & 44.91   \\
SENet-FER2013~\cite{152_hu2018squeeze} (CVPR'18) & $\times$ & V & 28    & \textbf{--}    & 56.69    \\
ResNet-FER2013~\cite{90_he2016deep} (CVPR'16)   & $\times$ & V & 26    & \textbf{--}    & 57.44    \\
MANet-RAFDB~\cite{91_zhao2021learning} (TIP'21) & $\times$ & V & 51    & \textbf{--}    & 56.19   \\

HiCMAE-T~\cite{1_sun2024hicmae} (IF'24)          & \checkmark & V &   8    & 50.52  & 58.37  \\
HiCMAE-S~\cite{1_sun2024hicmae} (IF'24)          & \checkmark & V &  18    & 51.53  & 59.25  \\
HiCMAE-B~\cite{1_sun2024hicmae} (IF'24)          & \checkmark & V &  32    & 52.31  & 59.87  \\

\midrule

\tabincell{l}{ResNet-FER2013+\\HuBERT-CH~\cite{64_lian2023mer} (MM'23)}   & \checkmark & A+V & 121    & \textbf{--}    & 69.11    \\
\tabincell{l}{MANet-RAFDB+\\HuBERT-CH~\cite{64_lian2023mer} (MM'23)}      & \checkmark & A+V & 146    & \textbf{--}    & 70.32   \\

HiCMAE-T~\cite{1_sun2024hicmae} (IF'24)          & \checkmark & A+V  & 20     & 59.91  & 68.56 \\
HiCMAE-S~\cite{1_sun2024hicmae} (IF'24)          & \checkmark & A+V  & 46     & 63.18  & 70.22 \\
HiCMAE-B~\cite{1_sun2024hicmae} (IF'24)          & \checkmark & A+V  & 81     & 64.15  & \underline{71.33} \\

\rowcolor{cvprblue!20}
\textbf{AVF-MAE++ (B)}          & \checkmark & A+V  & 169     & 64.87  & 69.56 \\
\rowcolor{cvprblue!20}
\textbf{AVF-MAE++ (L)}          & \checkmark & A+V  & 303     & \underline{66.34}  & 70.79 \\
\rowcolor{cvprblue!20}
\textbf{AVF-MAE++ (H)}          & \checkmark & A+V  & 521    & \textbf{68.20}  & \textbf{72.26} \\

\hline
\end{tabular}
}
\label{tab_mer-multi_sota}
\end{table}

\begin{table}[htbp]
\centering
\caption{Performance comparisons of the AVF-MAE++ with state-of-the-art methods on AVCAffe.} 
\setlength{\abovecaptionskip}{0.3em}  
\setlength{\arrayrulewidth}{0.40pt}   
\renewcommand\arraystretch{1.07}      
\resizebox{0.9\linewidth}{!}{
\begin{tabular}{lccccccccccc}
\toprule
Method   & SSL &   Modality    & \#Params (M)    & Arousal             & Valence            \\ \hline

\tabincell{l}{VGG-16+\\MC3-18~\cite{61_sarkar2023avcaffe} (AAAI'23)}          &  $\times$  &   A+V  &  47     & 38.90  & 41.70  \\
\tabincell{l}{VGG-16+\\3D ResNet-18~\cite{61_sarkar2023avcaffe} (AAAI'23)}    &  $\times$  &   A+V  &  69     & 37.30  & 39.40  \\
\tabincell{l}{VGG-16+\\R(2+1)D-18~\cite{61_sarkar2023avcaffe} (AAAI'23)}      &  $\times$  &   A+V  &  67     & 40.50  & 39.50  \\
\tabincell{l}{ResNet-18+\\MC3-18~\cite{61_sarkar2023avcaffe} (AAAI'23)}       &  $\times$  &   A+V  &  44     & 36.00  & 39.20  \\
\tabincell{l}{ResNet-18+\\3D ResNet-18~\cite{61_sarkar2023avcaffe} (AAAI'23)} &  $\times$  &   A+V  &  66     & 35.10  & 39.10  \\
\tabincell{l}{ResNet-18+\\R(2+1)D-18~\cite{61_sarkar2023avcaffe} (AAAI'23)}   &  $\times$  &   A+V  &  64     & 39.50  & 37.70  \\

HiCMAE-T~\cite{1_sun2024hicmae} (IF'24)           & \checkmark & A+V  & 20     & 39.64  & 36.74 \\
HiCMAE-S~\cite{1_sun2024hicmae} (IF'24)           & \checkmark & A+V  & 46     & 42.13  & 42.65 \\
HiCMAE-B~\cite{1_sun2024hicmae} (IF'24)           & \checkmark & A+V  & 81     & 43.18  & 44.20 \\

\rowcolor{cvprblue!20}
\textbf{AVF-MAE++ (B)}          & \checkmark & A+V  & 169     & 43.02  & 46.93 \\
\rowcolor{cvprblue!20}
\textbf{AVF-MAE++ (L)}          & \checkmark & A+V  & 303     & \underline{45.21}  & \underline{47.83} \\
\rowcolor{cvprblue!20}
\textbf{AVF-MAE++ (H)}          & \checkmark & A+V  & 521     & \textbf{47.25}  & \textbf{49.66} \\

\hline
\end{tabular}
}
\label{tab_avcaffe_sota}
\end{table}

\begin{table}[htbp]
\centering
\caption{Performance comparisons of AVF-MAE++ with advanced methods on Werewolf-XL.}
\setlength{\abovecaptionskip}{0.3em} 
\setlength{\arrayrulewidth}{0.40pt}  
\renewcommand\arraystretch{1.0}      
\resizebox{0.9\linewidth}{!}{
\begin{tabular}{lccccccccc}
\toprule
Method & SSL & Modality & Arousal  & Valence & Dominance \\  
\hline

eGeMAPS~\cite{98_eyben2015geneva} (TAFFC'15) & $\times$ &  A  & 23.45    & 8.08   & 31.15     \\
VGGish~\cite{137_hershey2017cnn} (ICASSP'17)   & $\times$ &  A  & 22.88    & 5.69   & 29.59     \\

HiCMAE-T~\cite{1_sun2024hicmae} (IF'24)          & \checkmark &  A  & 26.54   & 12.94    & 37.88     \\
HiCMAE-S~\cite{1_sun2024hicmae} (IF'24)          & \checkmark &  A  & 28.40    & 15.46   & 37.83     \\
HiCMAE-B~\cite{1_sun2024hicmae} (IF'24)          & \checkmark &  A  & 30.04    & 17.63   & 36.60     \\

\hline

HOG~\cite{136_dalal2005histograms} (CVPR'05) & $\times$ &  V  & 20.82    & 52.54   &
24.76     \\
VGGFace~\cite{106_parkhi2015deep} (BMVC'15)   & $\times$ &  V  & 7.24    & 62.96   &
14.30     \\

SVFAP~\cite{3_sun2024svfap} (TAFFC'24)     & \checkmark &  V   & 23.51    & 67.11  & 34.61    \\

HiCMAE-T~\cite{1_sun2024hicmae} (IF'24)          & \checkmark &  V   & 22.45    & 66.55  & 33.57    \\
HiCMAE-S~\cite{1_sun2024hicmae} (IF'24)          & \checkmark &  V   & 23.11    & 67.05   & 34.00     \\
HiCMAE-B~\cite{1_sun2024hicmae} (IF'24)          & \checkmark &  V   & 24.04    & 67.03   & 34.91    \\

\hline

Zhang et al.~\cite{62_zhang2021werewolf} (TAFFC'23)  & $\times$ &  A+V  & 16.41    & 63.14   &
35.40    \\

HiCMAE-T~\cite{1_sun2024hicmae} (IF'24)          & \checkmark &  A+V & 30.47    & 68.50   & 42.37  \\
HiCMAE-S~\cite{1_sun2024hicmae} (IF'24)          & \checkmark &  A+V & 31.08    & 68.92   & 41.38   \\
HiCMAE-B~\cite{1_sun2024hicmae} (IF'24)          & \checkmark &  A+V & 33.74    & 69.23   & 40.66  \\

\rowcolor{cvprblue!20}
\textbf{AVF-MAE++ (B)}          & \checkmark &  A+V & \underline{44.33}    & 71.22   & \textbf{52.59}   \\
\rowcolor{cvprblue!20}
\textbf{AVF-MAE++ (L)}    & \checkmark &  A+V & 43.54    & \underline{72.09}   & 52.07   \\
\rowcolor{cvprblue!20}
\textbf{AVF-MAE++ (H)}          & \checkmark &  A+V & \textbf{44.99}    & \textbf{72.19}   & \underline{52.35}  \\

\hline
\end{tabular}
}
\label{tab_werewolf-xl_sota}
\end{table}

\begin{table*}[t!]
\centering
\caption{The comparative results of SOTA MER methods with AVF-MAE++ in terms of Unweighted Average Recall (UAR) and Unweighted F1-score (UF1) on five popular MER datasets.} 
\setlength{\abovecaptionskip}{0.6em}  
\setlength{\arrayrulewidth}{0.40pt}   
\renewcommand\arraystretch{1.14}      
\resizebox{\textwidth}{!}{
\begin{tabular}{lccccccccccccc}
\toprule[0.40pt] 
\multirow{2}{*}{Method} & \multirow{2}{*}{Venue} & \multirow{2}{*}{SSL}  
& \multicolumn{2}{c}{SAMM}  & \multicolumn{2}{c}{CASME II} & \multicolumn{2}{c}{SMIC} & \multicolumn{2}{c}{CAS(ME)$^3$} & \multicolumn{2}{c}{MMEW}\\ 
\cmidrule(lr){4-5} \cmidrule(lr){6-7} \cmidrule(lr){8-9} \cmidrule(lr){10-11} \cmidrule(lr){12-13}
                       &       &       &         UAR  & UF1  & UAR  & UF1 & UAR  & UF1 & UAR  & UF1 & UAR  & UF1  \\ 
\midrule
\multicolumn{4}{l}{\textit{\textcolor{sgreen}{Traditional methods}}} \\
LBP-TOP~\cite{80_zhao2007dynamic}   & T-PAMI'07       & $\times$  & 41.02  & 39.54  & 74.29 & 70.26  & 52.80 & 20.00  & 21.39 & 21.78 & 63.61 & 64.23 \\
Bi-WOOF~\cite{138_liong2018less}   & IMAGE'18       & $\times$  & 51.39  & 52.11  & 80.26 & 78.05  & 58.29 & 57.27  & \textbf{--} & \textbf{--} & \textbf{--} & \textbf{--} \\
\midrule
\multicolumn{4}{l}{\textit{\textcolor{sgreen}{Deep learning methods}}} \\
AlexNet~\cite{139_zhang2022review}       & IJCNN'22   & $\times$  & 66.42  & 61.04  & 83.12 & 79.94  & 63.73 & 62.01  & 26.34 & 25.70 & \textbf{--} & \textbf{--} \\

GoogLeNet~\cite{140_ballester2016performance}       & AAAI'16   & $\times$  & 59.92  & 51.24  & 64.14 & 59.89  & 55.11 & 51.23  & \textbf{--} & \textbf{--} & \textbf{--} & \textbf{--} \\

STSTNet~\cite{75_liong2019shallow}     & FG'19     & $\times$  & 68.10  & 65.88  & 86.86 & 83.82  & 70.13 & 68.01  & 37.92 & 37.95 & 82.53 & 80.37 \\

VGG16~\cite{141_sengupta2019going}       & 	Frontiers in Neuroscience'19   & $\times$  & 47.93  & 48.70  & 82.02 & 81.66  & 59.64 & 58.00  & \textbf{--} & \textbf{--} & \textbf{--} & \textbf{--} \\

CapsuleNet~\cite{76_van2019capsulenet}    & FG'19      & $\times$  & 59.89  & 62.09  & 70.18 & 70.68  & 58.77 & 58.20  & \textbf{--} & \textbf{--} & 68.34 & 67.62 \\

RCN-A~\cite{78_xia2020revealing}    & TIP'20      & $\times$  & 67.20  & 76.01  & 81.20 & 85.12  & 64.40 & 63.26  & 38.93 & 39.28 & \textbf{--} & \textbf{--} \\

EMR~\cite{79_liu2019neural}    & FG'19      & $\times$  & 71.52  & 77.54  & 82.09 & 82.93  & 75.30 & 74.61  & 36.56 & 36.13 & 82.66 & 81.49 \\ 

OFF-ApexNet~\cite{142_gan2019off}       & IMAGE'19   & $\times$  & 53.92  & 54.09  & 86.81 & 87.64  & 66.95 & 68.17  & \textbf{--} & \textbf{--} & \textbf{--} & \textbf{--} \\

$\mu$-BERT~\cite{74_nguyen2023micron}    & CVPR'23      & $\times$  & \textbf{84.75}  & \textbf{--}  & 89.14 & 90.34  & \textbf{83.84} & \textbf{85.50}  & 61.25 & 56.04 & \textbf{--} & \textbf{--} \\

FeatRef~\cite{81_zhou2022feature}    & PR'22      & $\times$  & 71.55  & 73.72  & 88.73 & 89.15  & 70.83 & 70.11  & 34.13 & 34.93 & \textbf{--} & 82.11 \\

Dual-Inception~\cite{143_zhou2019dual}       & FG'19   & $\times$  & 56.63  & 58.68  & 85.60 & 86.21  & 67.26 & 66.45  & \textbf{--} & \textbf{--} & \textbf{--} & \textbf{--} \\

SLSTT-LSTM~\cite{144_zhang2022short}       & TAFFC'22   & $\times$ & 64.30  & 71.50  & 88.50 & 90.10  & 72.00 & 74.00  & \textbf{--} & \textbf{--} & \textbf{--} & \textbf{--} \\

HTNet~\cite{77_wang2024htnet}     & Neurocomputing'24     & $\times$  & 81.24  & 81.31  & 95.16 & \textbf{95.32}  & 79.05 & 80.49  & 54.15 & 57.67 & \textbf{--} & \textbf{84.33} \\

HiCMAE~\cite{1_sun2024hicmae}       & IF'24   & \checkmark  & \textbf{--}  & \textbf{--}  & 90.21 & 92.03  & 81.03 & 80.33  & \textbf{--} & \textbf{--} & \textbf{--} & \textbf{--} \\

\rowcolor{cvprblue!20}
\textbf{AVF-MAE++ (B)}       & \textbf{--}   & \checkmark  & 80.43 & 81.58 & \underline{96.67} & 93.58 & 80.56 & 83.23 & 61.06 & 63.18 & 83.96 & \underline{83.76} \\
\rowcolor{cvprblue!20}
\textbf{AVF-MAE++ (L)}       & \textbf{--}   & \checkmark & \underline{81.55} & \textbf{82.53}  & \textbf{96.72}  & 94.03  & \underline{81.67}  & \underline{83.79}  & \textbf{66.02}  & \textbf{67.88}   & \underline{85.81}  & 83.41 \\
\rowcolor{cvprblue!20}
\textbf{AVF-MAE++ (H) }     & \textbf{--}    & \checkmark  & 81.01  & \underline{81.62}  & 96.54 & \underline{94.11} & 81.64  & 83.55  & \underline{65.88}  & \underline{65.34}  & \textbf{86.23}  & \textbf{84.33} \\

\hline
\end{tabular}
}
\label{tab:sota-mer}
\vspace{-1.0em}
\end{table*}

\subsection{Main Results}
To thoroughly demonstrate the effectiveness of AVF-MAE++, we adapt its pre-trained representations to 17 benchmark datasets spanning three downstream AVFA tasks. Furthermore, we provide extensive comparisons between AVF-MAE++ and other state-of-the art AVFA methods on MAFW (11-class)~\cite{63_liu2022mafw}, MAFW (43-class)~\cite{63_liu2022mafw}, MSP-IMPROV~\cite{68_busso2016msp}, CRMEA-D (4-class)~\cite{66_cao2014crema}, DFEW~\cite{9_jiang2020dfew}, AVCAffe~\cite{61_sarkar2023avcaffe}, IEMOCAP~\cite{65_busso2008iemocap}, CRMEA-D (6-class)~\cite{66_cao2014crema}, RAVDESS~\cite{67_livingstone2018ryerson}, MER-MULTI~\cite{64_lian2023mer}, MER24-T\&V~\cite{56_lian2024mer}, and Werewolf-XL~\cite{62_zhang2021werewolf} datasets, as displayed in the tables above.

\noindent \textbf{Categorical Emotion Analysis.} We conduct comparisons with state-of-the-art CEA approaches on multiple benchmark datasets. Our analysis yields the following insight: (1) SSL-based models outperform purely supervised counterparts, as they are more effective in capturing generalized AVFA representations. (2) Audio-visual SSL models typically surpass uni-modal SSL methods by exploiting the complementary relationships between intra- and inter-modal features, thereby enhancing performance. For instance, AVF-MAE++ outperforms UniLearn~\cite{84_chen2024unilearn}, which jointly pre-trains on images and videos, by 2.33\% UAR and 1.80\% WAR on MAFW (11 classes). (3) As the model size of AVF-MAE++ increases, the improvements from Base to Large remain consistently significant across datasets. In contrast, the performance gap between Large and Huge narrows on some datasets, which is consistent with findings reported in the broader video and image domains~\cite{4_wang2023videomae,47_xie2023data}.
(4) Despite introducing the progressive semantics injection strategy to alleviate overfitting, we still observe a slight drop in performance on smaller downstream datasets (\textit{e.g.}, RAVDESS~\cite{67_livingstone2018ryerson}). This suggests that larger models are more prone to overfitting on limited data, which remains a major constraint on further improvement. In addition, we report results for 11 single-labeled emotions along with the overall metrics on the MAFW (11-class) dataset, as shown in Tab.~\ref{tab_mafw_sota_single}. The results demonstrate that our approach achieves superior performance on most categories, highlighting both its strong learning capacity and the benefits of scaling audio-visual MAE.

\noindent \textbf{Dimensional Emotion Analysis.} Following the evaluation protocol of HiCMAE~\cite{1_sun2024hicmae}, we compare AVF-MAE++ with prior approaches on two benchmark datasets. The results show that AVF-MAE++ delivers substantial improvements over existing baselines. In particular, AVF-MAE++ (H) surpasses the best reported performance on AVCAffe~\cite{61_sarkar2023avcaffe} by 4.07\% WA-F1 in Arousal and 5.46\% WA-F1 in Valence. Moreover, it achieves the highest improvement of 11.69\% PCC across dimensions on Werewolf-XL~\cite{62_zhang2021werewolf}.

\noindent \textbf{Micro-Expression Recognition.} To further demonstrate the broad applicability of our approach, we extend evaluation to the MER task. Unlike the two AVFA tasks discussed earlier, MER datasets generally lack audio modalities. Hence, we employ only the pre-trained video encoder to build a progressive fine-tuning pipeline on five representative MER benchmarks. As reported in Tab.~\ref{tab:sota-mer}, AVF-MAE++ achieves strong results, with the most notable gain of 10.21\% UF1 over HTNet~\cite{77_wang2024htnet} on CAS(ME)$^3$~\cite{71_li2022cas}, underscoring the generalization strength of the learned affective features. Moreover, on some datasets, we observe clearer performance drops when scaling from AVF-MAE++ (L) to AVF-MAE++ (H), reinforcing the overfitting tendency noted in the CEA analysis.

\begin{figure*}[!t] 
\centering
\includegraphics[width=\textwidth]{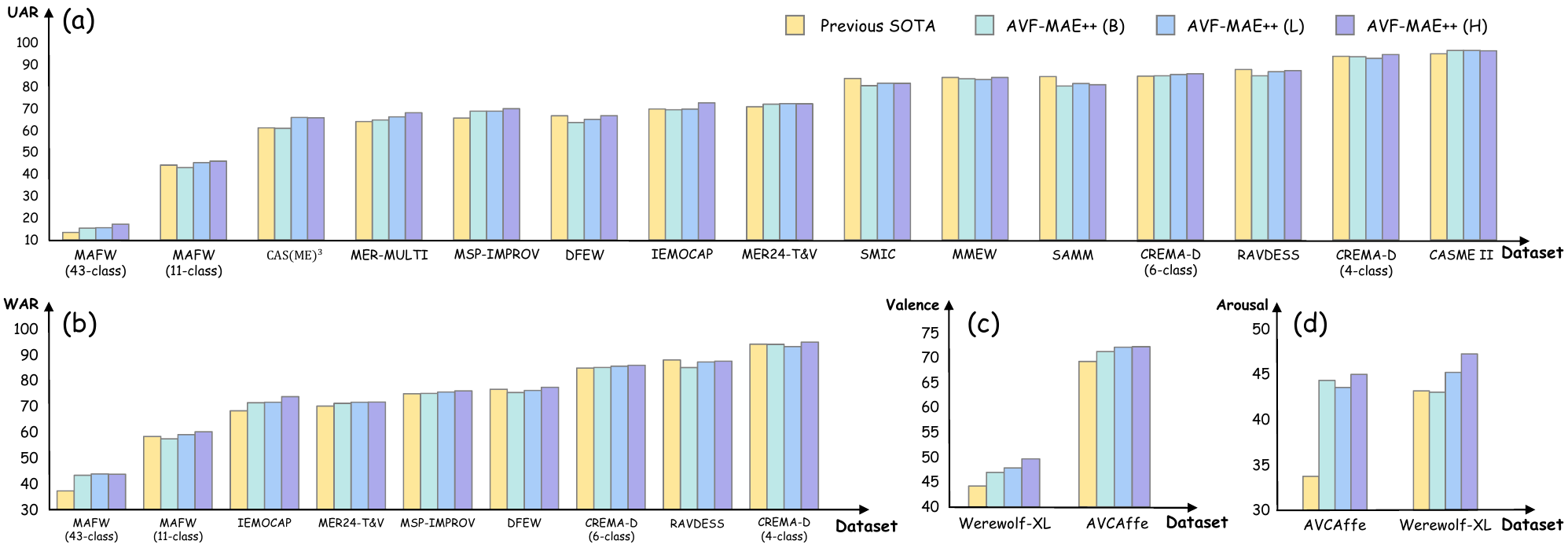}
\caption{Qualitative performance comparisons of our AVF-MAE++ across three versions and SOTA methods. Previous SOTA indicates the previously best results. (a) Comparison of UAR metric. (b) Comparison of WAR metric. (c) Comparison of Valence metric. (d) Comparison of Arousal metric.}
\label{vis}
\end{figure*}

\noindent \textbf{Qualitative Analysis of Model Performance.} To more intuitively demonstrate the performance gains brought by AVF-MAE++, we conduct a qualitative analysis by comparing three scales of AVF-MAE++ models with the state-of-the-art results on the corresponding tasks, as shown in Fig.~\ref{vis}. The overall results indicate that AVF-MAE++ achieves competitive performance across different scales, with particularly notable improvements on fine-grained emotion recognition tasks (\textit{e.g.}, MAFW-43) and continuous affective dimension prediction tasks (\textit{e.g.}, Werewolf-XL, AVCAffe). For classification metrics (UAR and WAR), AVF-MAE++ achieves consistent performance gains on mainstream multimodal emotion recognition datasets such as MER-MULTI, MSP-IMPROV, and IEMOCAP, demonstrating its adaptability and robustness in real-world scenarios. For continuous affect prediction, the model also shows significant advantages: based on the average performance across three scales, AVF-MAE++ achieves an improvement of about 10.5\% on Arousal and 3.9\% on Valence over the historical best results on Werewolf-XL, highlighting its effectiveness in modeling emotion intensity and polarity. Overall, the performance of AVF-MAE++ consistently improves with increasing model scale, further validating its generality and advancement in multimodal emotion analysis.

\begin{table}[t!]
\centering
\caption{Ablation comparisons of our dual masking \& modality encoder (\textit{i.e.}, Improved LGI-Former) with HiCMAE~\cite{1_sun2024hicmae}. We only report results in terms of WAR (\%). MER24: MER24-T\&V.}
\setlength{\abovecaptionskip}{0.3em} 
\setlength{\arrayrulewidth}{0.37pt}  
\resizebox{0.8\linewidth}{!}{%
\begin{tabular}{lcccccc}
\toprule[0.37pt]
Method  & Time & Speedup & \#PS & MAFW & MER24 \\
\midrule[0.37pt]
HiCMAE~\cite{1_sun2024hicmae}  & 115.45h & \textbf{--} & 99 & 56.17 & 70.95 \\
Dual masking + Vanilla LGI-Former  & 77.45h & \textcolor{sgreen}{\textbf{1.49$\times$}} & 142 & 55.11 & 69.42 \\
Dual masking + Improved LGI-Former  & 79.07h & \textcolor{sgreen}{\textbf{1.46$\times$}} & 163 & 56.12 & 70.36 \\
\hline
\end{tabular}
}
\label{tab-masking}
\end{table}

\begin{table}[t!]
\centering
\caption{Ablation study on the components of IAV-CL Module.}
\setlength{\abovecaptionskip}{0.3em} 
\setlength{\arrayrulewidth}{0.37pt}  
\renewcommand{\arraystretch}{0.81}   
\small
\resizebox{0.8\linewidth}{!}{%
\begin{tabular}{ccccccccc}
\toprule
\multirow{2}{*}{\raisebox{-1.3ex}{\begin{tabular}[c]{@{}c@{}}DiER \\Units\end{tabular}}} &  \multirow{2}{*}{\raisebox{-1.3ex}{\begin{tabular}[c]{@{}c@{}}HAFE \\Layer\end{tabular}}}  & \multicolumn{2}{c}{MAFW}  & \multicolumn{2}{c}{MER24-T\&V} & \multicolumn{2}{c}{IEMOCAP} \\
\cmidrule(lr){3-4} \cmidrule(lr){5-6}  \cmidrule(lr){7-8}
& &  UAR    & WAR  & WAR   & WA-F1  & UAR    & WAR  \\
\midrule
$\times$     & $\times$            & 41.89  & 56.31   & 70.21  & 69.62 & 67.46  & 69.33     \\
\checkmark   & $\times$           &  42.58   & 56.81   & 70.97  & 70.26 & 68.56  & 70.02     \\
$\times$    & \checkmark         &  42.55   & 56.77   & 70.65  & 70.13  & 68.63  & 69.86    \\
\rowcolor{cvprblue!20}
\checkmark   & \checkmark       &  \textbf{42.96}   & \textbf{57.02}   & \textbf{71.40}  & \textbf{70.72} & \textbf{68.86}  & \textbf{70.45} \\
\hline
\end{tabular}
}
\label{tab-two-subs}
\end{table}

\begin{table}[t!]
\centering
\caption{Ablation study on the stacked number of DiER Units.}
\setlength{\abovecaptionskip}{0.3em} 
\setlength{\arrayrulewidth}{0.37pt}   
\renewcommand{\arraystretch}{0.94}   
\resizebox{0.8\linewidth}{!}{%
\begin{tabular}{ccccccc}
\toprule
\multirow{2}{*}{\raisebox{-1.6ex}{\begin{tabular}[c]{@{}c@{}}Stacked \\Number\end{tabular}}} & \multicolumn{2}{c}{=~1} & \multicolumn{2}{c}{=~2} & \multicolumn{2}{c}{=~4} \\ \cmidrule(l){2-3}   \cmidrule(l){4-5}   \cmidrule(l){6-7}
& UAR & WAR & UAR & WAR & UAR & WAR  \\ \midrule
MAFW & 43.07 & 56.83 & \textbf{43.22} & \textbf{57.69} & 43.14 & 57.05  \\ 
MER24-T\&V & 61.57 & 69.98 & \textbf{62.46} & \textbf{71.09} & 62.11 & 70.83  \\ 
\hline
\end{tabular}
}
\label{tab-stacked-number}
\end{table}

\subsection{Ablation Studies}
\noindent \textbf{Influence of Dual Masking \& Improved Modality Encoder.} Tab.~\ref{tab-masking} shows the effects of our proposed audio-visual dual masking strategy and improved modality encoder on AVFA performance. In particular, we adopt AVF-MAE++ (B) for fair comparison against the audio-visual encoder-only masking scheme and the vanilla ViT~\cite{24_dosovitskiy2020image} used in HiCMAE-B~\cite{1_sun2024hicmae}, with pre-training conducted on VoxCeleb2-dev~\cite{57_chung2018voxceleb2} for 100 epochs. Results indicate that both the proposed dual masking and the improved modality encoder contribute positively, delivering 1.46$\times$ faster training while maintaining competitive accuracy.

\begin{table}[t!]
\centering
\caption{Ablation comparisons on the pre-training data scaling.}
\setlength{\abovecaptionskip}{0.3em}     
\setlength{\arrayrulewidth}{0.37pt}
\renewcommand{\arraystretch}{1.005}      
\resizebox{0.8\linewidth}{!}{%
\begin{tabular}{ccccccc}
\toprule
\raisebox{-1.8ex}{Method} & \multirow{2}{*}{\raisebox{-1.8ex}{\begin{tabular}[c]{@{}c@{}}Pre-training \\Dataset\end{tabular}}}  & \multicolumn{2}{c}{\raisebox{-0.7ex}{MAFW}} & \multicolumn{2}{c}{\raisebox{-0.7ex}{MER24-T\&V}} \\ \cmidrule(l){3-4} \cmidrule(l){5-6}  
&  & UAR   & WAR   & WAR   & WA-F1    \\ \midrule
AVF-MAE++ (B) & VoxCeleb2-dev                                & 42.80      & 56.64       & 70.93      & 70.51      \\
AVF-MAE++ (B)   & Unlabeled Hybrid   & \textbf{42.96}      & \textbf{57.02}      & \textbf{71.40}      & \textbf{70.72}     \\
$\Delta$\textcolor{sgreen}{\textit{\textbf{Metrics}}}  & \textbf{--}       & \textcolor{sgreen}{\textbf{+ {0.16}}\%}      & \textcolor{sgreen}{\textbf{+ {0.38}}\%}      & \textcolor{sgreen}{\textbf{+ {0.47}}\%}      & \textcolor{sgreen}{\textbf{+ {0.21}}\%}      \\
\hline

AVF-MAE++ (L) & VoxCeleb2-dev                                & 42.67      & 57.01       & 70.21      & 69.02      \\
AVF-MAE++ (L)   & Unlabeled Hybrid   & \textbf{43.22}      & \textbf{57.69}      & \textbf{71.09}      & \textbf{70.32}     \\
$\Delta$\textcolor{sgreen}{\textit{\textbf{Metrics}}}  & \textbf{--}       & \textcolor{sgreen}{\textbf{+ {0.55}}\%}      & \textcolor{sgreen}{\textbf{+ {0.68}}\%}      & \textcolor{sgreen}{\textbf{+ {0.88}}\%}      & \textcolor{sgreen}{\textbf{+ {1.30}}\%}      \\
\hline

AVF-MAE++ (H) & VoxCeleb2-dev                                & 43.59      & 57.22       & 70.45      & 69.42      \\
AVF-MAE++ (H)   & Unlabeled Hybrid   & \textbf{44.02}      & \textbf{57.79}      & \textbf{71.23}      & \textbf{70.41}     \\
$\Delta$\textcolor{sgreen}{\textit{\textbf{Metrics}}}  & \textbf{--}       & \textcolor{sgreen}{\textbf{+ {0.43}}\%}      & \textcolor{sgreen}{\textbf{+ {0.57}}\%}      & \textcolor{sgreen}{\textbf{+ {0.78}}\%}      & \textcolor{sgreen}{\textbf{+ {0.99}}\%}      \\

\hline
\end{tabular}
}
\vspace{1em}
\label{tab-data-scaling}
\end{table}

\begin{table*}[t!]
\centering
\caption{The comprehensive ablation studies for our introduced AVF-MAE++.}
\setlength{\abovecaptionskip}{0.3em}
\setlength{\arrayrulewidth}{0.37pt}
\renewcommand{\arraystretch}{1.065}
\resizebox{0.96\linewidth}{!}{%
\begin{tabular}{lcccccccc}
\toprule
\raisebox{-2.3ex}{Method} & \multirow{2}{*}{\raisebox{-2.6ex}{\begin{tabular}[c]{@{}c@{}}Pre-training Dataset\end{tabular}}} & \multirow{2}{*}{\raisebox{-2.6ex}{\begin{tabular}[c]{@{}c@{}}Pre-training Time\end{tabular}}} & \multirow{2}{*}{\raisebox{-2.6ex}{\begin{tabular}[c]{@{}c@{}}MAFW \\WAR (\%)\end{tabular}}} & \multirow{2}{*}{\raisebox{-2.6ex}{\begin{tabular}[c]{@{}c@{}}MER24-T\&V \\WAR (\%)\end{tabular}}} \\ 
\\ \midrule
HiCMAE-B (Baseline) & VoxCeleb2-dev               & 115.45h                       & 56.17       & 70.95          \\

HiCMAE-B + Dual Masking & VoxCeleb2-dev               & 81.46h (\textcolor{sgreen}{1.42$\times$})                       & 54.63       & 68.93          \\ 

Dual masking + Vanilla LGI-Former   & VoxCeleb2-dev & 77.45h (\textcolor{sgreen}{1.49$\times$})        & 55.11      & 69.42           \\
Dual masking + Improved LGI-Former   & VoxCeleb2-dev  & 79.07h (\textcolor{sgreen}{1.46$\times$})       & 56.12      & 70.36           \\
\hline

+ Pre-training Data Scaling & Unlabeled Hybrid       & 84.23h (\textcolor{sgreen}{1.37$\times$})                               & 56.47       & 70.67            \\
+ IAV-CL Module   & Unlabeled Hybrid & 85.87h (\textcolor{sgreen}{1.34$\times$})       & 57.02      & 71.40           \\
+ PSI Strategy \{\textit{i.e.}, \textbf{AVF-MAE++ (B)}\} & Unlabeled Hybrid     & 85.87h (\textcolor{sgreen}{1.34$\times$})                                 & \textbf{57.50} (\textcolor[RGB]{192,0,0}{+1.33})       & \textbf{72.11} (\textcolor[RGB]{192,0,0}{+1.16})            \\
\hline

\multirow{2}{*}{\raisebox{-1.68ex}{\begin{tabular}[c]{@{}c@{}}\textcolor{sgreen}{Model} \\\textcolor{sgreen}{Scaling}\end{tabular}}} \textbf{AVF-MAE++ (L)}   & Unlabeled Hybrid & 88.58h (\textcolor{sgreen}{1.30$\times$})        & \textbf{59.13} (\textcolor[RGB]{192,0,0}{+2.96})      & \textbf{72.33} (\textcolor[RGB]{192,0,0}{+1.38})          \\

~~~~~~~~~~\textbf{AVF-MAE++ (H)}   & Unlabeled Hybrid  & 93.52h~(\textcolor{sgreen}{1.23$\times$})       & \textbf{60.24} (\textcolor[RGB]{192,0,0}{+4.07})      & \textbf{72.28} (\textcolor[RGB]{192,0,0}{+1.33})          \\

\hline
\end{tabular}
}
\label{tab-overall-ablation}
\vspace{1em}
\end{table*}

\begin{figure}[t!]
\setlength{\belowcaptionskip}{-1.3em}
\centering
\includegraphics[width=0.85\linewidth]{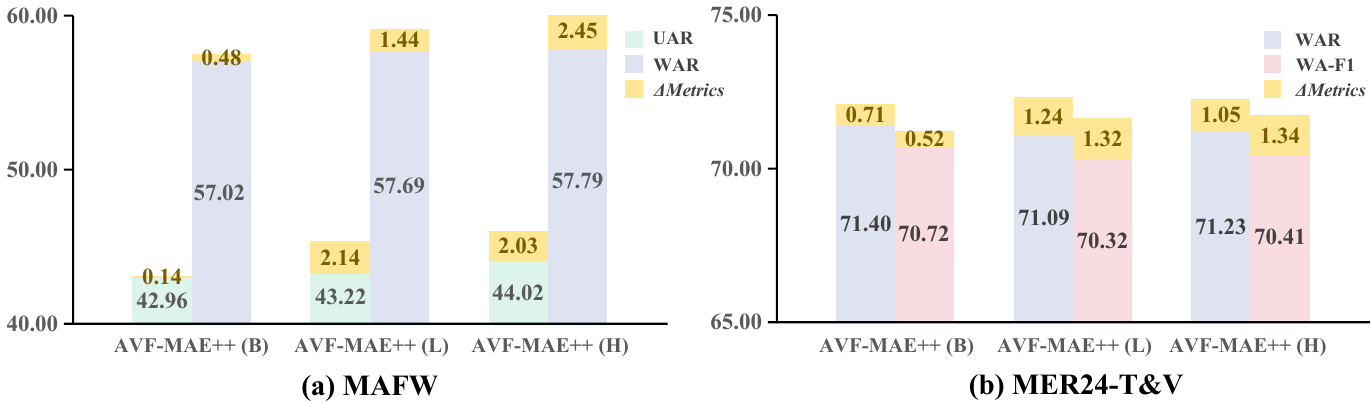}
\vspace{-0.6em}
\caption{Ablation explorations on the progressive training.}
\label{fig-PSI-ablation}
\end{figure}

\noindent \textbf{Evaluation on components of IAV-CL Module.} We assess the contributions of the DiER Units and HAFE Layer in the IAV-CL Module using AVF-MAE++ (B), as reported in Tab.~\ref{tab-two-subs}. The models are first pre-trained on the constructed hybrid dataset and subsequently fine-tuned on IEMOCAP~\cite{65_busso2008iemocap}. For the ablation on the HAFE Layer, we replace it with the original fusion components from HiCMAE~\cite{1_sun2024hicmae} to realize hierarchical integration. Results in Tab.~\ref{tab-two-subs} demonstrate that combining DiER Units with the HAFE Layer yields the greatest performance gains, confirming their effectiveness in modeling intra- and inter-modal correlations.

\noindent \textbf{Ablation Study on the Number of DiER Units.} To explore the optimal depth of DiER Unit stacking, we perform ablation experiments with varying numbers of units using AVF-MAE++ (L), as shown in Tab.~\ref{tab-stacked-number}. The findings reveal that performance does not scale linearly with more layers; excessive stacking introduces overly dense interactions, leading to higher complexity and reduced stability.

\noindent \textbf{Effectiveness on Data Scaling.} As illustrated in Tab.~\ref{tab-data-scaling}, we evaluate the impact of scaling pre-training data on AVF-MAE++ with VoxCeleb2-dev~\cite{57_chung2018voxceleb2} and our unlabeled hybrid dataset. The results show that scaling data consistently improves all metrics, highlighting the crucial role of dataset size and diversity in AVFA masked autoencoding.

\noindent \textbf{Contribution of the PSI Strategy.} We analyze the impact of the proposed PSI strategy, as shown in Fig.~\ref{fig-PSI-ablation}. The results reveal that AVF-MAE++ achieves enhanced performance, confirming its effectiveness in facilitating a smooth transition from pre-training to fine-tuning.

\noindent \textbf{Overall ablation studies.} We conduct a comprehensive set of ablation experiments to systematically examine the contributions of each proposed improvement under fair conditions, as summarized in Tab.~\ref{tab-overall-ablation}. The results clearly indicate that every component brings positive gains to the overall framework. Most of the parameter growth originates from LGI-Former, which increases attention parameters compared to the Vanilla ViT~\cite{24_dosovitskiy2020image}, yet effectively lowers FLOPs, consistent with observations in~\cite{2_sun2023mae}. With our careful design, notable acceleration in pre-training is also achieved, as reported in Tab.~\ref{tab-overall-ablation}.

\subsection{Uni-modal Comparison Results}
We additionally report uni-modal results of AVF-MAE++ (H) on three representative CEA and DEA datasets, as shown in Tab.~\ref{tab-uni-modal}. For more detailed comparisons, please refer to Tab.~\ref{tab_mafw_sota_single}, \ref{tab_cremad_sota_six_classes}, and \ref{tab_werewolf-xl_sota}. The findings show that our uni-modal models consistently deliver competitive performance, validating the effectiveness of the proposed model design.

\begin{table*}[t]
\centering
\caption{Results of uni-modal AVF-MAE++ across three representative CEA and DEA datasets. Note that we average the metrics for the three dimensions of Werewolf-XL~\cite{62_zhang2021werewolf} dataset.}
\setlength{\abovecaptionskip}{0.3em}
\renewcommand{\arraystretch}{1.0}
\setlength{\arrayrulewidth}{0.22pt}
\small
\scalebox{5.2}{}{%
\begin{tabular}{lccccccc}
\toprule[0.22pt]
\multirow{2}{*}{\makecell{Method}} & \multirow{2}{*}{\makecell{Modality}}  & \multicolumn{2}{c}{MAFW (11-class)} & \multicolumn{2}{c}{CREMA-D (6-class)} & \multicolumn{1}{c}{Werewolf-XL} \\
\cmidrule[0.22pt](l){3-4}   \cmidrule[0.22pt](l){5-6}   \cmidrule[0.22pt](l){7-7} 
                                &  & UAR        & WAR        & UAR        & WAR         & Average              \\ \midrule[0.22pt]

AVF-MAE++~(H)                   & Audio       & 27.15        & 38.78         & 73.51           & 73.02         & 33.22             \\
AVF-MAE++~(H)             & Video            & 42.24        & 55.61         & 79.23            & 79.97         & 45.38       \\
\hline
\end{tabular}
}
\label{tab-uni-modal}
\end{table*}

\begin{table}[t!]
\centering
\caption{The results of our cross-dataset study in terms of WAR metric.}
\setlength{\abovecaptionskip}{0.3em} 
\setlength{\arrayrulewidth}{0.40pt}  
\renewcommand\arraystretch{1.1}      
\begin{tabular}{l|ccc}
\toprule
Source Dataset $\rightarrow$ Targeted Dataset & AVF-MAE++ & HiCMAE & MAE-DFER \\
\hline
MSP-IMPROV $\rightarrow$ CREMA-D (4-class) & 94.53 & 93.21 & 90.88 \\
RAVDESS $\rightarrow$ IEMOCAP & 70.95 & 68.92 & 66.10 \\
\bottomrule
\end{tabular}
\label{tab:cross-dataset-study}
\end{table}

\subsection{Cross-Dataset Studies}
We perform cross-dataset evaluations using AVF-MAE++ (B), HiCMAE~\cite{1_sun2024hicmae}, and MAE-DFER~\cite{2_sun2023mae}, models trained on the MSP-IMPROV and RAVDESS datasets, and then test them on the CREMA-D (4-class) and IEMOCAP datasets, respectively.  As shown in Tab.~\ref{tab:cross-dataset-study}, all three models exhibit reasonable generalization ability, while our AVF-MAE++ consistently achieves the best performance in both transfer scenarios, obtaining 94.53\% WAR on CREMA-D and 70.95\% WAR on IEMOCAP. In the MSP-IMPROV → CREMA-D cross-dataset experiment (trained on MSP-IMPROV, tested on CREMA-D), AVF-MAE++ improves performance by +1.32\% over HiCMAE and +3.65\% over MAE-DFER. Likewise, in the RAVDESS → IEMOCAP experiment (trained on RAVDESS, tested on IEMOCAP), our method obtains +2.03\% and +4.85\% gains against HiCMAE and MAE-DFER, respectively. These results demonstrate that AVF-MAE++ not only generalizes well across datasets but also provides clear improvements over prior MAE-based methods, highlighting the robustness and transferability of the learned AVFA representations. For more detailed performance comparisons, please refer to Tab.~\ref{tab_cremad_sota_four_cls} and Tab.~\ref{tab_iemocap_sota}.

\subsection{Qualitative Analysis}

\begin{figure*}[t!]
\setlength{\belowcaptionskip}{-1.5em}
\centering
\includegraphics[width=\textwidth]{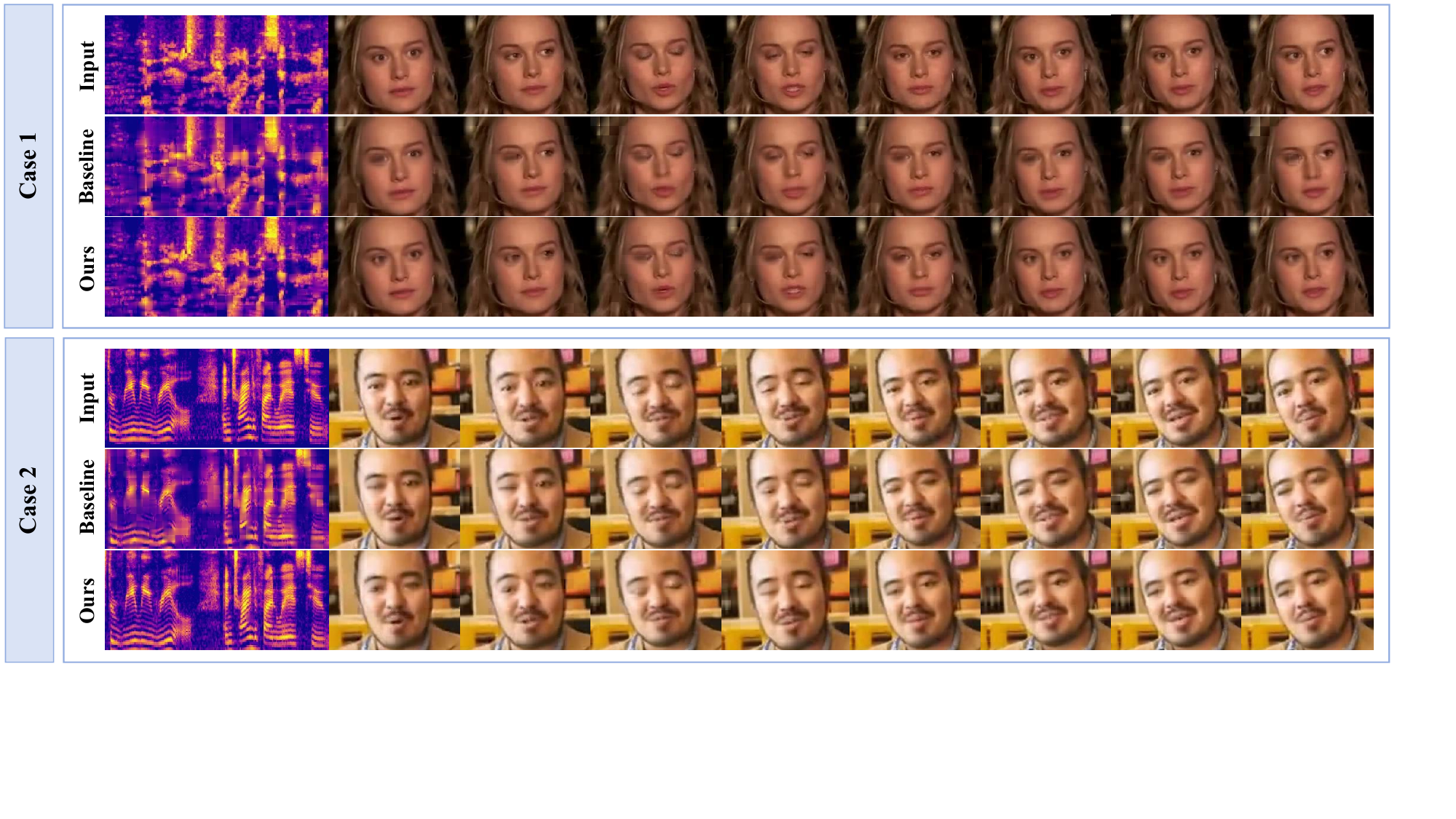}
\vspace{-6.2em}
\caption{The detailed qualitative visualizations of the overall audio-visual reconstructions of our proposed AVF-MAE++ and baseline method.} 
\label{fig-overall_reconstruction}
\vspace{1em}
\end{figure*}

\noindent \textbf{Audio-visual reconstruction visualizations.}~Fig.~\ref{fig-overall_reconstruction} shows the qualitative visualizations of overall audio-visual reconstructions. It can be observed that AVF-MAE++ exhibits superior capability in detail preservation, particularly in capturing the dynamic variations and continuity of micro-expressions. This indicates that the model is not only able to capture subtle temporal transitions of facial expressions but also maintain high consistency in local regions such as the eyes and lips along the spatial dimension. From these outcomes, we can conclude that our method can capture crucial factors of facial emotional expressions more effectively, such as eyes and lip movements, promoting discriminative AVFA representations learning, which leads to better reconstruction quality.

\begin{figure*}[t!]
\setlength{\belowcaptionskip}{-1.5em}
\centering
\includegraphics[width=\textwidth]{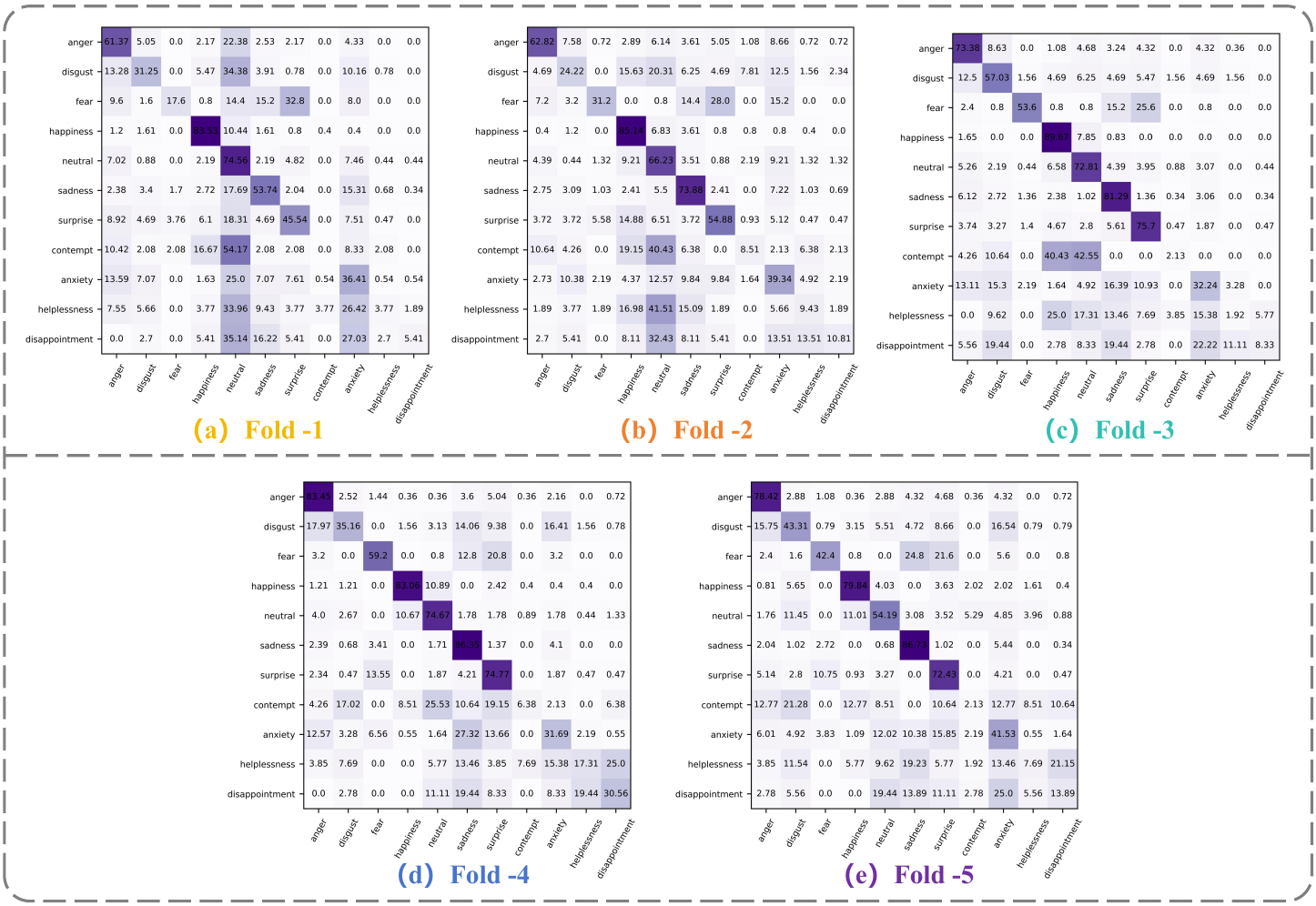}
\vspace{-1.4em}
\caption{The confusion matrices of AVF-MAE++ (H) for five folds on MAFW (11-class) dataset.}
\label{fig-MAFW-Fold5}
\vspace{1.0em}
\end{figure*}

\noindent \textbf{Confusion Matrices.} As shown in Fig.~\ref{fig-MAFW-Fold5}, we present the detailed confusion matrices of the proposed AVF-MAE++ (H) across five folds on the MAFW (11-class) dataset. The results demonstrate that our model consistently delivers stable performance and maintains good balance across all categories. Notably, even under the challenging conditions of class imbalance and high inter-class similarity, AVF-MAE++ is still able to effectively distinguish different emotion categories, exhibiting a relatively low confusion rate. This indicates that the learned multi-modal representations possess not only strong discriminative power but also good generalization ability across categories. Overall, these findings further highlight the robustness and generalizability of our model, thereby providing strong evidence of the effectiveness of our design.

\subsection{Downstream Application Explorations of AVF-MAE++}
AVF-MAE++ is not only an effective pretraining framework but also demonstrates excellent scalability across various downstream tasks. To be specific, we highlight its effectiveness across three representative applications:

\noindent\textbf{Video-level Deepfake Detection.}
We extend the idea of AVF-MAE++ to the deepfake detection task. Specifically, we first perform self-supervised training of the audio and video encoders using AVF-MAE++, and then fine-tune them on our subsequent deepfake detection framework \cite{wu2025hola}. As shown in Tab.~\ref{tab:main-compare}, our method achieves 98.72\% accuracy on video-level deepfake detection, outperforming the second-best method (TimeSformer, ACC = 86.04\%) by 12.68\%. Moreover, it attains the best performance in terms of WA-F1 (98.54\%), UAR (98.22\%), and AUC (0.9983). Compared with large vision-language models, our approach also demonstrates superior performance under the same evaluation setting, as shown in Tab.~\ref{tab:main-compare-mllms}. These results highlight the strong potential and broad applicability of AVF-MAE++ in deepfake detection.

\begin{table}[t!]
\centering
\caption{Performance comparisons of advanced baselines and AVF-MAE++ in terms of ACC, WA-F1, UAR, and AUC metrics on the video-level deepfake detection track. Note that we highlight the best performance in bold and underline the second performance.}
\vspace{-0.5em} 
\setlength{\arrayrulewidth}{0.40pt}
\renewcommand{\arraystretch}{1.25} 
\resizebox{\linewidth}{!}{%
\begin{tabular}{lcccccccc}
\toprule
Method & Venue & ACC~(\%) & WA-F1~(\%) & UAR~(\%) & AUC \\ 
\midrule
ECAPA-TDNN~\cite{new_32_desplanques2020ecapa} & Interspeech & 80.04 & 77.14 & 65.37 & 0.8569\\
Res2Net~\cite{34_gao2019res2net} & TPAMI & 83.10 & 82.78 & 75.88 & 0.9059 \\
CAM++~\cite{38_wang2023cam++}  & Interspeech & 74.52 & 64.28 & 50.81 & 0.5182 \\
ResNet-SE~\cite{hu2018squeeze} & CVPR & 81.88 & 81.35 & 74.25 & 0.8942 \\
PANNS~\cite{27_kong2020panns}  & TASLP & 82.39 & 80.70 & 70.00 & \underline{0.9102} \\

I3D~\cite{43_carreira2017quo}  &  CVPR & 85.00 & 84.76 & \underline{78.97} & 0.9058 \\

C3D~\cite{32_tran2015learning} &  CVPR & 74.31 & 64.03 & 50.63 & 0.6506 \\

Video Swin-T~\cite{19_liu2022video} & CVPR & 76.61 & 68.71 & 54.88 & 0.7560 \\

TimeSformer~\cite{30_bertasius2021space}  & ICML & \underline{86.04} & \underline{85.39} & 77.93 & 0.9097 \\

\rowcolor{gray!20}
\textbf{AVF-MAE++~(ours)} & \textbf{--} & \textbf{98.72} & \textbf{98.54} & \textbf{98.22} & \textbf{0.9983} \\
\hline
\end{tabular}
}
\label{tab:main-compare}
\end{table}

\begin{table}[t!]
\centering
\caption{Performance comparisons of advanced VLMs in terms of ACC, WA-F1, and UAR on video-level deepfake detection.}
\vspace{-0.5em} 
\setlength{\arrayrulewidth}{0.40pt}
\renewcommand{\arraystretch}{1.20} 
\resizebox{\linewidth}{!}{%
\begin{tabular}{lcccccccc}
\toprule
Method & Organization & ACC~(\%) & WA-F1~(\%) & UAR~(\%)  \\ 
\midrule

o3-mini-high~\cite{openai2025o3mini} & OpenAI & 46.06 & 48.34 & 50.60 \\ 
Gemini-2.5-pro~\cite{Gemini-2.5-pro-preview-03-25} & Google & 73.30 & 62.01 & 50.00 \\
Gemini-2.5-pro-thinking~\cite{Gemini-2.5-pro-preview-03-25} & Google & \textbf{73.88} & \textbf{62.39} & \textbf{51.00} \\
Claude-3.7-Sonnet-250219 & Anthropic & 29.03 & 18.59 & 49.60 \\
GPT-4o-240513~\cite{openai2024gpt4o} & OpenAI & 60.80 & 59.61 & 47.31\\ 

QVQ-Max-250325~\cite{Qwen-QVQ-72B}  & Alibaba & \underline{73.32} & \underline{62.03} & 50.00  \\ 

Qwen2.5-VL-32B-Instruct~\cite{Qwen2d5-vl}  & Alibaba & 32.40 & 25.79 & 50.55  \\ 
Qwen2.5-VL-72B-Instruct~\cite{Qwen2d5-vl}  & Alibaba & 30.90 & 22.53 & 50.25  \\ 
Grok 4~\cite{grok-4} & xAI & 73.30 & 62.01 & 50.00  \\
Phi-3.5-Vision-Instruct~\cite{48_abdin2024phi} & Microsoft & 50.70 & 53.51 & 48.63  \\
Doubao-Seed-1.6-250615~\cite{Doubao-1.5-vision-pro-32k} & ByteDance & 30.10 & 19.93 & \underline{50.77} \\

\hline
\rowcolor{gray!20}
\textbf{AVF-MAE++} (Reference) & \textbf{--} & 98.72 & 98.54 & 98.22 \\
\hline
\end{tabular}
}
\label{tab:main-compare-mllms}
\end{table}

\noindent\textbf{Audio-Visual Talking Head Generation.}
To explore the impact of AVF-MAE++ on the talking head generation task, we integrate it into the StableAvatar \cite{stableavatar} framework. Specifically, we replace the original Wav2vec 2.0 with the audio encoder pretrained by AVF-MAE++, enabling more robust and semantically rich audio representations. In addition, we introduce the vision encoder of AVF-MAE++ as an independent branch to extract frame-level visual features, which are fused with audio embeddings through the audio adapter. As shown in Tab.~\ref{tab:talking_face_generation}, our method achieves the best performance across three datasets on both FID and FVD, reducing FID by an average of 4.5\% and FVD by 4.3\% compared to StableAvatar. These results demonstrate significant improvements in generation quality and temporal coherence, further validating the scalability and effectiveness of AVF-MAE++ for talking head generation.

\begin{table}[t!]
\centering
\caption{Comparison of different methods on HDTF, AVSpeech, and Long100 datasets in terms of FID and FVD.}
\label{tab:talking_face_generation}
\begin{tabular}{lcccccc}
\toprule
\multirow[c]{2}{*}{Method} & \multicolumn{2}{c}{HDTF} & \multicolumn{2}{c}{AVSpeech} & \multicolumn{2}{c}{Long100} \\
\cmidrule(lr){2-3} \cmidrule(lr){4-5} \cmidrule(lr){6-7}
 & FID $\downarrow$ & FVD $\downarrow$ & FID $\downarrow$ & FVD $\downarrow$ & FID $\downarrow$ & FVD $\downarrow$ \\
\midrule
SadTalker~\cite{zhang2022sadtalker} & 53.12 & 567 & 120.57 & 1468 & 194.88 & 2261 \\
Aniportrait~\cite{wei2024aniportrait} & 46.35 & 537 & 118.86 & 1524 & 190.66 & 2095 \\
Sonic~\cite{Ji_sonic} & 62.17 & 552 & 187.42 & 2051 & 278.40 & 2877 \\
EchoMimic~\cite{echomimic} & 63.43 & 593 & 108.13 & 1123 & 178.12 & 1885 \\
Hallo3~\cite{hallo3} & 44.31 & 438 & 98.14 & 987 & 170.44 & 1724 \\
FantasyTalking~\cite{wang2025fantasytalking} & 46.74 & 479 & 80.01 & 823 & 175.78 & 1789 \\
HunyuanAvatar~\cite{hunyuanavatar} & 52.16 & 625 & 77.53 & 868 & 172.94 & 1743 \\
MultiTalk~\cite{multitalk} & 46.94 & 446 & 75.67 & 804 & 175.52 & 1768 \\
OmniAvatar~\cite{ominiavatar} & 41.79 & 424 & 72.56 & 744 & 168.49 & 1621 \\
StableAvatar~\cite{stableavatar} & \underline{38.14} & \underline{375} & \underline{68.12} & \underline{640} & \underline{57.18} & \underline{504} \\
StableAvatar w/ AVF-MAE++ & \textbf{37.84} & \textbf{369} & \textbf{68.03} & \textbf{639} & \textbf{56.22} & \textbf{498} \\
\hline
\end{tabular}
\end{table}

\begin{table*}[t!]
\centering
\caption{Performance comparisons of MLLMs with and without our AVF-MAE++ in terms of the ACC metric across four MER datasets.}
\resizebox{\textwidth}{!}{%
\begin{tabular}{lccccc}
\toprule
Method & Open-source & CASME II & \text{CAS(ME)}$^3$ & DFME TestA & DFME TestB \\
\midrule
Gemini-2.5-Pro \cite{Gemini-2.5-pro-preview-03-25} & $\times$ & 30.23 & 20.92 & 34.81 & 28.43 \\
\hline
Qwen2.5-VL-7B \cite{Qwen2d5-vl} & \checkmark & 16.28 & 25.07 & 20.46 & 16.05 \\
Qwen2.5-VL-7B (w/ AVF-MAE++) & \checkmark & 20.04 & 28.33 & 22.01 & 19.20 \\
\hline
Qwen-VL-Max \cite{Qwen-VL} & \checkmark & 10.08 & 19.05 & 20.25 & 16.72 \\
Qwen-VL-Max (w/ AVF-MAE++) & \checkmark & 15.32  & 23.01 & 23.08 & 18.31 \\
\hline
\end{tabular}%
}
\label{tab:mllm_benefit}
\end{table*}

\noindent\textbf{Benefiting MLLMs in Micro-Expression Recognition (MER).} 
To validate the effectiveness of AVF-MAE++, we integrate it as an auxiliary expert module into representative MLLMs for the challenging micro-expression recognition task. As shown in Tab.~\ref{tab:mllm_benefit}, AVF-MAE++ consistently improves baseline models across four benchmark datasets. Qwen2.5-VL-7B achieves an average accuracy gain of +2.93\% when equipped with AVF-MAE++, with notable improvements of +3.76\% on CASME II and +3.15\% on DFME TestB. Similarly, Qwen-VL-Max obtains an average improvement of +3.41\%, including significant gains of +5.24\% on CASME II and +3.96\% on CAS(ME)$^3$. These substantial improvements demonstrate that AVF-MAE++ effectively enhances multi-scale video feature modeling and addresses cross-dataset domain gaps in MER. While Gemini-2.5 Pro remains the strongest standalone model on certain benchmarks, the integration of AVF-MAE++ enables open-source models to approach or even surpass proprietary systems, highlighting the complementary value of our approach.

\section{Conclusion and Discussions}
\label{sec:Conclusions}
In this paper, we explore the scaling properties of audio-visual masked autoencoders (MAE) for efficient affective video facial analysis (AVFA). By introducing the dual masking strategy, refined model architecture, and a progressive training pipeline, we successfully develop AVF-MAE++, the first hundred-million audio-visual MAE series tailored for the affective computing domain. Comprehensive experiments across 17 datasets verify the superiority of our proposed AVF-MAE++ series models, highlighting that audio-visual MAEs can serve as scalable and versatile representation learners for AVFA.

Nevertheless, several challenges remain. Even with the progressive training pipeline, overfitting persists on smaller datasets, limiting the potential performance improvements. Moreover, scaling up model capacity shows diminishing returns on certain benchmarks. In addition, the data scale we investigate is still modest compared to large-scale general domains~\cite{47_xie2023data}, leaving the training of audio-visual MAEs on billions of AVFA videos unexplored. We identify these as key directions for future research and encourage further explorations from the research community.

\printbibliography

@String(CVPR= {IEEE Conf. Comput. Vis. Pattern Recog.})

@String(ICPR = {Int. Conf. Pattern Recog.})

@String(BMVC= {Brit. Mach. Vis. Conf.})

@String(ICME = {Int. Conf. Multimedia and Expo})

@String(ICASSP=	{ICASSP})

@String(ICIP = {IEEE Int. Conf. Image Process.})

@String(ACCV  = {ACCV})

@String(ICLR = {Int. Conf. Learn. Represent.})

@String(AAAI = {AAAI})

@String(CVPR  = {CVPR})

@String(ICPR  = {ICPR})

@String(BMVC  =	{BMVC})

@String(ICME  =	{ICME})

@String(ICIP  = {ICIP})

@String(ICLR  = {ICLR})

@article{1_sun2024hicmae,
  title={HiCMAE: Hierarchical Contrastive Masked Autoencoder for Self-Supervised Audio-Visual Emotion Recognition},
  author={Sun, Licai and Lian, Zheng and Liu, Bin and Tao, Jianhua},
  journal={Information Fusion},
  volume={108},
  pages={102382},
  year={2024},
  publisher={Elsevier}
}

@inproceedings{2_sun2023mae,
  title={Mae-dfer: Efficient masked autoencoder for self-supervised dynamic facial expression recognition},
  author={Sun, Licai and Lian, Zheng and Liu, Bin and Tao, Jianhua},
  booktitle={Proceedings of the 31st ACM International Conference on Multimedia},
  pages={6110--6121},
  year={2023}
}

@article{3_sun2024svfap,
  title={SVFAP: Self-supervised video facial affect perceiver},
  author={Sun, Licai and Lian, Zheng and Wang, Kexin and He, Yu and Xu, Mingyu and Sun, Haiyang and Liu, Bin and Tao, Jianhua},
  journal={IEEE Transactions on Affective Computing},
  year={2024},
  publisher={IEEE}
}

@inproceedings{4_wang2023videomae,
  title={Videomae v2: Scaling video masked autoencoders with dual masking},
  author={Wang, Limin and Huang, Bingkun and Zhao, Zhiyu and Tong, Zhan and He, Yinan and Wang, Yi and Wang, Yali and Qiao, Yu},
  booktitle={Proceedings of the IEEE/CVF Conference on Computer Vision and Pattern Recognition},
  pages={14549--14560},
  year={2023}
}

@article{5_wu2023emotions,
  title={eMotions: A Large-Scale Dataset for Emotion Recognition in Short Videos},
  author={Wu, Xuecheng and Sun, Heli and Xue, Junxiao and Zhai, Ruofan and Kong, Xiangyan and Nie, Jiayu and He, Liang},
  journal={arXiv preprint arXiv:2311.17335},
  year={2023}
}

@article{6_zong2024self,
  title={Self-Supervised Multimodal Learning: A Survey},
  author={Zong, Yongshuo and Mac Aodha, Oisin and others},
  journal={IEEE Transactions on Pattern Analysis and Machine Intelligence},
  year={2024},
  publisher={IEEE}
}

@article{7_tong2022videomae,
  title={Videomae: Masked autoencoders are data-efficient learners for self-supervised video pre-training},
  author={Tong, Zhan and Song, Yibing and Wang, Jue and Wang, Limin},
  journal={Advances in neural information processing systems},
  volume={35},
  pages={10078--10093},
  year={2022}
}

@inproceedings{9_jiang2020dfew,
  title={Dfew: A large-scale database for recognizing dynamic facial expressions in the wild},
  author={Jiang, Xingxun and Zong, Yuan and Zheng, Wenming and Tang, Chuangao and Xia, Wanchuang and Lu, Cheng and Liu, Jiateng},
  booktitle={Proceedings of the 28th ACM international conference on multimedia},
  pages={2881--2889},
  year={2020}
}

@article{10_cheng2024emotion,
  title={Emotion-LLaMA: Multimodal Emotion Recognition and Reasoning with Instruction Tuning},
  author={Cheng, Zebang and Cheng, Zhi-Qi and He, Jun-Yan and Sun, Jingdong and Wang, Kai and Lin, Yuxiang and Lian, Zheng and Peng, Xiaojiang and Hauptmann, Alexander},
  journal={arXiv preprint arXiv:2406.11161},
  year={2024}
}

@inproceedings{11_wang2022one,
  title={One-shot talking face generation from single-speaker audio-visual correlation learning},
  author={Wang, Suzhen and Li, Lincheng and Ding, Yu and Yu, Xin},
  booktitle={Proceedings of the AAAI Conference on Artificial Intelligence},
  volume={36},
  number={3},
  pages={2531--2539},
  year={2022}
}

@article{13_lian2023explainable,
  title={Explainable multimodal emotion reasoning},
  author={Lian, Zheng and Sun, Licai and Xu, Mingyu and Sun, Haiyang and Xu, Ke and Wen, Zhuofan and Chen, Shun and Liu, Bin and Tao, Jianhua},
  journal={arXiv preprint arXiv:2306.15401},
  year={2023}
}

@article{14_pantic2000automatic,
  title={Automatic analysis of facial expressions: The state of the art},
  author={Pantic, Maja and Rothkrantz, Leon J. M.},
  journal={IEEE Transactions on pattern analysis and machine intelligence},
  volume={22},
  number={12},
  pages={1424--1445},
  year={2000},
  publisher={IEEE}
}

@incollection{16_mehrabian2017communication,
  title={Communication without words},
  author={Mehrabian, Albert},
  booktitle={Communication theory},
  pages={193--200},
  year={2017},
  publisher={Routledge}
}

@article{17_wu2014survey,
  title={Survey on audiovisual emotion recognition: databases, features, and data fusion strategies},
  author={Wu, Chung-Hsien and Lin, Jen-Chun and Wei, Wen-Li},
  journal={APSIPA transactions on signal and information processing},
  volume={3},
  pages={e12},
  year={2014},
  publisher={Cambridge University Press}
}

@article{18_zhang2023transformer,
  title={Transformer-based multimodal emotional perception for dynamic facial expression recognition in the wild},
  author={Zhang, Xiaoqin and Li, Min and Lin, Sheng and Xu, Hang and Xiao, Guobao},
  journal={IEEE Transactions on Circuits and Systems for Video Technology},
  year={2023},
  publisher={IEEE}
}

@article{19_zhang2017learning,
  title={Learning affective features with a hybrid deep model for audio--visual emotion recognition},
  author={Zhang, Shiqing and Zhang, Shiliang and Huang, Tiejun and Gao, Wen and Tian, Qi},
  journal={IEEE transactions on circuits and systems for video technology},
  volume={28},
  number={10},
  pages={3030--3043},
  year={2017},
  publisher={IEEE}
}

@article{20_hossain2019emotion,
  title={Emotion recognition using deep learning approach from audio--visual emotional big data},
  author={Hossain, M Shamim and Muhammad, Ghulam},
  journal={Information Fusion},
  volume={49},
  pages={69--78},
  year={2019},
  publisher={Elsevier}
}

@inproceedings{21_zhang2024mart,
  title={Mart: Masked affective representation learning via masked temporal distribution distillation},
  author={Zhang, Zhicheng and Zhao, Pancheng and Park, Eunil and Yang, Jufeng},
  booktitle={Proceedings of the IEEE/CVF Conference on Computer Vision and Pattern Recognition},
  pages={12830--12840},
  year={2024}
}

@inproceedings{22_he2022masked,
  title={Masked autoencoders are scalable vision learners},
  author={He, Kaiming and Chen, Xinlei and Xie, Saining and Li, Yanghao and Doll{\'a}r, Piotr and Girshick, Ross},
  booktitle={Proceedings of the IEEE/CVF conference on computer vision and pattern recognition},
  pages={16000--16009},
  year={2022}
}

@article{23_bao2021beit,
  title={Beit: Bert pre-training of image transformers},
  author={Bao, Hangbo and Dong, Li and Piao, Songhao and Wei, Furu},
  journal={arXiv preprint arXiv:2106.08254},
  year={2021}
}

@article{24_dosovitskiy2020image,
  title={An Image is Worth 16x16 Words: Transformers for Image Recognition at Scale},
  author={Dosovitskiy, Alexey and Beyer, Lucas and Kolesnikov, Alexander and Weissenborn, Dirk and Zhai, Xiaohua and Unterthiner, Thomas and  Dehghani, Mostafa and Minderer, Matthias and Heigold, Georg and Gelly, Sylvain and Uszkoreit, Jakob and Houlsby, Neil},
  journal={ICLR},
  year={2021}
}

@inproceedings{25_chumachenko2024mma,
  title={MMA-DFER: MultiModal Adaptation of unimodal models for Dynamic Facial Expression Recognition in-the-wild},
  author={Chumachenko, Kateryna and Iosifidis, Alexandros and Gabbouj, Moncef},
  booktitle={Proceedings of the IEEE/CVF Conference on Computer Vision and Pattern Recognition},
  pages={4673--4682},
  year={2024}
}

@article{26_zhao2023prompting,
  title={Prompting visual-language models for dynamic facial expression recognition},
  author={Zhao, Zengqun and Patras, Ioannis},
  journal={arXiv preprint arXiv:2308.13382},
  year={2023}
}

@inproceedings{27_zhang2023weakly,
  title={Weakly supervised video emotion detection and prediction via cross-modal temporal erasing network},
  author={Zhang, Zhicheng and Wang, Lijuan and Yang, Jufeng},
  booktitle={Proceedings of the IEEE/CVF Conference on Computer Vision and Pattern Recognition},
  pages={18888--18897},
  year={2023}
}

@article{28_goncalves2024versatile,
  title={Versatile Audio-Visual Learning for Emotion Recognition},
  author={Goncalves, Lucas and Leem, Seong-Gyun and Lin, Wei-Cheng and Sisman, Berrak and Busso, Carlos},
  journal={IEEE Transactions on Affective Computing},
  year={2024},
  publisher={IEEE}
}

@article{29_ma2019audio,
  title={Audio-visual emotion fusion (AVEF): A deep efficient weighted approach},
  author={Ma, Yaxiong and Hao, Yixue and Chen, Min and Chen, Jincai and Lu, Ping and Ko{\v{s}}ir, Andrej},
  journal={Information Fusion},
  volume={46},
  pages={184--192},
  year={2019},
  publisher={Elsevier}
}

@article{30_wang2023cam++,
  title={Cam++: A fast and efficient network for speaker verification using context-aware masking},
  author={Wang, Hui and Zheng, Siqi and Chen, Yafeng and Cheng, Luyao and Chen, Qian},
  journal={arXiv preprint arXiv:2303.00332},
  year={2023}
}

@inproceedings{31_liu2022video,
  title={Video swin transformer},
  author={Liu, Ze and Ning, Jia and Cao, Yue and Wei, Yixuan and Zhang, Zheng and Lin, Stephen and Hu, Han},
  booktitle={Proceedings of the IEEE/CVF conference on computer vision and pattern recognition},
  pages={3202--3211},
  year={2022}
}

@article{32_hsu2021hubert,
  title={Hubert: Self-supervised speech representation learning by masked prediction of hidden units},
  author={Hsu, Wei-Ning and Bolte, Benjamin and Tsai, Yao-Hung Hubert and Lakhotia, Kushal and Salakhutdinov, Ruslan and Mohamed, Abdelrahman},
  journal={IEEE/ACM transactions on audio, speech, and language processing},
  volume={29},
  pages={3451--3460},
  year={2021},
  publisher={IEEE}
}

@article{33_sun2023efficient,
  title={Efficient multimodal transformer with dual-level feature restoration for robust multimodal sentiment analysis},
  author={Sun, Licai and Lian, Zheng and Liu, Bin and Tao, Jianhua},
  journal={IEEE Transactions on Affective Computing},
  volume={15},
  number={1},
  pages={309--325},
  year={2023},
  publisher={IEEE}
}

@inproceedings{34_tsai2019multimodal,
  title={Multimodal transformer for unaligned multimodal language sequences},
  author={Tsai, Yao-Hung Hubert and Bai, Shaojie and Liang, Paul Pu and Kolter, J Zico and Morency, Louis-Philippe and Salakhutdinov, Ruslan},
  booktitle={Proceedings of the conference. Association for computational linguistics. Meeting},
  volume={2019},
  pages={6558},
  year={2019},
  organization={NIH Public Access}
}

@inproceedings{35_shi2022learning,
  title={Learning Audio-Visual Speech Representation by Masked Multimodal Cluster Prediction},
  author={Shi, Bowen and Hsu, Wei-Ning and Lakhotia, Kushal and Mohamed, Abdelrahman},
  booktitle={International Conference on Learning Representations},
  year={2022}
}

@inproceedings{36_guo2024crossmae,
  title={CrossMAE: Cross-Modality Masked Autoencoders for Region-Aware Audio-Visual Pre-Training},
  author={Guo, Yuxin and Sun, Siyang and Ma, Shuailei and Zheng, Kecheng and Bao, Xiaoyi and Ma, Shijie and Zou, Wei and Zheng, Yun},
  booktitle={Proceedings of the IEEE/CVF Conference on Computer Vision and Pattern Recognition},
  pages={26721--26731},
  year={2024}
}

@article{37_liu2024masked,
  title={Masked co-attention model for audio-visual event localization},
  author={Liu, Hengwei and Gu, Xiaodong},
  journal={Applied Intelligence},
  volume={54},
  number={2},
  pages={1691--1705},
  year={2024},
  publisher={Springer}
}

@inproceedings{38_woo2024speech,
  title={Speech Guided Masked Image Modeling for Visually Grounded Speech},
  author={Woo, Jongbhin and Ryu, Hyeonggon and Senocak, Arda and Chung, Joon Son},
  booktitle={ICASSP 2024-2024 IEEE International Conference on Acoustics, Speech and Signal Processing (ICASSP)},
  pages={8361--8365},
  year={2024},
  organization={IEEE}
}

@article{39_gong2022contrastive,
  title={Contrastive audio-visual masked autoencoder},
  author={Gong, Yuan and Rouditchenko, Andrew and Liu, Alexander H and Harwath, David and Karlinsky, Leonid and Kuehne, Hilde and Glass, James},
  journal={arXiv preprint arXiv:2210.07839},
  year={2022}
}

@inproceedings{40_georgescu2023audiovisual,
  title={Audiovisual masked autoencoders},
  author={Georgescu, Mariana-Iuliana and Fonseca, Eduardo and Ionescu, Radu Tudor and Lucic, Mario and Schmid, Cordelia and Arnab, Anurag},
  booktitle={Proceedings of the IEEE/CVF International Conference on Computer Vision},
  pages={16144--16154},
  year={2023}
}

@inproceedings{41_mo2024unveiling,
  title={Unveiling the Power of Audio-Visual Early Fusion Transformers with Dense Interactions through Masked Modeling},
  author={Mo, Shentong and Morgado, Pedro},
  booktitle={Proceedings of the IEEE/CVF Conference on Computer Vision and Pattern Recognition},
  pages={27186--27196},
  year={2024}
}

@article{42_huang2024mavil,
  title={Mavil: Masked audio-video learners},
  author={Huang, Po-Yao and Sharma, Vasu and Xu, Hu and Ryali, Chaitanya and Li, Yanghao and Li, Shang-Wen and Ghosh, Gargi and Malik, Jitendra and Feichtenhofer, Christoph and others},
  journal={Advances in Neural Information Processing Systems},
  volume={36},
  year={2024}
}

@phdthesis{43_sadok2024audiovisual,
  title={Audiovisual speech representation learning applied to emotion recognition},
  author={Sadok, Samir},
  year={2024},
  school={CentraleSup{\'e}lec}
}

@article{44_xiang2024multimae,
  title={MultiMAE-DER: Multimodal Masked Autoencoder for Dynamic Emotion Recognition},
  author={Xiang, Peihao and Lin, Chaohao and Wu, Kaida and Bai, Ou},
  journal={arXiv preprint arXiv:2404.18327},
  year={2024}
}

@inproceedings{45_singh2023effectiveness,
  title={The effectiveness of MAE pre-pretraining for billion-scale pretraining},
  author={Singh, Mannat and Duval, Quentin and Alwala, Kalyan Vasudev and Fan, Haoqi and Aggarwal, Vaibhav and Adcock, Aaron and Joulin, Armand and Doll{\'a}r, Piotr and Feichtenhofer, Christoph and Girshick, Ross and others},
  booktitle={Proceedings of the IEEE/CVF International Conference on Computer Vision},
  pages={5484--5494},
  year={2023}
}

@inproceedings{46_han2024efficient,
  title={Efficient MAE towards Large-Scale Vision Transformers},
  author={Han, Qiu and Zhang, Gongjie and Huang, Jiaxing and Gao, Peng and Wei, Zhang and Lu, Shijian},
  booktitle={Proceedings of the IEEE/CVF Winter Conference on Applications of Computer Vision},
  pages={606--615},
  year={2024}
}

@inproceedings{47_xie2023data,
  title={On data scaling in masked image modeling},
  author={Xie, Zhenda and Zhang, Zheng and Cao, Yue and Lin, Yutong and Wei, Yixuan and Dai, Qi and Hu, Han},
  booktitle={Proceedings of the IEEE/CVF Conference on Computer Vision and Pattern Recognition},
  pages={10365--10374},
  year={2023}
}

@article{48_feichtenhofer2022masked,
  title={Masked autoencoders as spatiotemporal learners},
  author={Feichtenhofer, Christoph and Li, Yanghao and He, Kaiming and others},
  journal={Advances in neural information processing systems},
  volume={35},
  pages={35946--35958},
  year={2022}
}

@article{49_huang2022masked,
  title={Masked autoencoders that listen},
  author={Huang, Po-Yao and Xu, Hu and Li, Juncheng and Baevski, Alexei and Auli, Michael and Galuba, Wojciech and Metze, Florian and Feichtenhofer, Christoph},
  journal={Advances in Neural Information Processing Systems},
  volume={35},
  pages={28708--28720},
  year={2022}
}

@article{50_qing2023mar,
  title={Mar: Masked autoencoders for efficient action recognition},
  author={Qing, Zhiwu and Zhang, Shiwei and Huang, Ziyuan and Wang, Xiang and Wang, Yuehuan and Lv, Yiliang and Gao, Changxin and Sang, Nong},
  journal={IEEE Transactions on Multimedia},
  volume={26},
  pages={218--233},
  year={2023},
  publisher={IEEE}
}

@inproceedings{52_cheng2020look,
  title={Look, listen, and attend: Co-attention network for self-supervised audio-visual representation learning},
  author={Cheng, Ying and Wang, Ruize and Pan, Zhihao and Feng, Rui and Zhang, Yuejie},
  booktitle={Proceedings of the 28th ACM International Conference on Multimedia},
  pages={3884--3892},
  year={2020}
}

@article{53_hu2023cross,
  title={Cross-modal global interaction and local alignment for audio-visual speech recognition},
  author={Hu, Yuchen and Li, Ruizhe and Chen, Chen and Zou, Heqing and Zhu, Qiushi and Chng, Eng Siong},
  journal={arXiv preprint arXiv:2305.09212},
  year={2023}
}

@inproceedings{55_fan2020cn,
  title={Cn-celeb: a challenging chinese speaker recognition dataset},
  author={Fan, Yue and Kang, JW and Li, LT and Li, KC and Chen, HL and Cheng, ST and Zhang, PY and Zhou, ZY and Cai, YQ and Wang, Dong},
  booktitle={ICASSP 2020-2020 IEEE International Conference on Acoustics, Speech and Signal Processing (ICASSP)},
  pages={7604--7608},
  year={2020},
  organization={IEEE}
}

@article{56_lian2024mer,
  title={MER 2024: Semi-Supervised Learning, Noise Robustness, and Open-Vocabulary Multimodal Emotion Recognition},
  author={Lian, Zheng and Sun, Haiyang and Sun, Licai and Wen, Zhuofan and Zhang, Siyuan and Chen, Shun and Gu, Hao and Zhao, Jinming and Ma, Ziyang and Chen, Xie and others},
  journal={arXiv preprint arXiv:2404.17113},
  year={2024}
}

@article{57_chung2018voxceleb2,
  title={Voxceleb2: Deep speaker recognition},
  author={Chung, Joon Son and Nagrani, Arsha and Zisserman, Andrew},
  journal={arXiv preprint arXiv:1806.05622},
  year={2018}
}

@article{58_ephrat2018looking,
  title={Looking to listen at the cocktail party: A speaker-independent audio-visual model for speech separation},
  author={Ephrat, Ariel and Mosseri, Inbar and Lang, Oran and Dekel, Tali and Wilson, Kevin and Hassidim, Avinatan and Freeman, William T and Rubinstein, Michael},
  journal={arXiv preprint arXiv:1804.03619},
  year={2018}
}

@inproceedings{59_zhu2022celebv,
  title={CelebV-HQ: A large-scale video facial attributes dataset},
  author={Zhu, Hao and Wu, Wayne and Zhu, Wentao and Jiang, Liming and Tang, Siwei and Zhang, Li and Liu, Ziwei and Loy, Chen Change},
  booktitle={European conference on computer vision},
  pages={650--667},
  year={2022},
  organization={Springer}
}

@inproceedings{61_sarkar2023avcaffe,
  title={AVCAffe: a large scale audio-visual dataset of cognitive load and affect for remote work},
  author={Sarkar, Pritam and Posen, Aaron and Etemad, Ali},
  booktitle={Proceedings of the AAAI Conference on Artificial Intelligence},
  volume={37},
  number={1},
  pages={76--85},
  year={2023}
}

@article{62_zhang2021werewolf,
  title={Werewolf-XL: A database for identifying spontaneous affect in large competitive group interactions},
  author={Zhang, Kejun and Wu, Xinda and Xie, Xinhang and Zhang, Xiaoran and Zhang, Hui and Chen, Xiaoyu and Sun, Lingyun},
  journal={IEEE Transactions on Affective Computing},
  volume={14},
  number={2},
  pages={1201--1214},
  year={2021},
  publisher={IEEE}
}

@inproceedings{63_liu2022mafw,
  title={Mafw: A large-scale, multi-modal, compound affective database for dynamic facial expression recognition in the wild},
  author={Liu, Yuanyuan and Dai, Wei and Feng, Chuanxu and Wang, Wenbin and Yin, Guanghao and Zeng, Jiabei and Shan, Shiguang},
  booktitle={Proceedings of the 30th ACM International Conference on Multimedia},
  pages={24--32},
  year={2022}
}

@inproceedings{64_lian2023mer,
  title={Mer 2023: Multi-label learning, modality robustness, and semi-supervised learning},
  author={Lian, Zheng and Sun, Haiyang and Sun, Licai and Chen, Kang and Xu, Mngyu and Wang, Kexin and Xu, Ke and He, Yu and Li, Ying and Zhao, Jinming and others},
  booktitle={Proceedings of the 31st ACM International Conference on Multimedia},
  pages={9610--9614},
  year={2023}
}

@article{65_busso2008iemocap,
  title={IEMOCAP: Interactive emotional dyadic motion capture database},
  author={Busso, Carlos and Bulut, Murtaza and Lee, Chi-Chun and Kazemzadeh, Abe and Mower, Emily and Kim, Samuel and Chang, Jeannette N and Lee, Sungbok and Narayanan, Shrikanth S},
  journal={Language resources and evaluation},
  volume={42},
  pages={335--359},
  year={2008},
  publisher={Springer}
}

@article{66_cao2014crema,
  title={Crema-d: Crowd-sourced emotional multimodal actors dataset},
  author={Cao, Houwei and Cooper, David G and Keutmann, Michael K and Gur, Ruben C and others},
  journal={IEEE transactions on affective computing},
  volume={5},
  number={4},
  pages={377--390},
  year={2014},
  publisher={IEEE}
}

@article{67_livingstone2018ryerson,
  title={The Ryerson Audio-Visual Database of Emotional Speech and Song (RAVDESS): A dynamic, multimodal set of facial and vocal expressions in North American English},
  author={Livingstone, Steven R and Russo, Frank A},
  journal={PloS one},
  volume={13},
  number={5},
  pages={e0196391},
  year={2018},
  publisher={Public Library of Science San Francisco, CA USA}
}

@article{68_busso2016msp,
  title={MSP-IMPROV: An acted corpus of dyadic interactions to study emotion perception},
  author={Busso, Carlos and Parthasarathy, Srinivas and Burmania, Alec and AbdelWahab, Mohammed and Sadoughi, Najmeh and Provost, Emily Mower},
  journal={IEEE Transactions on Affective Computing},
  volume={8},
  number={1},
  pages={67--80},
  year={2016},
  publisher={IEEE}
}

@article{69_ben2021video,
  title={Video-based facial micro-expression analysis: A survey of datasets, features and algorithms},
  author={Ben, Xianye and Ren, Yi and Zhang, Junping and Wang, Su-Jing and Kpalma, Kidiyo and Meng, Weixiao and Liu, Yong-Jin},
  journal={IEEE transactions on pattern analysis and machine intelligence},
  volume={44},
  number={9},
  pages={5826--5846},
  year={2021},
  publisher={IEEE}
}

@article{70_yan2014casme,
  title={CASME II: An improved spontaneous micro-expression database and the baseline evaluation},
  author={Yan, Wen-Jing and Li, Xiaobai and Wang, Su-Jing and Zhao, Guoying and Liu, Yong-Jin and Chen, Yu-Hsin and Fu, Xiaolan},
  journal={PloS one},
  volume={9},
  number={1},
  pages={e86041},
  year={2014},
  publisher={Public Library of Science San Francisco, USA}
}

@article{71_li2022cas,
  title={CAS (ME) 3: A third generation facial spontaneous micro-expression database with depth information and high ecological validity},
  author={Li, Jingting and Dong, Zizhao and Lu, Shaoyuan and Wang, Su-Jing and Yan, Wen-Jing and Ma, Yinhuan and Liu, Ye and Huang, Changbing and Fu, Xiaolan},
  journal={IEEE Transactions on Pattern Analysis and Machine Intelligence},
  volume={45},
  number={3},
  pages={2782--2800},
  year={2022},
  publisher={IEEE}
}

@article{72_davison2016samm,
  title={Samm: A spontaneous micro-facial movement dataset},
  author={Davison, Adrian K and Lansley, Cliff and Costen, Nicholas and Tan, Kevin and Yap, Moi Hoon},
  journal={IEEE transactions on affective computing},
  volume={9},
  number={1},
  pages={116--129},
  year={2016},
  publisher={IEEE}
}

@inproceedings{73_li2013spontaneous,
  title={A spontaneous micro-expression database: Inducement, collection and baseline},
  author={Li, Xiaobai and Pfister, Tomas and Huang, Xiaohua and Zhao, Guoying and Pietik{\"a}inen, Matti},
  booktitle={2013 10th IEEE International Conference and Workshops on Automatic face and gesture recognition (fg)},
  pages={1--6},
  year={2013},
  organization={IEEE}
}

@inproceedings{74_nguyen2023micron,
  title={Micron-bert: Bert-based facial micro-expression recognition},
  author={Nguyen, Xuan-Bac and Duong, Chi Nhan and Li, Xin and Gauch, Susan and Seo, Han-Seok and Luu, Khoa},
  booktitle={Proceedings of the IEEE/CVF Conference on Computer Vision and Pattern Recognition},
  pages={1482--1492},
  year={2023}
}

@inproceedings{75_liong2019shallow,
  title={Shallow triple stream three-dimensional cnn (ststnet) for micro-expression recognition},
  author={Liong, Sze-Teng and Gan, Yee Siang and See, John and Khor, Huai-Qian and Huang, Yen-Chang},
  booktitle={2019 14th IEEE international conference on automatic face \& gesture recognition (FG 2019)},
  pages={1--5},
  year={2019},
  organization={IEEE}
}

@inproceedings{76_van2019capsulenet,
  title={CapsuleNet for micro-expression recognition},
  author={Van Quang, Nguyen and Chun, Jinhee and Tokuyama, Takeshi},
  booktitle={2019 14th IEEE International Conference on Automatic Face \& Gesture Recognition (FG 2019)},
  pages={1--7},
  year={2019},
  organization={IEEE}
}

@article{77_wang2024htnet,
  title={Htnet for micro-expression recognition},
  author={Wang, Zhifeng and Zhang, Kaihao and Luo, Wenhan and Sankaranarayana, Ramesh},
  journal={Neurocomputing},
  volume={602},
  pages={128196},
  year={2024},
  publisher={Elsevier}
}

@article{78_xia2020revealing,
  title={Revealing the invisible with model and data shrinking for composite-database micro-expression recognition},
  author={Xia, Zhaoqiang and Peng, Wei and Khor, Huai-Qian and Feng, Xiaoyi and Zhao, Guoying},
  journal={IEEE Transactions on Image Processing},
  volume={29},
  pages={8590--8605},
  year={2020},
  publisher={IEEE}
}

@inproceedings{79_liu2019neural,
  title={A neural micro-expression recognizer},
  author={Liu, Yuchi and Du, Heming and Zheng, Liang and Gedeon, Tom},
  booktitle={2019 14th IEEE international conference on automatic face \& gesture recognition (FG 2019)},
  pages={1--4},
  year={2019},
  organization={IEEE}
}

@article{80_zhao2007dynamic,
  title={Dynamic texture recognition using local binary patterns with an application to facial expressions},
  author={Zhao, Guoying and Pietikainen, Matti},
  journal={IEEE transactions on pattern analysis and machine intelligence},
  volume={29},
  number={6},
  pages={915--928},
  year={2007},
  publisher={IEEE}
}

@article{81_zhou2022feature,
  title={Feature refinement: An expression-specific feature learning and fusion method for micro-expression recognition},
  author={Zhou, Ling and Mao, Qirong and Huang, Xiaohua and Zhang, Feifei and Zhang, Zhihong},
  journal={Pattern Recognition},
  volume={122},
  pages={108275},
  year={2022},
  publisher={Elsevier}
}

@article{82_chen2024finecliper,
  title={Finecliper: Multi-modal fine-grained clip for dynamic facial expression recognition with adapters},
  author={Chen, Haodong and Huang, Haojian and Dong, Junhao and Zheng, Mingzhe and Shao, Dian},
  journal={arXiv preprint arXiv:2407.02157},
  year={2024}
}

@article{84_chen2024unilearn,
  title={UniLearn: Enhancing Dynamic Facial Expression Recognition through Unified Pre-Training and Fine-Tuning on Images and Videos},
  author={Chen, Yin and Li, Jia and Zhang, Yu and Hu, Zhenzhen and Shan, Shiguang and Wang, Meng and Hong, Richang},
  journal={arXiv preprint arXiv:2409.06154},
  year={2024}
}

@article{85_zhang2023transformer,
  title={Transformer-based multimodal emotional perception for dynamic facial expression recognition in the wild},
  author={Zhang, Xiaoqin and Li, Min and Lin, Sheng and Xu, Hang and Xiao, Guobao},
  journal={IEEE Transactions on Circuits and Systems for Video Technology},
  year={2023},
  publisher={IEEE}
}

@article{86_hsu2021hubert,
  title={Hubert: Self-supervised speech representation learning by masked prediction of hidden units},
  author={Hsu, Wei-Ning and Bolte, Benjamin and Tsai, Yao-Hung Hubert and Lakhotia, Kushal and Salakhutdinov, Ruslan and Mohamed, Abdelrahman},
  journal={IEEE/ACM transactions on audio, speech, and language processing},
  volume={29},
  pages={3451--3460},
  year={2021},
  publisher={IEEE}
}

@article{87_chen2022wavlm,
  title={Wavlm: Large-scale self-supervised pre-training for full stack speech processing},
  author={Chen, Sanyuan and Wang, Chengyi and Chen, Zhengyang and Wu, Yu and Liu, Shujie and Chen, Zhuo and Li, Jinyu and Kanda, Naoyuki and Yoshioka, Takuya and Xiao, Xiong and others},
  journal={IEEE Journal of Selected Topics in Signal Processing},
  volume={16},
  number={6},
  pages={1505--1518},
  year={2022},
  publisher={IEEE}
}

@inproceedings{89_zhang2022wenetspeech,
  title={Wenetspeech: A 10000+ hours multi-domain mandarin corpus for speech recognition},
  author={Zhang, Binbin and Lv, Hang and Guo, Pengcheng and Shao, Qijie and Yang, Chao and Xie, Lei and Xu, Xin and Bu, Hui and Chen, Xiaoyu and Zeng, Chenchen and others},
  booktitle={ICASSP 2022-2022 IEEE International Conference on Acoustics, Speech and Signal Processing (ICASSP)},
  pages={6182--6186},
  year={2022},
  organization={IEEE}
}

@inproceedings{90_he2016deep,
  title={Deep residual learning for image recognition},
  author={He, Kaiming and Zhang, Xiangyu and Ren, Shaoqing and Sun, Jian},
  booktitle={Proceedings of the IEEE conference on computer vision and pattern recognition},
  pages={770--778},
  year={2016}
}

@article{91_zhao2021learning,
  title={Learning deep global multi-scale and local attention features for facial expression recognition in the wild},
  author={Zhao, Zengqun and Liu, Qingshan and Wang, Shanmin},
  journal={IEEE Transactions on Image Processing},
  volume={30},
  pages={6544--6556},
  year={2021},
  publisher={IEEE}
}

@inproceedings{92_zhao2021former,
  title={Former-dfer: Dynamic facial expression recognition transformer},
  author={Zhao, Zengqun and Liu, Qingshan},
  booktitle={Proceedings of the 29th ACM International Conference on Multimedia},
  pages={1553--1561},
  year={2021}
}

@inproceedings{93_radford2023robust,
  title={Robust speech recognition via large-scale weak supervision},
  author={Radford, Alec and Kim, Jong Wook and Xu, Tao and Brockman, Greg and McLeavey, Christine and Sutskever, Ilya},
  booktitle={International conference on machine learning},
  pages={28492--28518},
  year={2023},
  organization={PMLR}
}

@article{94_oquab2023dinov2,
  title={Dinov2: Learning robust visual features without supervision},
  author={Oquab, Maxime and Darcet, Timoth{\'e}e and Moutakanni, Th{\'e}o and Vo, Huy and Szafraniec, Marc and Khalidov, Vasil and Fernandez, Pierre and Haziza, Daniel and Massa, Francisco and El-Nouby, Alaaeldin and others},
  journal={arXiv preprint arXiv:2304.07193},
  year={2023}
}

@inproceedings{95_tran2023saaml,
  title={Saaml: A framework for semi-supervised affective adaptation via metric learning},
  author={Tran, Minh and Kim, Yelin and Su, Che-Chun and Kuo, Cheng-Hao and Soleymani, Mohammad},
  booktitle={Proceedings of the 31st ACM International Conference on Multimedia},
  pages={6004--6015},
  year={2023}
}

@article{96_baevski2020wav2vec,
  title={wav2vec 2.0: A framework for self-supervised learning of speech representations},
  author={Baevski, Alexei and Zhou, Yuhao and others},
  journal={Advances in neural information processing systems},
  volume={33},
  pages={12449--12460},
  year={2020}
}

@inproceedings{97_lee2020parameter,
  title={Parameter efficient multimodal transformers for video representation learning},
  author={Lee, Sangho and Yu, Youngjae and Kim, Gunhee and Breuel, Thomas and Kautz, Jan and Song, Yale},
  booktitle={9th International Conference on Learning Representations, ICLR 2021},
  year={2021}
}

@article{98_eyben2015geneva,
  title={The Geneva minimalistic acoustic parameter set (GeMAPS) for voice research and affective computing},
  author={Eyben, Florian and Scherer, Klaus R and Schuller, Bj{\"o}rn W and Sundberg, Johan and Andr{\'e}, Elisabeth and Busso, Carlos and Devillers, Laurence Y and Epps, Julien and Laukka, Petri and Narayanan, Shrikanth S and others},
  journal={IEEE transactions on affective computing},
  volume={7},
  number={2},
  pages={190--202},
  year={2015},
  publisher={IEEE}
}

@article{99_schuller2018age,
  title={The age of artificial emotional intelligence},
  author={Schuller, Dagmar and Schuller, Bj{\"o}rn W},
  journal={Computer},
  volume={51},
  number={9},
  pages={38--46},
  year={2018},
  publisher={IEEE}
}

@inproceedings{100_chen2023cn,
  title={CN-CVS: A mandarin audio-visual dataset for large vocabulary continuous visual to speech synthesis},
  author={Chen, Chen and Wang, Dong and Zheng, Thomas Fang},
  booktitle={ICASSP 2023-2023 IEEE International Conference on Acoustics, Speech and Signal Processing (ICASSP)},
  pages={1--5},
  year={2023},
  organization={IEEE}
}

@article{101_loshchilov2017decoupled,
  title={Decoupled weight decay regularization},
  author={Loshchilov, I},
  journal={arXiv preprint arXiv:1711.05101},
  year={2017}
}

@inproceedings{102_hoffer2020augment,
  title={Augment your batch: Improving generalization through instance repetition},
  author={Hoffer, Elad and Ben-Nun, Tal and Hubara, Itay and Giladi, Niv and Hoefler, Torsten and Soudry, Daniel},
  booktitle={Proceedings of the IEEE/CVF Conference on Computer Vision and Pattern Recognition},
  pages={8129--8138},
  year={2020}
}

@inproceedings{103_cubuk2020randaugment,
  title={Randaugment: Practical automated data augmentation with a reduced search space},
  author={Cubuk, Ekin D and Zoph, Barret and Shlens, Jonathon and Le, Quoc V},
  booktitle={Proceedings of the IEEE/CVF conference on computer vision and pattern recognition workshops},
  pages={702--703},
  year={2020}
}

@inproceedings{104_szegedy2016rethinking,
  title={Rethinking the inception architecture for computer vision},
  author={Szegedy, Christian and Vanhoucke, Vincent and Ioffe, Sergey and Shlens, Jon and Wojna, Zbigniew},
  booktitle={Proceedings of the IEEE conference on computer vision and pattern recognition},
  pages={2818--2826},
  year={2016}
}

@article{105_zhang2017mixup,
  title={mixup: Beyond empirical risk minimization},
  author={Zhang, Hongyi},
  journal={arXiv preprint arXiv:1710.09412},
  year={2017}
}

@inproceedings{106_parkhi2015deep,
  title={Deep face recognition},
  author={Parkhi, Omkar and Vedaldi, Andrea and Zisserman, Andrew},
  booktitle={BMVC 2015-Proceedings of the British Machine Vision Conference 2015},
  year={2015},
  organization={British Machine Vision Association}
}

@inproceedings{107_tran2015learning,
  title={Learning spatiotemporal features with 3d convolutional networks},
  author={Tran, Du and Bourdev, Lubomir and Fergus, Rob and Torresani, Lorenzo and Paluri, Manohar},
  booktitle={Proceedings of the IEEE international conference on computer vision},
  pages={4489--4497},
  year={2015}
}

@inproceedings{108_yoon2020attentive,
  title={Attentive modality hopping mechanism for speech emotion recognition},
  author={Yoon, Seunghyun and Dey, Subhadeep and Lee, Hwanhee and Jung, Kyomin},
  booktitle={ICASSP 2020-2020 IEEE International Conference on Acoustics, Speech and Signal Processing (ICASSP)},
  pages={3362--3366},
  year={2020},
  organization={IEEE}
}

@inproceedings{109_yoon2018multimodal,
  title={Multimodal speech emotion recognition using audio and text},
  author={Yoon, Seunghyun and Byun, Seokhyun and Jung, Kyomin},
  booktitle={2018 IEEE spoken language technology workshop (SLT)},
  pages={112--118},
  year={2018},
  organization={IEEE}
}

@inproceedings{110_rajan2022cross,
  title={Is cross-attention preferable to self-attention for multi-modal emotion recognition?},
  author={Rajan, Vandana and Brutti, Alessio and Cavallaro, Andrea},
  booktitle={ICASSP 2022-2022 IEEE International Conference on Acoustics, Speech and Signal Processing (ICASSP)},
  pages={4693--4697},
  year={2022},
  organization={IEEE}
}

@article{111_chen2024static,
  title={From static to dynamic: Adapting landmark-aware image models for facial expression recognition in videos},
  author={Chen, Yin and Li, Jia and Shan, Shiguang and Wang, Meng and Hong, Richang},
  journal={IEEE Transactions on Affective Computing},
  year={2024},
  publisher={IEEE}
}

@inproceedings{112_gowda2024fe,
  title={Fe-adapter: Adapting image-based emotion classifiers to videos},
  author={Gowda, Shreyank N and Gao, Boyan and Clifton, David A},
  booktitle={2024 IEEE 18th International Conference on Automatic Face and Gesture Recognition (FG)},
  pages={1--6},
  year={2024},
  organization={IEEE}
}

@inproceedings{113_foteinopoulou2024emoclip,
  title={Emoclip: A vision-language method for zero-shot video facial expression recognition},
  author={Foteinopoulou, Niki Maria and Patras, Ioannis},
  booktitle={2024 IEEE 18th International Conference on Automatic Face and Gesture Recognition (FG)},
  pages={1--10},
  year={2024},
  organization={IEEE}
}

@article{114_tao20243,
  title={A3lign-DFER: Pioneering Comprehensive Dynamic Affective Alignment for Dynamic Facial Expression Recognition with CLIP},
  author={Tao, Zeng and Wang, Yan and Lin, Junxiong and Wang, Haoran and Mai, Xinji and Yu, Jiawen and Tong, Xuan and Zhou, Ziheng and Yan, Shaoqi and Zhao, Qing and others},
  journal={arXiv preprint arXiv:2403.04294},
  year={2024}
}

@inproceedings{115_tran2018closer,
  title={A closer look at spatiotemporal convolutions for action recognition},
  author={Tran, Du and Wang, Heng and Torresani, Lorenzo and Ray, Jamie and LeCun, Yann and Paluri, Manohar},
  booktitle={Proceedings of the IEEE conference on Computer Vision and Pattern Recognition},
  pages={6450--6459},
  year={2018}
}

@inproceedings{116_hara2018can,
  title={Can spatiotemporal 3d cnns retrace the history of 2d cnns and imagenet?},
  author={Hara, Kensho and Kataoka, Hirokatsu and Satoh, Yutaka},
  booktitle={Proceedings of the IEEE conference on Computer Vision and Pattern Recognition},
  pages={6546--6555},
  year={2018}
}

@article{117_liu2022clip,
  title={Clip-aware expressive feature learning for video-based facial expression recognition},
  author={Liu, Yuanyuan and Feng, Chuanxu and Yuan, Xiaohui and Zhou, Lin and Wang, Wenbin and Qin, Jie and Luo, Zhongwen},
  journal={Information Sciences},
  volume={598},
  pages={182--195},
  year={2022},
  publisher={Elsevier}
}

@article{118_liu2023expression,
  title={Expression snippet transformer for robust video-based facial expression recognition},
  author={Liu, Yuanyuan and Wang, Wenbin and Feng, Chuanxu and Zhang, Haoyu and Chen, Zhe and Zhan, Yibing},
  journal={Pattern Recognition},
  volume={138},
  pages={109368},
  year={2023},
  publisher={Elsevier}
}

@article{119_ma2022spatio,
  title={Spatio-temporal transformer for dynamic facial expression recognition in the wild},
  author={Ma, Fuyan and Sun, Bin and Li, Shutao},
  journal={arXiv preprint arXiv:2205.04749},
  year={2022}
}

@article{120_li2022nr,
  title={Nr-dfernet: Noise-robust network for dynamic facial expression recognition},
  author={Li, Hanting and Sui, Mingzhe and Zhu, Zhaoqing and others},
  journal={arXiv preprint arXiv:2206.04975},
  year={2022}
}

@inproceedings{121_wang2022dpcnet,
  title={Dpcnet: Dual path multi-excitation collaborative network for facial expression representation learning in videos},
  author={Wang, Yan and Sun, Yixuan and Song, Wei and Gao, Shuyong and Huang, Yiwen and Chen, Zhaoyu and Ge, Weifeng and Zhang, Wenqiang},
  booktitle={Proceedings of the 30th ACM International Conference on Multimedia},
  pages={101--110},
  year={2022}
}

@inproceedings{122_li2023intensity,
  title={Intensity-aware loss for dynamic facial expression recognition in the wild},
  author={Li, Hanting and Niu, Hongjing and Zhu, Zhaoqing and Zhao, Feng},
  booktitle={Proceedings of the AAAI Conference on Artificial Intelligence},
  volume={37},
  number={1},
  pages={67--75},
  year={2023}
}

@inproceedings{123_wang2023rethinking,
  title={Rethinking the learning paradigm for dynamic facial expression recognition},
  author={Wang, Hanyang and Li, Bo and Wu, Shuang and Shen, Siyuan and Liu, Feng and Ding, Shouhong and Zhou, Aimin},
  booktitle={Proceedings of the IEEE/CVF conference on computer vision and pattern recognition},
  pages={17958--17968},
  year={2023}
}

@inproceedings{124_li2024cliper,
  title={Cliper: A unified vision-language framework for in-the-wild facial expression recognition},
  author={Li, Hanting and Niu, Hongjing and Zhu, Zhaoqing and Zhao, Feng},
  booktitle={2024 IEEE International Conference on Multimedia and Expo (ICME)},
  pages={1--6},
  year={2024},
  organization={IEEE}
}

@inproceedings{125_goncalves2022auxformer,
  title={AuxFormer: Robust approach to audiovisual emotion recognition},
  author={Goncalves, Lucas and Busso, Carlos},
  booktitle={ICASSP 2022-2022 IEEE International Conference on Acoustics, Speech and Signal Processing (ICASSP)},
  pages={7357--7361},
  year={2022},
  organization={IEEE}
}

@inproceedings{126_ghaleb2019multimodal,
  title={Multimodal and temporal perception of audio-visual cues for emotion recognition},
  author={Ghaleb, Esam and Popa, Mirela and Asteriadis, Stylianos},
  booktitle={2019 8th international conference on affective computing and intelligent interaction (ACII)},
  pages={552--558},
  year={2019},
  organization={IEEE}
}

@article{127_goncalves2022robust,
  title={Robust audiovisual emotion recognition: Aligning modalities, capturing temporal information, and handling missing features},
  author={Goncalves, Lucas and Busso, Carlos},
  journal={IEEE Transactions on Affective Computing},
  volume={13},
  number={4},
  pages={2156--2170},
  year={2022},
  publisher={IEEE}
}

@article{128_lei2023audio,
  title={Audio-visual emotion recognition with preference learning based on intended and multi-modal perceived labels},
  author={Lei, Yuanyuan and Cao, Houwei},
  journal={IEEE Transactions on Affective Computing},
  volume={14},
  number={4},
  pages={2954--2969},
  year={2023},
  publisher={IEEE}
}

@inproceedings{129_keesing2023emotion,
  title={Emotion Recognition ToolKit (ERTK): Standardising Tools For Emotion Recognition Research},
  author={Keesing, Aaron and Koh, Yun Sing and Yogarajan, Vithya and Witbrock, Michael},
  booktitle={Proceedings of the 31st ACM International Conference on Multimedia},
  pages={9693--9696},
  year={2023}
}

@inproceedings{130_tran2022pre,
  title={A pre-trained audio-visual transformer for emotion recognition},
  author={Tran, Minh and Soleymani, Mohammad},
  booktitle={ICASSP 2022-2022 IEEE International Conference on Acoustics, Speech and Signal Processing (ICASSP)},
  pages={4698--4702},
  year={2022},
  organization={IEEE}
}

@article{131_zadeh2017tensor,
  title={Tensor fusion network for multimodal sentiment analysis},
  author={Zadeh, Amir and Chen, Minghai and Poria, Soujanya and Cambria, Erik and Morency, Louis-Philippe},
  journal={arXiv preprint arXiv:1707.07250},
  year={2017}
}

@inproceedings{132_ghaleb2020multimodal,
  title={Multimodal attention-mechanism for temporal emotion recognition},
  author={Ghaleb, Esam and Niehues, Jan and Asteriadis, Stylianos},
  booktitle={2020 IEEE International Conference on Image Processing (ICIP)},
  pages={251--255},
  year={2020},
  organization={IEEE}
}

@inproceedings{133_goncalves2023learning,
  title={Learning cross-modal audiovisual representations with ladder networks for emotion recognition},
  author={Goncalves, Lucas and Busso, Carlos},
  booktitle={ICASSP 2023-2023 IEEE International Conference on Acoustics, Speech and Signal Processing (ICASSP)},
  pages={1--5},
  year={2023},
  organization={IEEE}
}

@inproceedings{134_tseng2024av,
  title={Av-superb: A multi-task evaluation benchmark for audio-visual representation models},
  author={Tseng, Yuan and Berry, Layne and Chen, Yi-Ting and Chiu, I-Hsiang and Lin, Hsuan-Hao and Liu, Max and Peng, Puyuan and Shih, Yi-Jen and Wang, Hung-Yu and Wu, Haibin and others},
  booktitle={ICASSP 2024-2024 IEEE International Conference on Acoustics, Speech and Signal Processing (ICASSP)},
  pages={6890--6894},
  year={2024},
  organization={IEEE}
}

@article{135_mittal2022learning,
  title={Learning state-aware visual representations from audible interactions},
  author={Mittal, Himangi and Morgado, Pedro and Jain, Unnat and Gupta, Abhinav},
  journal={Advances in Neural Information Processing Systems},
  volume={35},
  pages={23765--23779},
  year={2022}
}

@inproceedings{136_dalal2005histograms,
  title={Histograms of oriented gradients for human detection},
  author={Dalal, Navneet and Triggs, Bill},
  booktitle={2005 IEEE computer society conference on computer vision and pattern recognition (CVPR'05)},
  volume={1},
  pages={886--893},
  year={2005},
  organization={Ieee}
}

@inproceedings{137_hershey2017cnn,
  title={CNN architectures for large-scale audio classification},
  author={Hershey, Shawn and Chaudhuri, Sourish and Ellis, Daniel PW and Gemmeke, Jort F and Jansen, Aren and Moore, R Channing and Plakal, Manoj and Platt, Devin and Saurous, Rif A and Seybold, Bryan and others},
  booktitle={2017 ieee international conference on acoustics, speech and signal processing (icassp)},
  pages={131--135},
  year={2017},
  organization={IEEE}
}

@article{138_liong2018less,
  title={Less is more: Micro-expression recognition from video using apex frame},
  author={Liong, Sze-Teng and See, John and Wong, KokSheik and Phan, Raphael C-W},
  journal={Signal Processing: Image Communication},
  volume={62},
  pages={82--92},
  year={2018},
  publisher={Elsevier}
}

@inproceedings{139_zhang2022review,
  title={A Review of Micro-expression Recognition based on Deep Learning},
  author={Zhang, He and Zhang, Hanling},
  booktitle={2022 International Joint Conference on Neural Networks (IJCNN)},
  pages={01--08},
  year={2022},
  organization={IEEE}
}

@inproceedings{140_ballester2016performance,
  title={On the performance of GoogLeNet and AlexNet applied to sketches},
  author={Ballester, Pedro and Araujo, Ricardo},
  booktitle={Proceedings of the AAAI conference on artificial intelligence},
  volume={30},
  number={1},
  year={2016}
}

@article{141_sengupta2019going,
  title={Going deeper in spiking neural networks: VGG and residual architectures},
  author={Sengupta, Abhronil and Ye, Yuting and Wang, Robert and Liu, Chiao and Roy, Kaushik},
  journal={Frontiers in neuroscience},
  volume={13},
  pages={95},
  year={2019},
  publisher={Frontiers Media SA}
}

@article{142_gan2019off,
  title={OFF-ApexNet on micro-expression recognition system},
  author={Gan, Yee Siang and Liong, Sze-Teng and Yau, Wei-Chuen and Huang, Yen-Chang and Tan, Lit-Ken},
  journal={Signal Processing: Image Communication},
  volume={74},
  pages={129--139},
  year={2019},
  publisher={Elsevier}
}

@inproceedings{143_zhou2019dual,
  title={Dual-inception network for cross-database micro-expression recognition},
  author={Zhou, Ling and Mao, Qirong and Xue, Luoyang},
  booktitle={2019 14th IEEE International Conference on Automatic Face \& Gesture Recognition (FG 2019)},
  pages={1--5},
  year={2019},
  organization={IEEE}
}

@article{144_zhang2022short,
  title={Short and long range relation based spatio-temporal transformer for micro-expression recognition},
  author={Zhang, Liangfei and Hong, Xiaopeng and Arandjelovi{\'c}, Ognjen and Zhao, Guoying},
  journal={IEEE Transactions on Affective Computing},
  volume={13},
  number={4},
  pages={1973--1985},
  year={2022},
  publisher={IEEE}
}

@article{145_su2020msaf,
  title={Msaf: Multimodal split attention fusion},
  author={Su, Lang and Hu, Chuqing and Li, Guofa and Cao, Dongpu},
  journal={arXiv preprint arXiv:2012.07175},
  year={2020}
}

@inproceedings{146_fukui2016multimodal,
  title={Multimodal Compact Bilinear Pooling for Visual Question Answering and Visual Grounding},
  author={Fukui, Akira and Park, Dong Huk and Yang, Daylen and Rohrbach, Anna and Darrell, Trevor and Rohrbach, Marcus},
  booktitle={Proceedings of the 2016 Conference on Empirical Methods in Natural Language Processing},
  pages={457--468},
  year={2016}
}

@inproceedings{147_joze2020mmtm,
  title={MMTM: Multimodal transfer module for CNN fusion},
  author={Joze, Hamid Reza Vaezi and Shaban, Amirreza and Iuzzolino, Michael L and Koishida, Kazuhito},
  booktitle={Proceedings of the IEEE/CVF conference on computer vision and pattern recognition},
  pages={13289--13299},
  year={2020}
}

@article{148_verbitskiy2022eranns,
  title={Eranns: Efficient residual audio neural networks for audio pattern recognition},
  author={Verbitskiy, Sergey and Berikov, Vladimir and Vyshegorodtsev, Viacheslav},
  journal={Pattern Recognition Letters},
  volume={161},
  pages={38--44},
  year={2022},
  publisher={Elsevier}
}

@article{149_fu2021cross,
  title={A cross-modal fusion network based on self-attention and residual structure for multimodal emotion recognition},
  author={Fu, Ziwang and Liu, Feng and Wang, Hanyang and Qi, Jiayin and Fu, Xiangling and Zhou, Aimin and Li, Zhibin},
  journal={arXiv preprint arXiv:2111.02172},
  year={2021}
}

@inproceedings{150_tsai2019multimodal,
  title={Multimodal transformer for unaligned multimodal language sequences},
  author={Tsai, Yao-Hung Hubert and Bai, Shaojie and Liang, Paul Pu and Kolter, J Zico and Morency, Louis-Philippe and Salakhutdinov, Ruslan},
  booktitle={Proceedings of the conference. Association for Computational Linguistics. Meeting},
  volume={2019},
  pages={6558},
  year={2019},
  organization={NIH Public Access}
}

@inproceedings{151_chumachenko2022self,
  title={Self-attention fusion for audiovisual emotion recognition with incomplete data},
  author={Chumachenko, Kateryna and Iosifidis, Alexandros and Gabbouj, Moncef},
  booktitle={2022 26th International Conference on Pattern Recognition (ICPR)},
  pages={2822--2828},
  year={2022},
  organization={IEEE}
}

@inproceedings{152_hu2018squeeze,
  title={Squeeze-and-excitation networks},
  author={Hu, Jie and Shen, Li and Sun, Gang},
  booktitle={Proceedings of the IEEE conference on computer vision and pattern recognition},
  pages={7132--7141},
  year={2018}
}

@article{153_ma2023emotion2vec,
  title={emotion2vec: Self-supervised pre-training for speech emotion representation},
  author={Ma, Ziyang and Zheng, Zhisheng and Ye, Jiaxin and Li, Jinchao and Gao, Zhifu and Zhang, Shiliang and Chen, Xie},
  journal={arXiv preprint arXiv:2312.15185},
  year={2023}
}

@article{154_kahou2016emonets,
  title={Emonets: Multimodal deep learning approaches for emotion recognition in video},
  author={Kahou, Samira Ebrahimi and Bouthillier, Xavier and Lamblin, Pascal and Gulcehre, Caglar and Michalski, Vincent and Konda, Kishore and Jean, S{\'e}bastien and Froumenty, Pierre and Dauphin, Yann and Boulanger-Lewandowski, Nicolas and others},
  journal={Journal on Multimodal User Interfaces},
  volume={10},
  pages={99--111},
  year={2016},
  publisher={Springer}
}

@inproceedings{155_radford2021learning,
  title={Learning transferable visual models from natural language supervision},
  author={Radford, Alec and Kim, Jong Wook and Hallacy, Chris and Ramesh, Aditya and Goh, Gabriel and Agarwal, Sandhini and Sastry, Girish and Askell, Amanda and Mishkin, Pamela and Clark, Jack and others},
  booktitle={International conference on machine learning},
  pages={8748--8763},
  year={2021},
  organization={PMLR}
}

@article{156_fang2024eva,
  title={Eva-02: A visual representation for neon genesis},
  author={Fang, Yuxin and Sun, Quan and Wang, Xinggang and Huang, Tiejun and Wang, Xinlong and Cao, Yue},
  journal={Image and Vision Computing},
  volume={149},
  pages={105171},
  year={2024},
  publisher={Elsevier}
}

@inproceedings{157_wang2024building,
  title={Building Robust Video-Level Deepfake Detection via Audio-Visual Local-Global Interactions},
  author={Wang, Yifan and Wu, Xuecheng and Zhang, Jia and Jing, Mohan and Lu, Keda and Yu, Jun and Su, Wen and Gao, Fang and Liu, Qingsong and Sun, Jianqing and others},
  booktitle={Proceedings of the 32nd ACM International Conference on Multimedia},
  pages={11370--11376},
  year={2024}
}

@inproceedings{158_he2020momentum,
  title={Momentum contrast for unsupervised visual representation learning},
  author={He, Kaiming and Fan, Haoqi and Wu, Yuxin and Xie, Saining and Girshick, Ross},
  booktitle={Proceedings of the IEEE/CVF conference on computer vision and pattern recognition},
  pages={9729--9738},
  year={2020}
}

@inproceedings{wang2024multimodal,
  title={Multimodal llm enhanced cross-lingual cross-modal retrieval},
  author={Wang, Yabing and Wang, Le and Zhou, Qiang and Wang, Zhibin and Li, Hao and Hua, Gang and Tang, Wei},
  booktitle={Proceedings of the 32nd ACM International Conference on Multimedia},
  pages={8296--8305},
  year={2024}
}

@article{wang2024dual,
  title={Dual-view Curricular Optimal Transport for Cross-lingual Cross-modal Retrieval},
  author={Wang, Yabing and Wang, Shuhui and Luo, Hao and Dong, Jianfeng and Wang, Fan and Han, Meng and Wang, Xun and Wang, Meng},
  journal={IEEE Transactions on Image Processing},
  volume={33},
  pages={1522--1533},
  year={2024},
  publisher={IEEE}
}

@inproceedings{wucvpr,
  title={AVF-MAE++: Scaling Affective Video Facial Masked Autoencoders via Efficient Audio-Visual Self-Supervised Learning},
  author={Wu, Xuecheng and Sun, Heli and Wang, Yifan and Nie, Jiayu and Zhang, Jie and Wang, Yabing and Xue, Junxiao and He, Liang},
  booktitle={Proceedings of the Computer Vision and Pattern Recognition Conference},
  pages={9142--9153},
  year={2025}
}

@misc{wu2025hola,
      title={HOLA: Enhancing Audio-visual Deepfake Detection via Hierarchical Contextual Aggregations and Efficient Pre-training}, 
      author={Xuecheng Wu and Danlei Huang and Heli Sun and Xinyi Yin and Yifan Wang and Hao Wang and Jia Zhang and Fei Wang and Peihao Guo and Suyu Xing and Junxiao Xue and Liang He},
      year={2025},
      eprint={2507.22781},
      archivePrefix={arXiv},
      primaryClass={cs.CV},
      url={https://arxiv.org/abs/2507.22781}, 
}

@article{new_32_desplanques2020ecapa,
  title={Ecapa-tdnn: Emphasized channel attention, propagation and aggregation in tdnn based speaker verification},
  author={Desplanques, Brecht and Thienpondt, Jenthe and Demuynck, Kris},
  journal={arXiv preprint arXiv:2005.07143},
  year={2020}
}

@article{34_gao2019res2net,
  title={Res2net: A new multi-scale backbone architecture},
  author={Gao, Shang-Hua and Cheng, Ming-Ming and Zhao, Kai and Zhang, Xin-Yu and Yang, Ming-Hsuan and Torr, Philip},
  journal={IEEE transactions on pattern analysis and machine intelligence},
  volume={43},
  number={2},
  pages={652--662},
  year={2019},
  publisher={IEEE}
}

@article{38_wang2023cam++,
  title={CAM++: A Fast and Efficient Network For Speaker Verification Using Context-Aware Masking},
  author={Wang, Hui and Zheng, Siqi and Chen, Yafeng and Cheng, Luyao and Chen, Qian},
  journal={arXiv preprint arXiv:2303.00332},
  year={2023}
}

@inproceedings{hu2018squeeze,
  title={Squeeze-and-excitation networks},
  author={Hu, Jie and Shen, Li and Sun, Gang},
  booktitle={Proceedings of the IEEE conference on computer vision and pattern recognition},
  pages={7132--7141},
  year={2018}
}

@article{27_kong2020panns,
  title={Panns: Large-scale pretrained audio neural networks for audio pattern recognition},
  author={Kong, Qiuqiang and Cao, Yin and Iqbal, Turab and Wang, Yuxuan and Wang, Wenwu and Plumbley, Mark D},
  journal={IEEE/ACM Transactions on Audio, Speech, and Language Processing},
  volume={28},
  pages={2880--2894},
  year={2020},
  publisher={IEEE}
}

@inproceedings{43_carreira2017quo,
  title={Quo vadis, action recognition? a new model and the kinetics dataset},
  author={Carreira, Joao and Zisserman, Andrew},
  booktitle={proceedings of the IEEE Conference on Computer Vision and Pattern Recognition},
  pages={6299--6308},
  year={2017}
}

@inproceedings{32_tran2015learning,
  title={Learning spatiotemporal features with 3d convolutional networks},
  author={Tran, Du and Bourdev, Lubomir and Fergus, Rob and others},
  booktitle={Proceedings of the IEEE international conference on computer vision},
  pages={4489--4497},
  year={2015}
}

@inproceedings{19_liu2022video,
  title={Video swin transformer},
  author={Liu, Ze and Ning, Jia and Cao, Yue and Wei, Yixuan and Zhang, Zheng and Lin, Stephen and Hu, Han},
  booktitle={Proceedings of the IEEE/CVF conference on computer vision and pattern recognition},
  pages={3202--3211},
  year={2022}
}

@inproceedings{30_bertasius2021space,
  title={Is space-time attention all you need for video understanding?},
  author={Bertasius, Gedas and Wang, Heng and Torresani, Lorenzo},
  booktitle={ICML},
  volume={2},
  number={3},
  pages={4},
  year={2021}
}

@misc{openai2025o3mini,
  author    = {OpenAI},
  title     = {OpenAI o3},
  howpublished = {\url{https://openai.com/index/introducing-o3-and-o4-mini/}},
  year      = {2025}
}

@misc{Gemini-2.5-pro-preview-03-25,
  author    = {Gemini Team},
  title     = {Gemini-2.5-pro-preview-03-25},
  howpublished = {\url{https://deepmind.google/technologies/gemini/pro/}},
  year      = {2025}
}

@misc{openai2024gpt4o,
  author    = {OpenAI},
  title     = {Hello GPT-4o},
  howpublished = {\url{https://openai.com/index/hello-gpt-4o/}},
  year      = {2024}
}

@misc{Qwen-QVQ-72B,
    title = {QVQ: To See the World with Wisdom},
    howpublished ={\url{https://qwenlm.github.io/blog/qvq-72b-preview/}},
    author = {Qwen Team},
    month = {December},
    year = {2024}
}

@article{Qwen2d5-vl,
  title={Qwen2.5-vl technical report},
  author={Bai, Shuai and Chen, Keqin and Liu, Xuejing and Wang, Jialin and Ge, Wenbin and Song, Sibo and Dang, Kai and Wang, Peng and Wang, Shijie and Tang, Jun and others},
  journal={arXiv preprint arXiv:2502.13923},
  year={2025}
}

@misc{grok-4,
  author    = {xAI},
  title     = {xAI Grok 4},
  howpublished = {\url{https://x.ai/news/grok-4}},
  year      = {2025}
}

@article{48_abdin2024phi,
  title={Phi-3 technical report: A highly capable language model locally on your phone},
  author={Abdin, Marah and Jacobs, Sam Ade and Awan, Ammar Ahmad and Aneja, Jyoti and Awadallah, Ahmed and Awadalla, Hany and Bach, Nguyen and Bahree, Amit and Bakhtiari, Arash and Behl, Harkirat and others},
  journal={arXiv preprint arXiv:2404.14219},
  year={2024}
}

@misc{Doubao-1.5-vision-pro-32k,
  author    = {ByteDance},
  title     = {Doubao-1.5-vision-pro-32k},
  howpublished = {\url{https://volcengine.com/product/doubao}},
  year      = {2025}
}

@article{zhang2022sadtalker,
  title={SadTalker: Learning Realistic 3D Motion Coefficients for Stylized Audio-Driven Single Image Talking Face Animation},
  author={Zhang, Wenxuan and Cun, Xiaodong and Wang, Xuan and Zhang, Yong and Shen, Xi and Guo, Yu and Shan, Ying and Wang, Fei},
  journal={arXiv preprint arXiv:2211.12194},
  year={2022}
}

@misc{wei2024aniportrait,
      title={AniPortrait: Audio-Driven Synthesis of Photorealistic Portrait Animations}, 
      author={Huawei Wei and Zejun Yang and Zhisheng Wang},
      year={2024},
      eprint={2403.17694},
      archivePrefix={arXiv},
      primaryClass={cs.CV}
}

@InProceedings{Ji_sonic,
    author    = {Ji, Xiaozhong and Hu, Xiaobin and Xu, Zhihong and Zhu, Junwei and Lin, Chuming and He, Qingdong and Zhang, Jiangning and Luo, Donghao and Chen, Yi and Lin, Qin and Lu, Qinglin and Wang, Chengjie},
    title     = {Sonic: Shifting Focus to Global Audio Perception in Portrait Animation},
    booktitle = {Proceedings of the IEEE/CVF Conference on Computer Vision and Pattern Recognition (CVPR)},
    month     = {June},
    year      = {2025},
    pages     = {193-203}
}

@inproceedings{echomimic,
  author    = {Zhiyuan Chen and Jiajiong Cao and Zhiquan Chen and Yuming Li and Chenguang Ma},
  title     = {EchoMimic: Lifelike Audio-Driven Portrait Animations through Editable Landmark Conditions},
  booktitle = {Proceedings of the 39th AAAI Conference on Artificial Intelligence (AAAI-25)},
  year      = {2025},
  publisher = {AAAI Press},
  doi       = {10.1609/aaai.v39i3.32241},
  url       = {https://doi.org/10.1609/aaai.v39i3.32241},
  pages     = {268:1--268:8},
}

@InProceedings{hallo3,
    author    = {Cui, Jiahao and Li, Hui and Zhan, Yun and Shang, Hanlin and Cheng, Kaihui and Ma, Yuqi and Mu, Shan and Zhou, Hang and Wang, Jingdong and Zhu, Siyu},
    title     = {Hallo3: Highly Dynamic and Realistic Portrait Image Animation with Video Diffusion Transformer},
    booktitle = {Proceedings of the IEEE/CVF Conference on Computer Vision and Pattern Recognition (CVPR)},
    month     = {June},
    year      = {2025},
    pages     = {21086-21095}
}

@article{wang2025fantasytalking,
   title={FantasyTalking: Realistic Talking Portrait Generation via Coherent Motion Synthesis},
   author={Wang, Mengchao and Wang, Qiang and Jiang, Fan and Fan, Yaqi and Zhang, Yunpeng and Qi, Yonggang and Zhao, Kun and Xu, Mu},
   journal={arXiv preprint arXiv:2504.04842},
   year={2025}
 }

@misc{hunyuanavatar,
      title={HunyuanVideo-Avatar: High-Fidelity Audio-Driven Human Animation for Multiple Characters}, 
      author={Yi Chen and Sen Liang and Zixiang Zhou and Ziyao Huang and Yifeng Ma and Junshu Tang and Qin Lin and Yuan Zhou and Qinglin Lu},
      year={2025},
      eprint={2505.20156},
      archivePrefix={arXiv},
      primaryClass={cs.CV},
      url={https://arxiv.org/abs/2505.20156}, 
}

@misc{multitalk,
      title={Let Them Talk: Audio-Driven Multi-Person Conversational Video Generation}, 
      author={Zhe Kong and Feng Gao and Yong Zhang and Zhuoliang Kang and Xiaoming Wei and Xunliang Cai and Guanying Chen and Wenhan Luo},
      year={2025},
      eprint={2505.22647},
      archivePrefix={arXiv},
      primaryClass={cs.CV},
      url={https://arxiv.org/abs/2505.22647}, 
}

@misc{ominiavatar,
      title={OmniAvatar: Efficient Audio-Driven Avatar Video Generation with Adaptive Body Animation}, 
      author={Qijun Gan and Ruizi Yang and Jianke Zhu and Shaofei Xue and Steven Hoi},
      year={2025},
      eprint={2506.18866},
      archivePrefix={arXiv},
      primaryClass={cs.CV},
      url={https://arxiv.org/abs/2506.18866}, 
}

@misc{stableavatar,
      title={StableAvatar: Infinite-Length Audio-Driven Avatar Video Generation}, 
      author={Shuyuan Tu and Yueming Pan and Yinming Huang and Xintong Han and Zhen Xing and Qi Dai and Chong Luo and Zuxuan Wu and Yu-Gang Jiang},
      year={2025},
      eprint={2508.08248},
      archivePrefix={arXiv},
      primaryClass={cs.CV},
      url={https://arxiv.org/abs/2508.08248}, 
}

@InProceedings{Syncnet,
  author       = "Chung, J.~S. and Zisserman, A.",
  title        = "Out of time: automated lip sync in the wild",
  booktitle    = "Workshop on Multi-view Lip-reading, ACCV",
  year         = "2016",
}

@article{Qwen-VL,
  title={Qwen-VL: A Versatile Vision-Language Model for Understanding, Localization, Text Reading, and Beyond},
  author={Bai, Jinze and Bai, Shuai and Yang, Shusheng and Wang, Shijie and Tan, Sinan and Wang, Peng and Lin, Junyang and Zhou, Chang and Zhou, Jingren},
  journal={arXiv preprint arXiv:2308.12966},
  year={2023}
}

@inproceedings{wu2025towards,
  title={Towards Emotion Analysis in Short-form Videos: A Large-Scale Dataset and Baseline},
  author={Wu, Xuecheng and Sun, Heli and Xue, Junxiao and Nie, Jiayu and Kong, Xiangyan and Zhai, Ruofan and Huang, Danlei and He, Liang},
  booktitle={Proceedings of the 2025 International Conference on Multimedia Retrieval},
  pages={1497--1506},
  year={2025}
}

@article{zhang2025hkd4vlm,
  title={HKD4VLM: A Progressive Hybrid Knowledge Distillation Framework for Robust Multimodal Hallucination and Factuality Detection in VLMs},
  author={Zhang, Zijian and Wu, Xuecheng and Huang, Danlei and Yan, Siyu and Peng, Chong and Cao, Xuezhi},
  journal={arXiv preprint arXiv:2506.13038},
  year={2025}
}

\end{document}